\documentclass{article} 
\usepackage{colm2024_conference}

\usepackage{graphicx} 
\usepackage{subcaption} 
\usepackage{csvsimple}
\usepackage{tabularx}
\usepackage{longtable}
\usepackage{float}
\usepackage{listings}

\usepackage{microtype}
\usepackage{hyperref}
\usepackage{url}
\usepackage{booktabs}
\usepackage{multirow}
\definecolor{darkblue}{rgb}{0, 0, 0.5}
\hypersetup{colorlinks=true, citecolor=darkblue, linkcolor=darkblue, urlcolor=darkblue}

\title{The Fine Line: \\ Navigating Large Language Model Pretraining with Down-streaming Capability Analysis}

\newcommand*\samethanks[1][\value{footnote}]{\footnotemark[#1]}
\author{Chen Yang\textsuperscript{2}\thanks{Equal Technical Contributions.},\ \ 
Junzhuo Li\textsuperscript{3}\samethanks[1]\ \ 
\\
\textbf{Xinyao Niu}\textsuperscript{4},\ 
\textbf{Xinrun Du}\textsuperscript{1}, \ 
\textbf{Songyang Gao}\textsuperscript{5},\ 
\textbf{Haoran Zhang}\textsuperscript{6}\ 
\\
\textbf{Zhaoliang Chen}\textsuperscript{7}, \
\textbf{Xingwei Qu}\textsuperscript{1 8},\ 
\textbf{Ruibin Yuan}\textsuperscript{1 8},\ 
\textbf{Yizhi Li}\textsuperscript{1 9}\ 
\\
\textbf{Jiaheng Liu}\textsuperscript{1},\ 
\textbf{Stephen W. Huang}\textsuperscript{10},\ 
\textbf{Shawn Yue}\textsuperscript{10}\ 
\\ 
\textbf{Ge Zhang}\textsuperscript{1 11 12}\samethanks[1]~~\samethanks[2] \ 
\\
    \textsuperscript{1}Multimodal Art Projection Research Community,\quad
    \textsuperscript{2}Peking University\quad
\\ 
    \textsuperscript{3}Tianjin University,\quad
    \textsuperscript{4}University of Melbourne,\quad
    \textsuperscript{5}Fudan University\quad
\\
    \textsuperscript{6}University of Illinois at Urbana-Champaign,\quad
    \textsuperscript{7}Emory University\quad
\\
    \textsuperscript{8}HKUST,\quad
    \textsuperscript{9}University of Manchester,\quad
    \textsuperscript{10}harmony.ai\quad
\\
    \textsuperscript{11}University of Waterloo,\quad
    \textsuperscript{12}Vector Institute\quad
}

\colmfinalcopy 
\begin{document}

\maketitle

\begin{abstract}

Uncovering early-stage metrics that reflect final model performance is one core principle for large-scale pretraining. The existing scaling law demonstrates the power-law correlation between pretraining loss and training flops, which serves as an important indicator of the current training state for large language models. However, this principle only focuses on the model's compression properties on the training data, resulting in an inconsistency with the ability improvements on the downstream tasks. Some follow-up works attempted to extend the scaling-law to more complex metrics (such as hyperparameters), but still lacked a comprehensive analysis of the dynamic differences among various capabilities during pretraining. 
To address the aforementioned limitations,
in this paper, we compare the model capabilities across numerous pretraining intermediate checkpoints, confirming the similar training dynamics of specific downstream metrics across different model sizes. This generalization holds for models with up to 67 billion parameters.
Besides, we provide empirical summaries, including performance comparisons of different models and capabilities, and tuition of key metrics for different training phases.
Based on these findings,
we provide a more user-friendly strategy for evaluating the optimization state, offering guidance for establishing a stable pretraining process.

\end{abstract}

\section{Introduction}


Large Language Models (LLMs) have shown great promise as highly capable AI assistants that excel in complex reasoning tasks requiring expert knowledge across a wide range of fields (e.g., programming and creative writing).  
However,
training large language models requires significant pretraining computation costs on large-scale training data.
To reduce the computation costs,
scaling law~\citep{kaplan2020scaling} is proposed to illustrate the power-law relationship between pretraining loss and computational effort,
which has provided valuable insights into model optimization with minimal computational cost. 
Recently, several findings, such as those exploring the phenomena of emergence~\citep{wei2022emergent} and the Broken Neural Scaling Laws~\citep{caballero2023broken}, indicate these laws might not fully capture model capabilities, particularly for downstream tasks.
Therefore, it is important to expand and refine our evaluative frameworks.

In this work,
we first investigate the dynamics of several open-sourced large language models (i.e., Baichuan-7B~\citep{yang2023baichuan}, DeepSeek-7B~\citep{deepseekai2024deepseek}, Amber-7B~\citep{liu2023llm360}, OpenLLaMA-7B~\citep{openlm2023openllama,touvron2023llama,together2023redpajama}, Yi-34B~\citep{ai2024yi} and DeepSeek-67B), and analyze their performance results across diverse tasks using the corresponding intermediate checkpoints based on the number of pre-trained tokens. Then, 
based on the theoretical framework of the Scaling Law~\citep{kaplan2020scaling}, we analyze the performance patterns of different models across various downstream tasks, and provide findings to facilitate further research on the training dynamics of the LLMs,
where the findings are shown as follows:





\begin{itemize}
    \item \textbf{Findings on task dynamic prediction:} 
Within the training process of LLMs, we observe that the dynamics of existing downstream tasks within a domain can predict the dynamics of unseen tasks. This suggests that a model's performance on known tasks can inform us about how it might perform on similar, yet unseen tasks within the same domain. (Section \ref{cross-task analysis})

    \item \textbf{Findings on cross-domain promotion:}
Similar to the human cognitive process, the enhancement of different abilities across various domains progresses from basic to advanced levels following curriculum learning. The curriculum between cross-domain tasks can guide the training direction of models and the insights gained from one domain can potentially promote the learning process of other domains. (Section \ref{cross-domain analysis})

    \item \textbf{Findings on the effect of training strategies, model architecture, etc. :}
    Based on the results of several 7b-scale models (Baichuan2-7b, DeepSeek-7b, OpenLLaMA-7b, and Amber-7b), we comprehensively analyze the effect of training strategies, dataset quality, and model architecture in training LLMs. For example, we observe that the training dataset, learning rate adjustments, batch size, and regularization techniques play a significant role in the learning efficiency in the early training stage  (Section \ref{analysis within the 7b scale}).

    \item \textbf{Findings on the effect of model scale on reasoning tasks:}
    Based on the results of Deepseek-7b, 34b and 67b,
    we observe that the model size and complexity significantly influence its ability on these reasoning tasks. Nonetheless, employing particular strategies can enhance smaller-scale models to obtain similar performance on commonsense reasoning when compared with their larger counterparts. (Section \ref{analysis across scales})

    \item \textbf{Findings on the scaling law:} 
    (1). We observe that larger training datasets lead to improved model performance on various benchmarks, which demonstrates the effect of extensive training data in training LLMs. However, the benefit of additional data diminishes as datasets grow, suggesting an approaching limit to performance gains. 
(2). We observe that the accuracies of the scaling law~\citep{hoffmann2022training} 
vary a lot across different models,
which indicates that factors such as model architecture and computational complexity significantly influence scaling efficiency. Notably, some models demonstrate better alignment with the law, suggesting potential advantages in data utilization and learning efficiency.
(3). Although the scaling law can provide a useful perspective on the impact of training data size, the actual performance scaling is nuanced, which reflects the complex interplay between data volume, model architecture, and computational strategies. This also highlights the importance of continued research into scaling laws and model optimization to maximize learning outcomes from available data (Section \ref{sec: scaling law}).
\end{itemize}

Moreover,
apart from the aforementioned findings,
it should be mentioned that we plan to publicly release the intermediate checkpoints of Amber-7B and OpenLLaMA-7B, 
which not only enhances the comprehension of scaling laws but also helps to develop more effective training strategies for LLMs.
In conclusion,
we hope our findings and open-sourced checkpoints can guide the developers to understand the optimization process of LLMs and facilitate the growth of foundation models.

\section{Related Work}

\subsection{Scaling Law}

Scaling laws have been identified as critical empirical relationships for LLMs that delineate the impact of changes in model size, dataset size, and computational budget on model performance. \citet{kaplan2020scaling} focused on the distribution of computational resources for the training of LLMs. However, this principle is particularly significant as it underscores the relative importance of different factors to model performance, such as model architecture and batch size are secondary to the overarching influence of model size and dataset scope in reducing final validation loss.  

Recent research, including studies by \citet{hoffmann2022training} and \citet{muennighoff2024scaling}, highlights the importance of a balanced approach in scaling both model size and training data to enhance model capabilities. Specifically, enlarging pre-training datasets has been emphasized as a crucial step forward. However, challenges arise in addressing data limitations, such as the potential pitfalls of up-sampling. \citet{hernandez2021scaling} demonstrated that increasing only 0.1\% of the training dataset by a factor of 100 significantly reduced model efficacy, pointing out the limitations of simple data amplification strategies. Meanwhile, \citet{muennighoff2024scaling} approach, which involved repeating the entire pre-training dataset across multiple epochs, showed promising results. 

\subsection{Emergent Abilities of Large Language Models}

The concept of emergent abilities in large language models (LLMs) draws researchers attention, as model sizes continue to scale. The phenomenon of "double descent," where a model's validation loss first worsens and then improves with increasing model complexity, was first proposed by \citet{nakkiran2019deep}. Furthermore, the "grokking" phenomenon, where models continue to improve even without a reduction in training loss, has been detailed by \citet{power2022grokking} and \citet{murty2023grokking}, providing insight into the non-linear learning trajectories of LLMs.

Recent research has extensively explored the emergent abilities of LLMs in downstream tasks. \citet{wei2022emergent} highlighted that emergent abilities are predominantly observed as models scale up, underscoring the significance of model size in achieving advanced capabilities. \citet{schaeffer2023emergent} contended that the performance of emergent abilities is highly dependent on the chosen evaluation metrics, suggesting a nuanced relationship between model performance and assessment methodologies. In our study, we concentrate on the emergent abilities of LLMs in downstream tasks, utilizing common evaluation frameworks as provided by \citet{2023opencompass}. Our aim is to conclude the dynamics of emergent abilities within the context of practical applications, contributing to a deeper understanding of how LLMs can performs through scaling and standard evaluation metrics.

\section{Methodology}

\subsection{Models}

In this study, our analysis extends to the intermediate checkpoints of a series of state-of-the-art large language models, each featuring unique architectural designs and training paradigms, showcasing the latest advancements in natural language processing and machine learning. Among these models, Amber-7B and OpenLLaMA-7B were replicated by us. The introduction to each model under scrutiny can be found in Appendix~\ref{a. models introduction}.

By examining these models across various stages of their training lifecycle, we aim to shed light on the dynamic interplay between model architecture, size, training strategies, and their impact on learning efficiency and task-specific performance. This comprehensive analysis not only assesses the current state of the art in language modeling but also explores the broader implications of model training and design choices on the future trajectory of large language model development.

\subsection{Datasets and evaluation metrics}

To rigorously assess the capabilities of our language models, we have curated a wide-ranging collection of datasets that span a broad array of cognitive and computational challenges. These datasets are crucial for evaluating the models' proficiency in different aspects of language understanding, reasoning, and generation. Specifically, our evaluation encompasses six categories: Code, Commonsense Reasoning, World Knowledge, Reading Comprehension, Math, and Examination.

Covering these six categories allows us to thoroughly examine the models' strengths and weaknesses across a diverse set of tasks, ensuring their robustness and adaptability. This comprehensive evaluation strategy enables us to identify areas where our models excel in understanding complex language constructs, applying commonsense knowledge, retrieving factual information, comprehending and analyzing written texts, solving mathematical problems, and performing in exam-like conditions. 

Our detailed evaluation setup and the metrics applied to each dataset are summarized in Table~\ref{tab:datasets_metrics_category}, showcasing the extensive measures taken to guarantee the robustness and flexibility of our language models across a wide range of tasks.

\begin{table}[ht]
\centering
\resizebox{\linewidth}{!}{
\begin{tabular}{lll}
\toprule
\textbf{Category} & \textbf{Dataset} & \textbf{Metrics} \\
\midrule
\multirow{2}{*}{Code} & HumanEval~\citep{chen2021evaluating} & \multirow{2}{*}{Pass@1} \\
& MBPP~\citep{austin2021program} & \\
\midrule
\multirow{8}{*}{\shortstack[l]{Commonsense\\Reasoning}} & PIQA~\citep{Bisk_Zellers_Le} & \multirow{7}{*}{Accuracy (ppl, 0-shot) }\\
& SIQA~\citep{sap2019socialiqa} &  \\
& HellaSwag~\citep{zellers2019hellaswag} &  \\
& WinoGrande~\citep{10.1145/3474381} & \\
& ARC Easy~\citep{clark2018think} &  \\
& ARC Challenge~\citep{clark2018think} &  \\
& OpenBookQA~\citep{mihaylov2018suit} &  \\
& CommonsenseQA~\citep{talmor2019commonsenseqa} & Accuracy (8-shot) \\
\midrule
\multirow{2}{*}{\shortstack[l]{World\\Knowledge}} & NaturalQuestions~\citep{10.1162/tacl_a_00276} & \multirow{2}{*}{Accuracy (0-shot)} \\
& TriviaQA~\citep{joshi2017triviaqa} &  \\
\midrule
\multirow{2}{*}{\shortstack[l]{Reading\\Comprehension}} & BoolQ~\citep{clark2019boolq} & Accuracy (ppl, 0-shot) \\
& SQuAD 2.0~\citep{rajpurkar2018know} & Accuracy (0-shot) \\
\midrule
\multirow{3}{*}{Math} & GSM8K~\citep{cobbe2021training} & \multirow{2}{*}{Accuracy (4-shot)} \\
& MATH~\citep{hendrycks2021measuring} &  \\
& TheoremQA~\citep{chen2023theoremqa} & Accuracy (0-shot) \\
\midrule
\multirow{3}{*}{Examination} & MMLU~\citep{hendrycks2021measuring} & \multirow{3}{*}{Accuracy (5-shot)}  \\
& CMMLU~\citep{li2023cmmlu} & \\
& CEVAL~\citep{huang2023ceval} & \\
\bottomrule
\end{tabular}}
\caption{Datasets and their evaluation metrics by category.}
\label{tab:datasets_metrics_category}
\end{table}


\section{Empirical Analysis}

\subsection{Intra-model analysis}
\label{intra-model analysis}


In this part of our discussion, we pay attention to an in-depth analysis of individual models, focusing on identifying whether the trends in their performance remain consistent across various metrics and observing the unique patterns that emerge across different benchmarks. 


\subsubsection{Cross-task analysis}
\label{cross-task analysis}

\begin{figure}[htbp]
    \centering    
    \begin{subfigure}[b]{0.3\textwidth}
        \includegraphics[width=\textwidth]{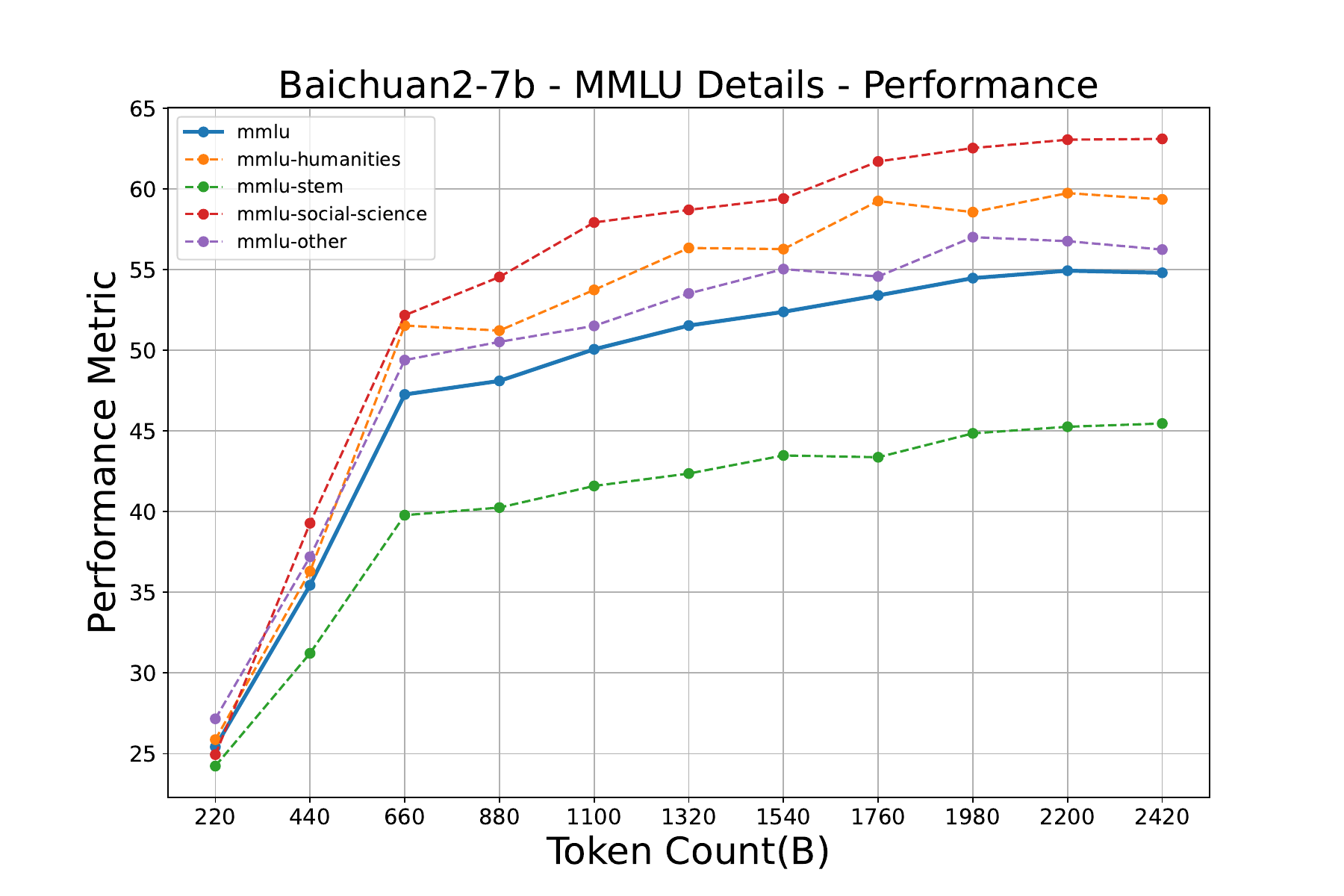}
        \caption{Baichuan-7B MMLU}
        \label{fig:baichuan7b-mmlu}
    \end{subfigure}
    \hfill
    \begin{subfigure}[b]{0.3\textwidth}
        \includegraphics[width=\textwidth]{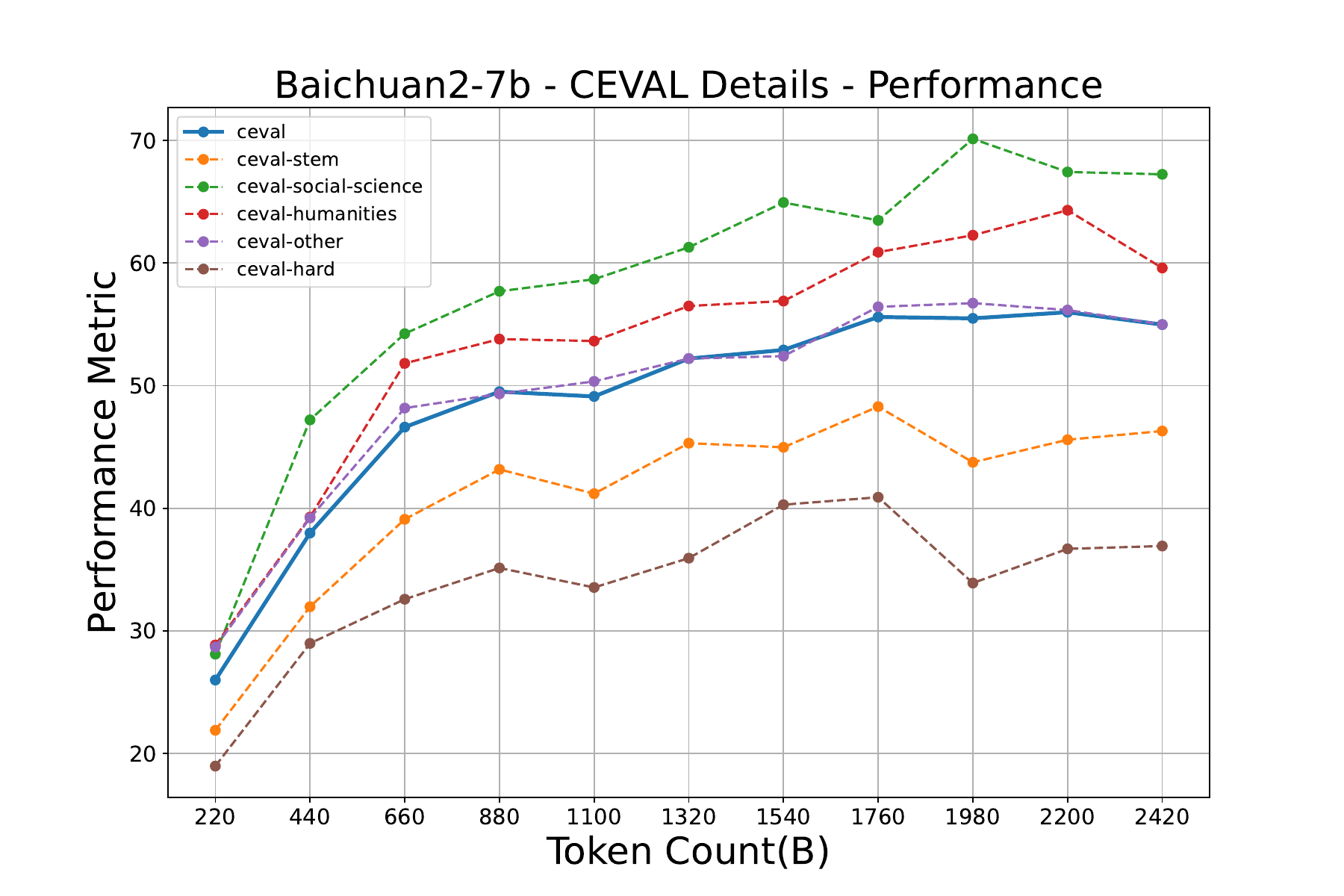}
        \caption{Baichuan-7B CEval}
        \label{fig:baichuan7b-ceval}
    \end{subfigure}
    \hfill
    \begin{subfigure}[b]{0.3\textwidth}
        \includegraphics[width=\textwidth]{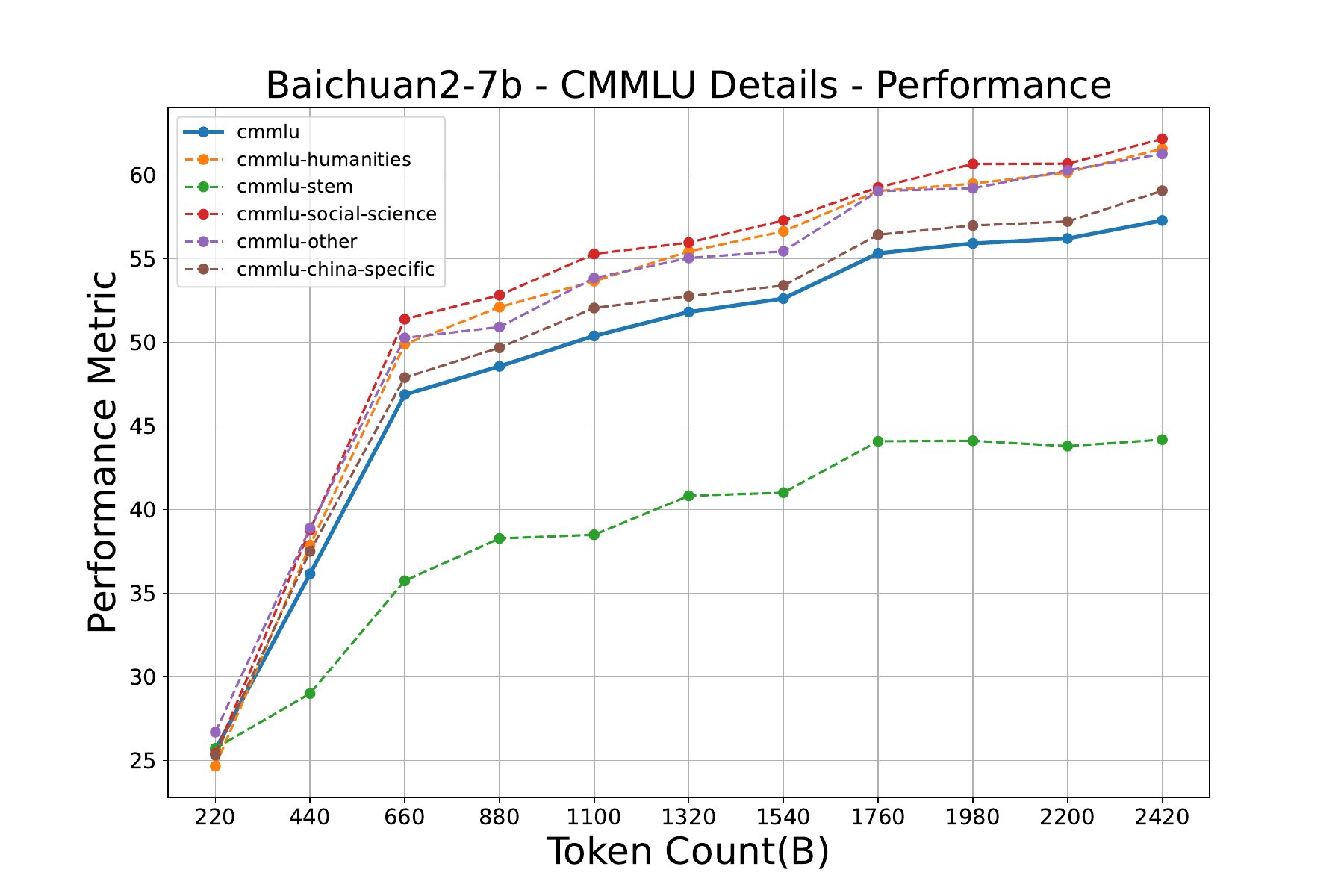}
        \caption{Baichuan-7B CMMLU}
        \label{fig:baichuan7b-cmmlu}
    \end{subfigure}
    
    \begin{subfigure}[b]{0.3\textwidth}
        \includegraphics[width=\textwidth]{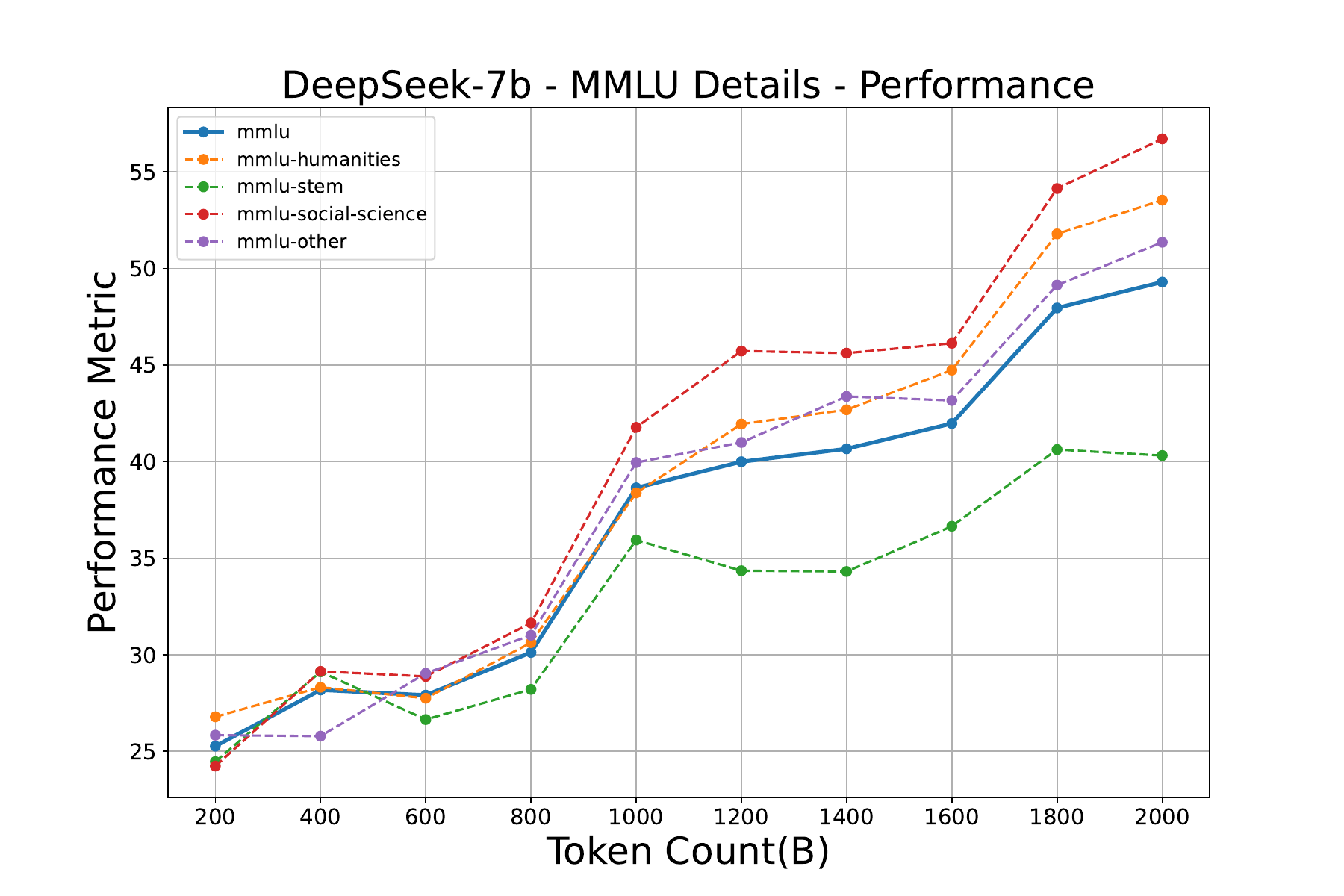}
        \caption{DeepSeek-7B MMLU}
        \label{fig:deepseek7b-mmlu}
    \end{subfigure}
    \hfill
    \begin{subfigure}[b]{0.3\textwidth}
        \includegraphics[width=\textwidth]{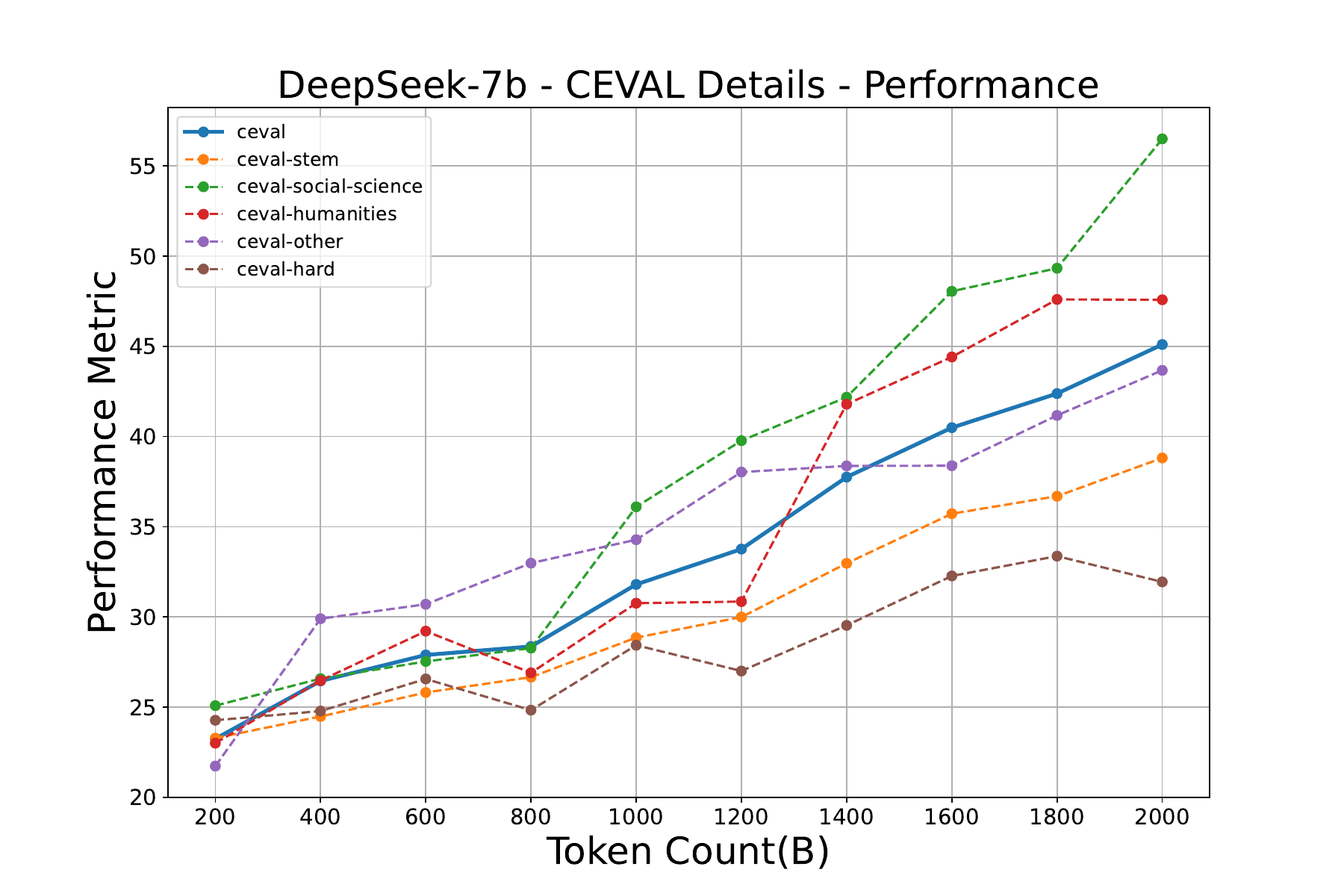}
        \caption{DeepSeek-7B CEval}
        \label{fig:deepseek7b-ceval}
    \end{subfigure}
    \hfill
    \begin{subfigure}[b]{0.3\textwidth}
        \includegraphics[width=\textwidth]{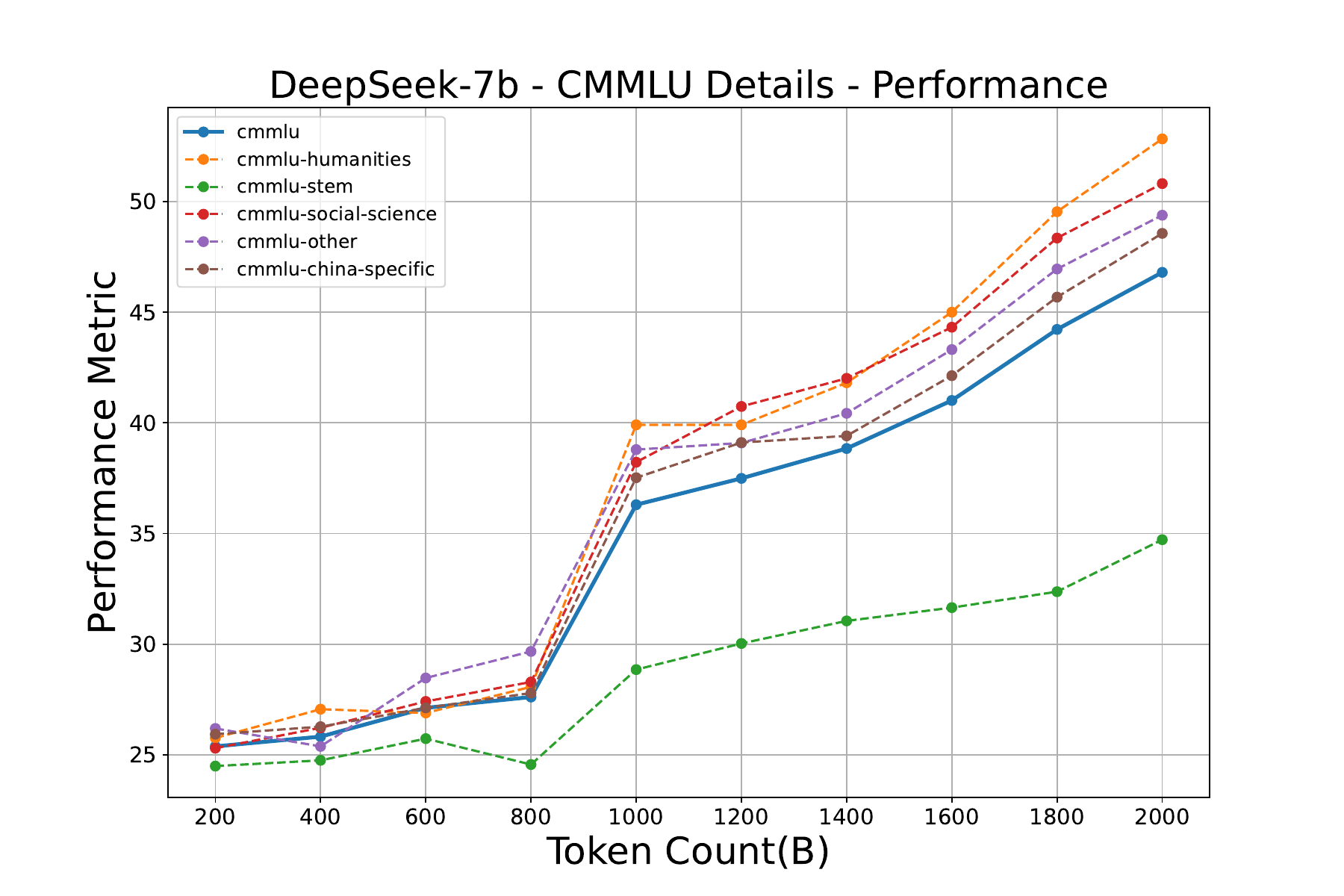}
        \caption{DeepSeek-7B CMMLU}
        \label{fig:deepseek7b-cmmlu}
    \end{subfigure}

    
    \begin{subfigure}[b]{0.3\textwidth}
        \includegraphics[width=\textwidth]{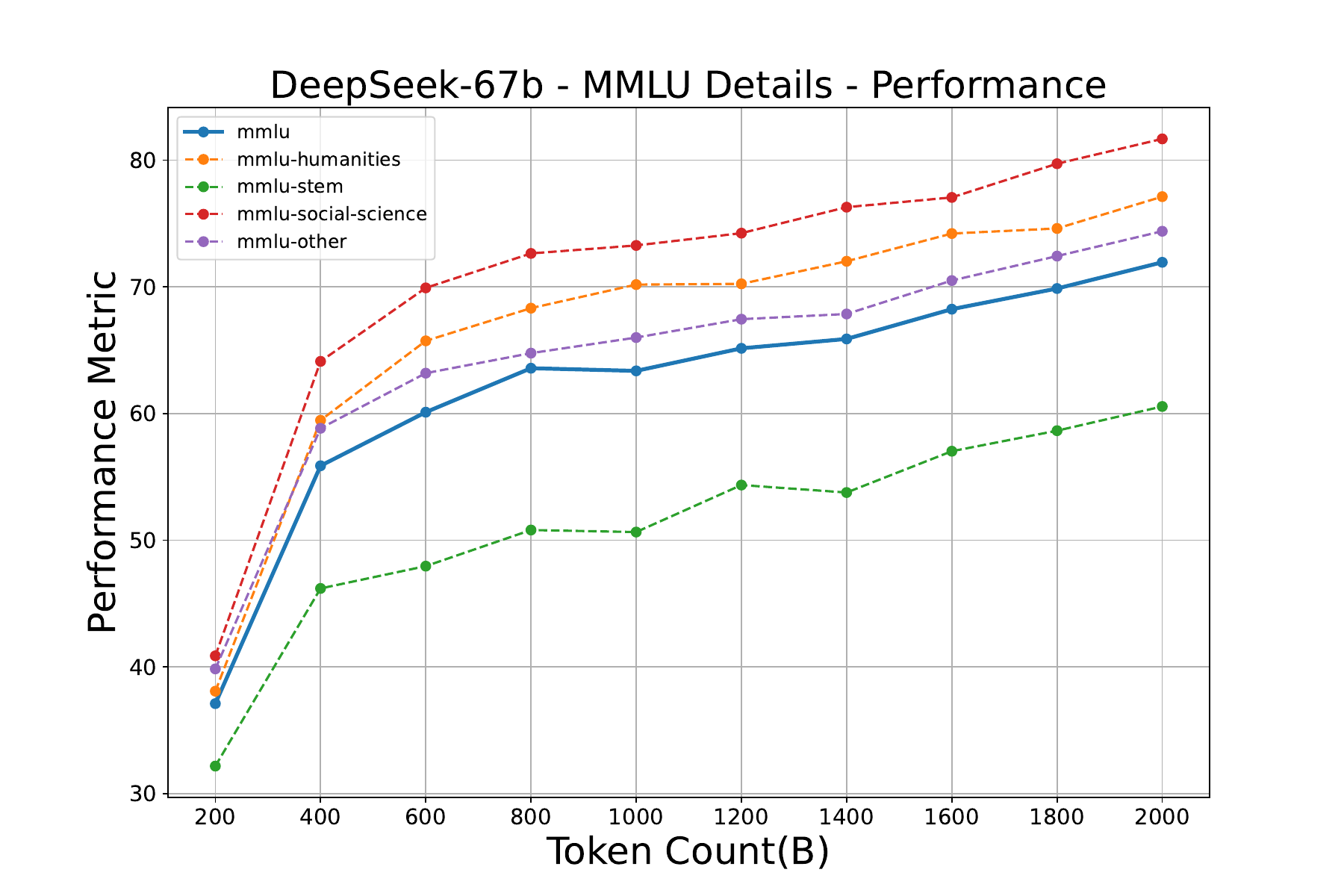}
        \caption{DeepSeek-67B MMLU}
        \label{fig:deepseek67b-mmlu}
    \end{subfigure}
    \hfill
    \begin{subfigure}[b]{0.3\textwidth}
        \includegraphics[width=\textwidth]{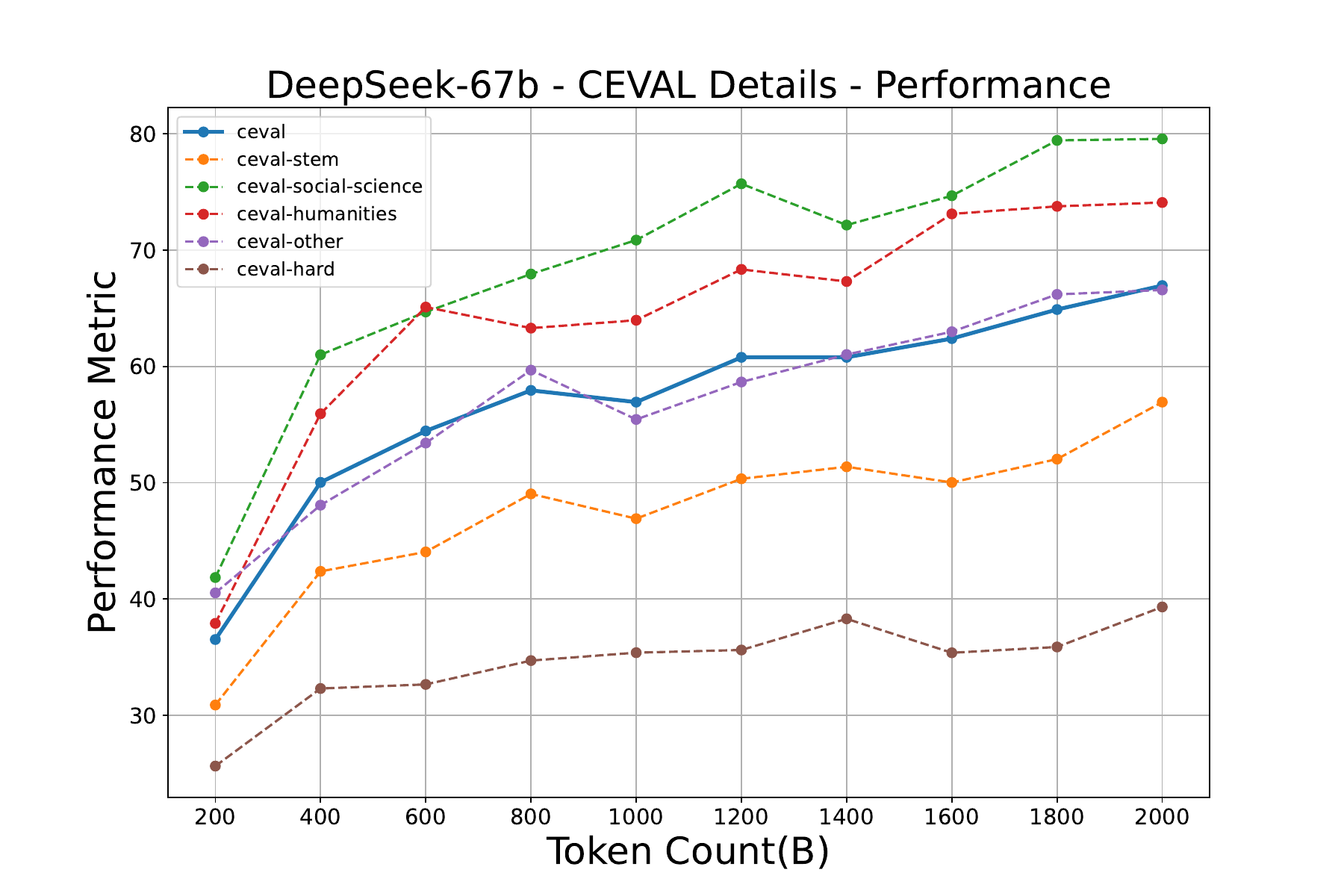}
        \caption{DeepSeek-67B CEval}
        \label{fig:deepseek67b-ceval}
    \end{subfigure}
    \hfill
    \begin{subfigure}[b]{0.3\textwidth}
        \includegraphics[width=\textwidth]{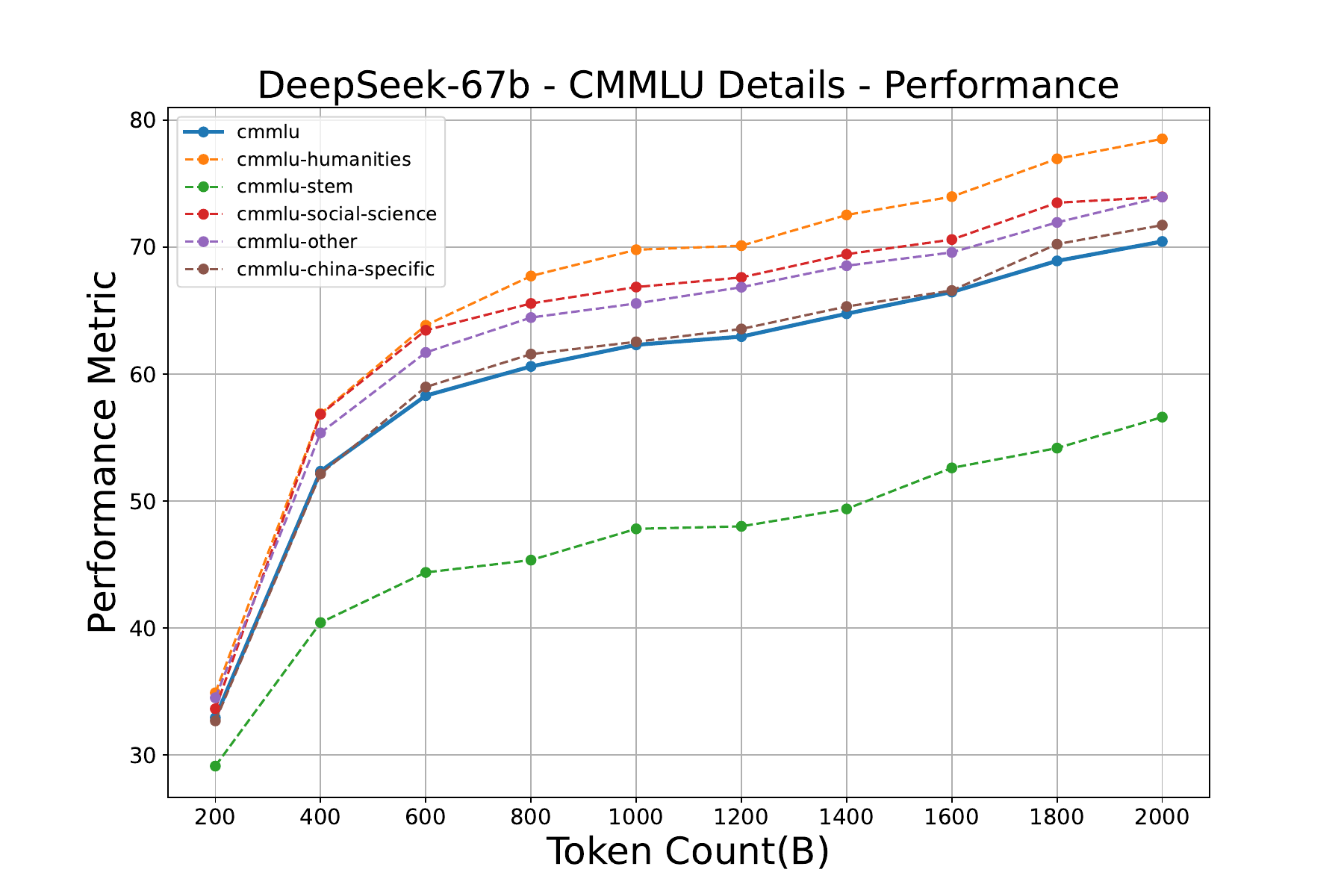}
        \caption{DeepSeek-67B CMMLU}
        \label{fig:deepseek67b-cmmlu}
    \end{subfigure}

    \caption{Comparative performance analysis of Baichuan-7B, DeepSeek-7B, and DeepSeek-67B models across MMLU, CMMLU, and CEval benchmarks.}
    \label{fig:models-comparison-exam}
\end{figure}

\paragraph{Analysis of examination domain}

The findings from our experiments are illustrated in Figure~\ref{fig:models-comparison-exam}, with extended experimental results accessible in Appendix~\ref{a. full models performance}. Analysis across the models, as demonstrated by the horizontal examination of the rows in the figure, reveals that the performance trends for the Examination benchmarks, namely MMLU, CMMLU, and CEVAL, exhibit remarkable consistency. This uniformity across various evaluation datasets within the Examination domain suggests inherent similarities in the tasks. 

Despite focusing on different aspects of examination-style questions and encompassing both English and Chinese languages, MMLU, CMMLU, and CEVAL appear to assess overlapping capabilities of the models, leading to similar performance trends. Furthermore, the training process of models demonstrates that the behavior of known tasks within a domain can predict the behavior of yet-to-be-encountered tasks. This suggests that understanding a model's performance on familiar tasks can offer valuable insights into its potential performance on analogous tasks that have not been previously explored within the same domain.

\subsubsection{Cross-domain analysis}
\label{cross-domain analysis}

\begin{figure}[htbp]
    \centering
    \begin{subfigure}[b]{0.3\textwidth}
        \includegraphics[width=\textwidth]{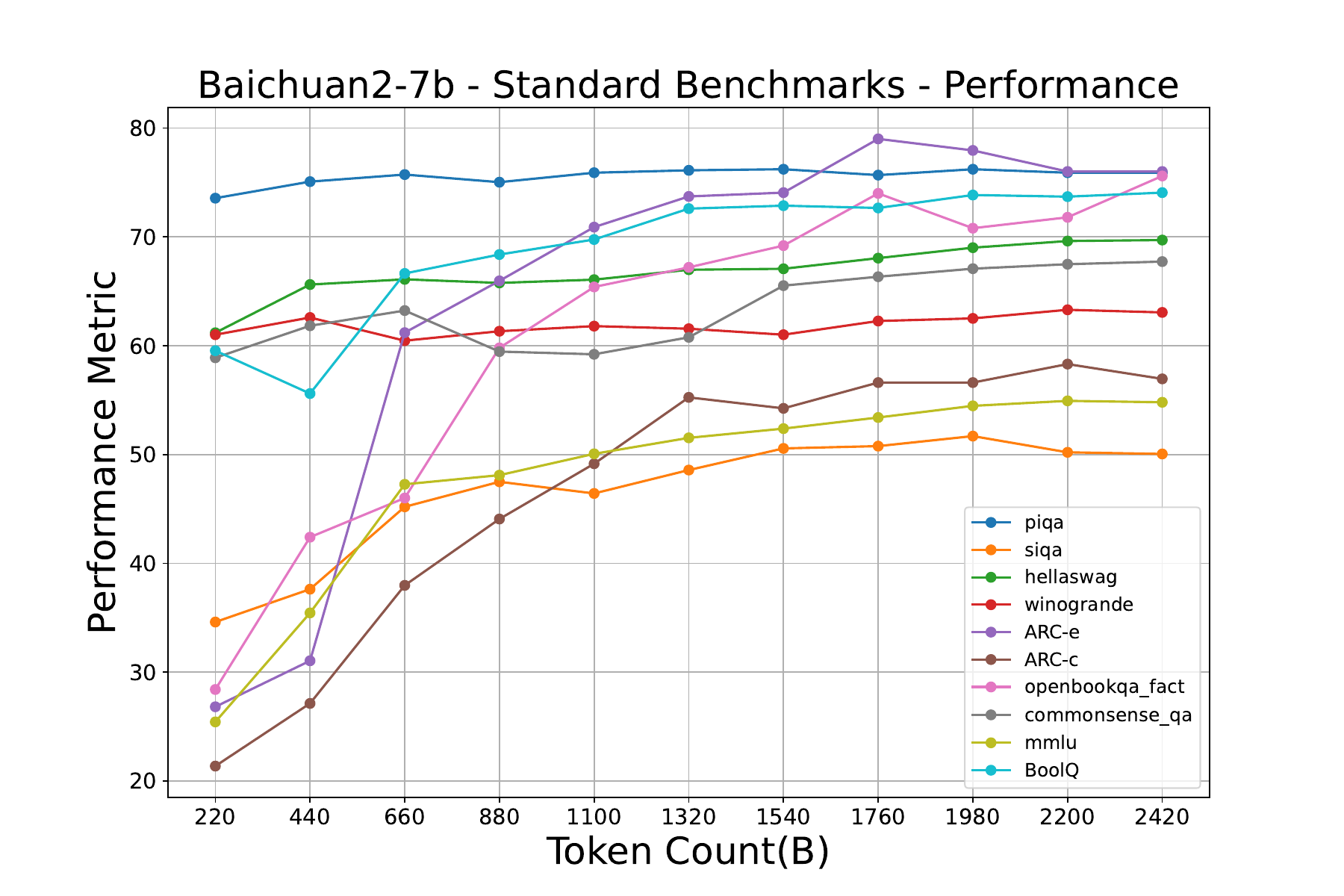}
        \caption{Baichuan-7B}
        \label{fig:Baichuan7b-standard-benchmarks}
    \end{subfigure}
    \hfill
    \begin{subfigure}[b]{0.3\textwidth}
        \includegraphics[width=\textwidth]{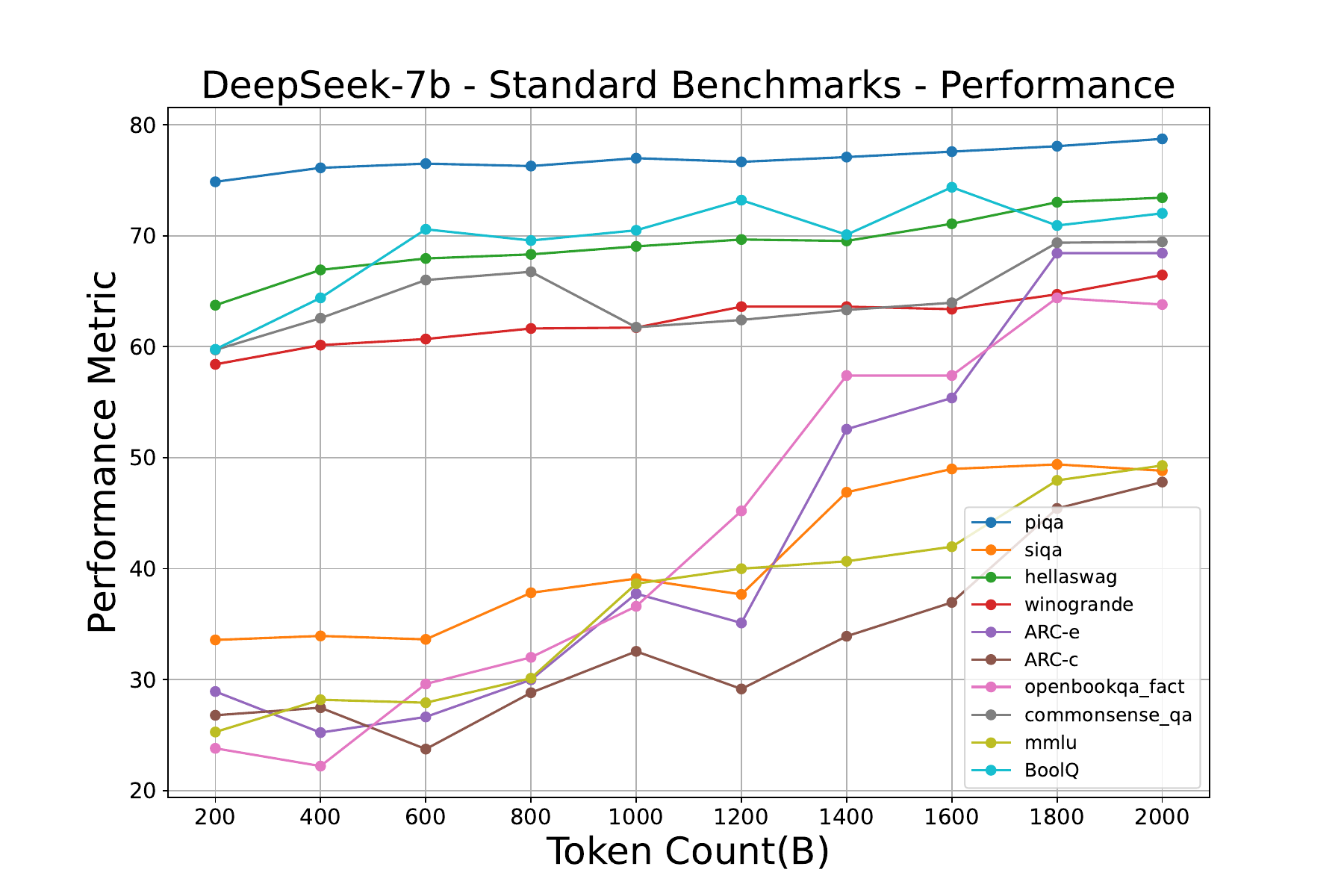}
        \caption{DeepSeek-7B}
        \label{fig:deepseek7b-Standard-Benchmarks}
    \end{subfigure}
    \hfill
    \begin{subfigure}[b]{0.3\textwidth}
        \includegraphics[width=\textwidth]{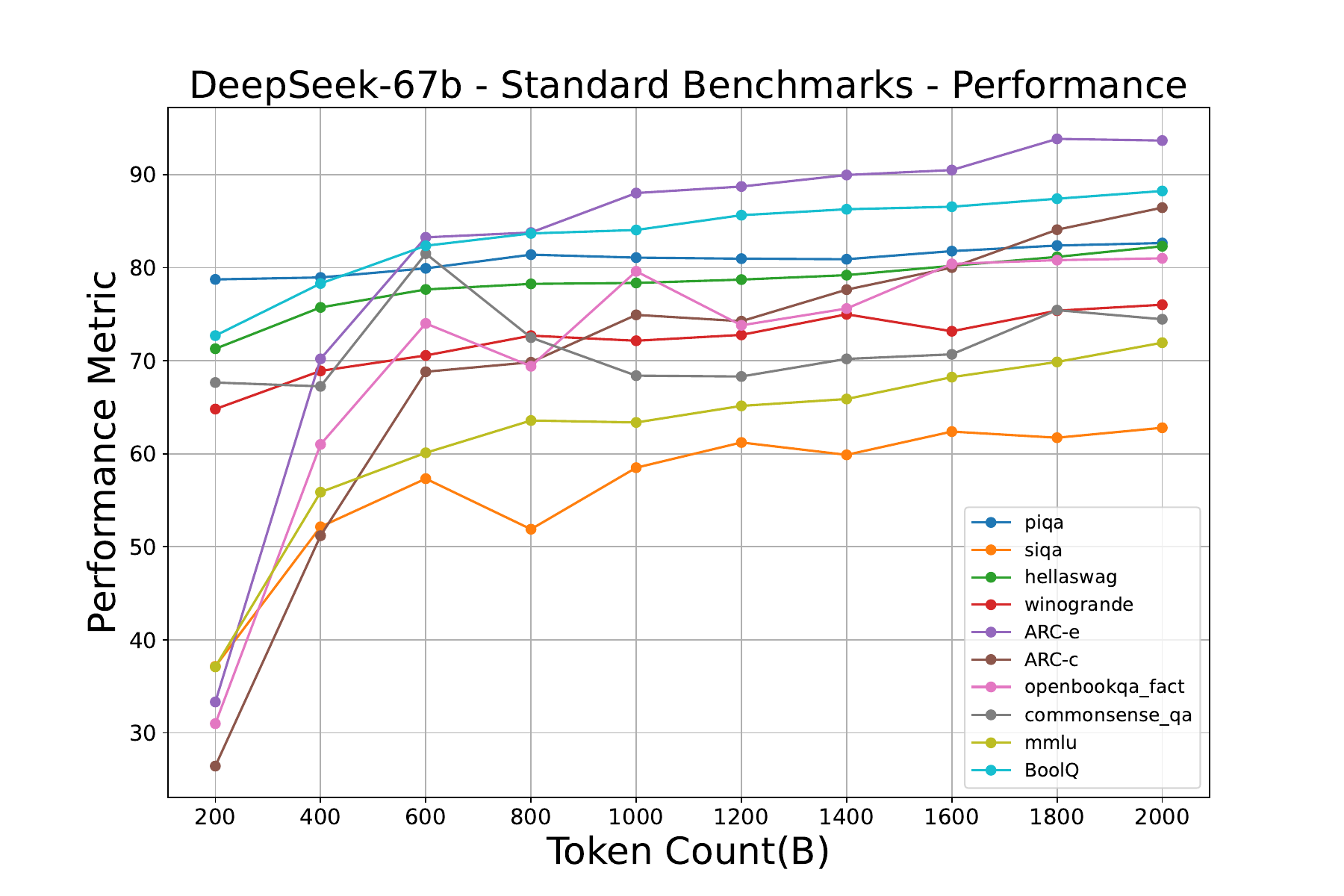}
        \caption{DeepSeek-67B}
        \label{fig:deepseek67b-Standard-Benchmarks}
    \end{subfigure}
    
    \begin{subfigure}[b]{0.3\textwidth}
        \includegraphics[width=\textwidth]{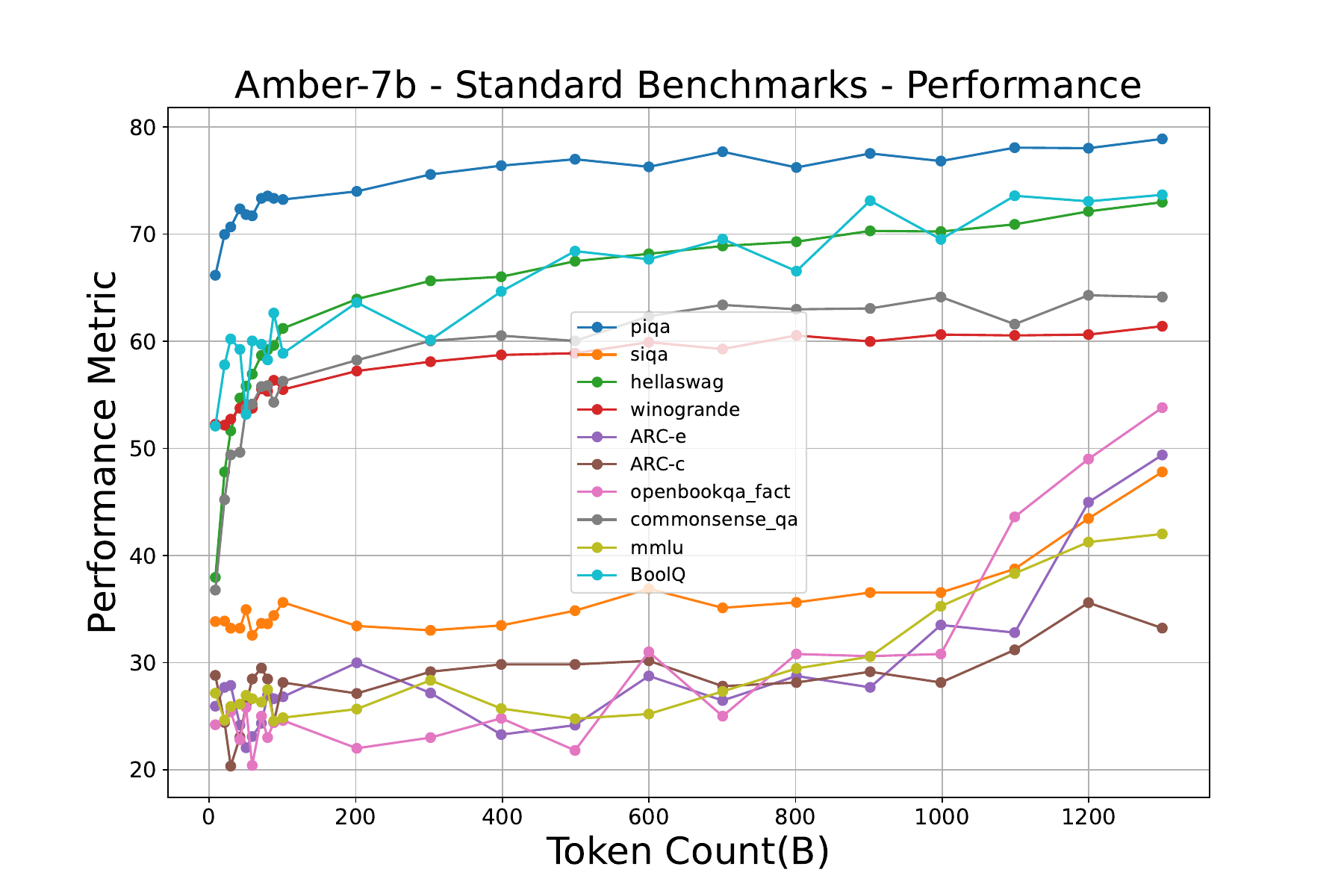}
        \caption{Amber-7B}
        \label{fig:amber7B-Standard-Benchmarks}
    \end{subfigure}
    \hfill
    \begin{subfigure}[b]{0.3\textwidth}
        \includegraphics[width=\textwidth]{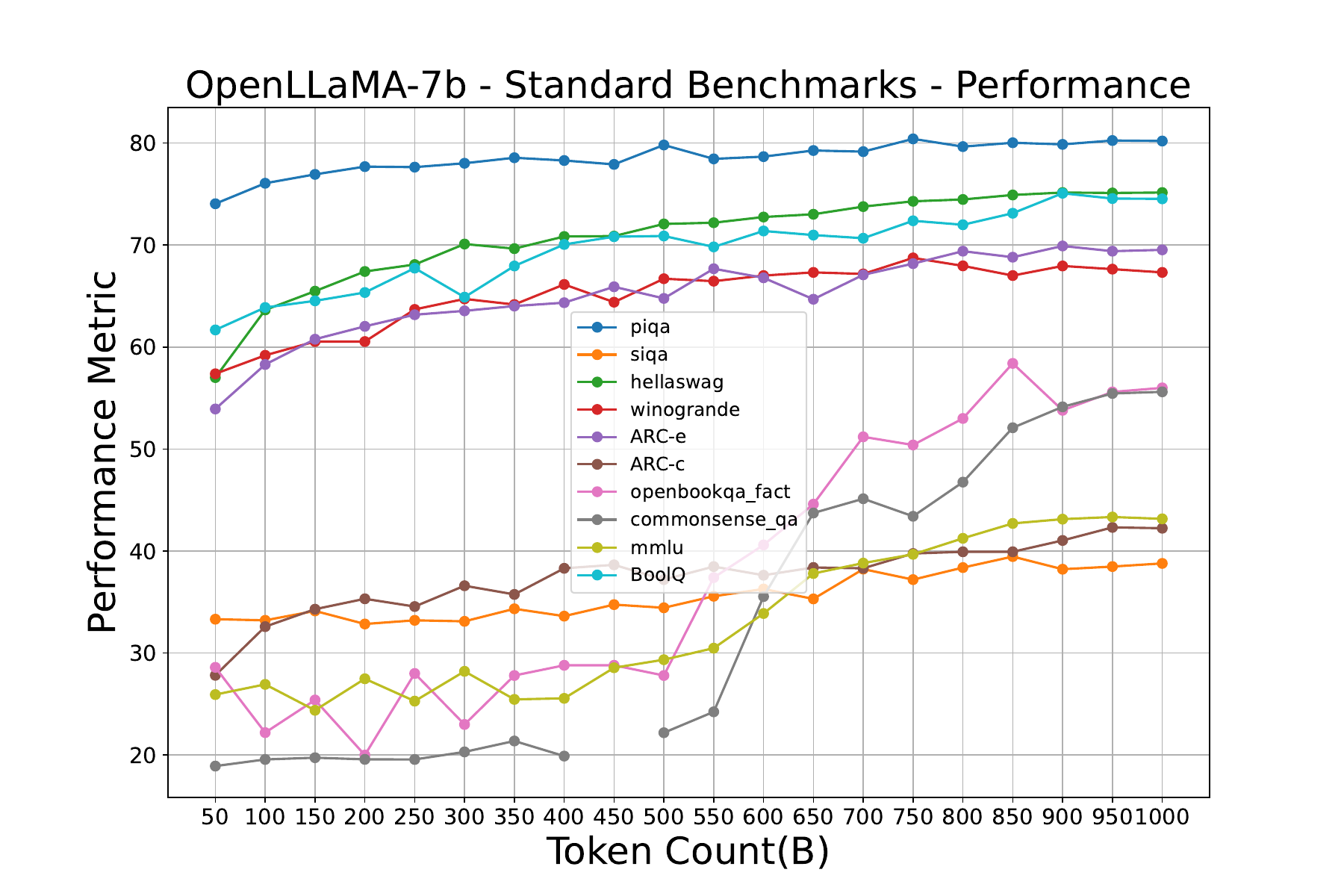}
        \caption{OpenLLaMA-7B}
        \label{fig:openllama7b-Standard-Benchmarks}
    \end{subfigure}
    \hfill
    \begin{subfigure}[b]{0.3\textwidth}
        \includegraphics[width=\textwidth]{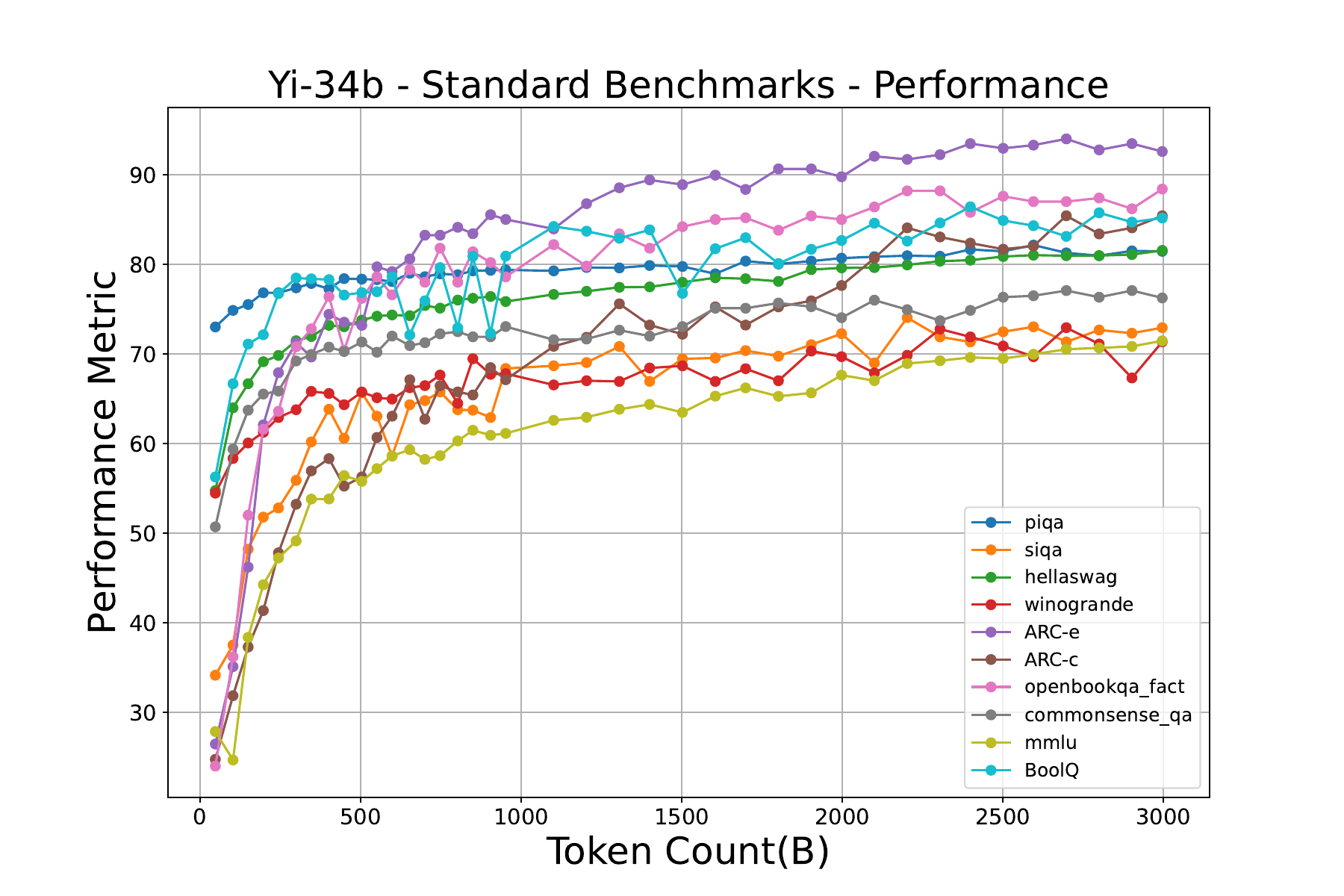}
        \caption{Yi-34B}
        \label{fig:yi34b-Standard-Benchmarks}
    \end{subfigure}
    
    \caption{Comparative performance analysis of Baichuan-7B, DeepSeek-7B, DeepSeek-67B, Amber-7B, OpenLLaMA-7B, and Yi-34B models across standard benchmarks.}
    \label{fig:models-comparison-standard-benckmarks}
\end{figure}

\paragraph{Analysis of standard benchmarks}
We define that standard benchmarks include PIQA, SIQA, HellaSwag, WinoGrande, ARC Easy and Challenge, OpenBookQA, CommonsenseQA, BoolQ, and MMLU. This categorization is predicated on the observation that these benchmarks, to varying degrees, evaluate the models' understanding of real-world knowledge, reading comprehension, and commonsense reasoning. The descriptions and observations on these datasets are as follows. 

First, PIQA, HellaSwag, BoolQ, WinoGrande, and CommonsenseQA are all datasets for evaluating models' common sense reasoning and understanding capabilities.  PIQA tests models on understanding physical laws, HellaSwag evaluates how well they predict story outcomes, BoolQ evaluates their comprehension and inference skills through yes/no questions based on text passages, WinoGrande examines their handling of language nuances and context-based ambiguity resolution, and CommonsenseQA assesses their general knowledge and reasoning. These datasets measure the most fundamental capabilities of the models. 
As shown in Figure \ref{fig:models-comparison-standard-benckmarks}, within each model, the accuracy on these datasets rapidly increases in the early stages of training (before 300B tokens) and gradually reaches a plateau. This trend is particularly evident in the checkpoints before 100B tokens for the Amber-7b model. 

Second, SIQA, ARC (both Easy and Challenge), OpenBookQA and MMLU target more advanced cognitive abilities compared to datasets like PIQA, HellaSwag, WinoGrande and CommonsenseQA, with accuracy improvements on these benchmarks typically occurring later in the training process. SIQA tests social comprehension, ARC spans basic to complex scientific reasoning, OpenBookQA requires factual integration with textual understanding and MMLU measures knowledge application across multiple disciplines. These datasets emerge as crucial in the mid-training phases, shifting from basic commonsense to intricate reasoning and domain-specific knowledge application. This progression underscores a layered approach in AI training, moving from foundational understanding to higher-order cognitive skills.

These observations underscore the importance of benchmark selection in model evaluation. The early plateau observed in Standard Benchmarks suggests a rapid acquisition of general knowledge and reasoning capabilities, followed by a phase where additional training yields diminishing returns on these particular tasks. Conversely, the mid-training spike in other benchmarks indicates areas where models may require more extensive training to fully capture and understand the complexities of the tasks.

\subsection{Cross-model analysis}
\label{cross-model analysis}

%

Our cross-model analysis aims to understand the nuances and intricacies of model performance across different architectures and scales. 

\subsubsection{Analysis within the 7b scale}
\label{analysis within the 7b scale}

\begin{figure}[ht]
    \centering
    \begin{subfigure}[b]{0.45\textwidth}
        \includegraphics[width=\textwidth]{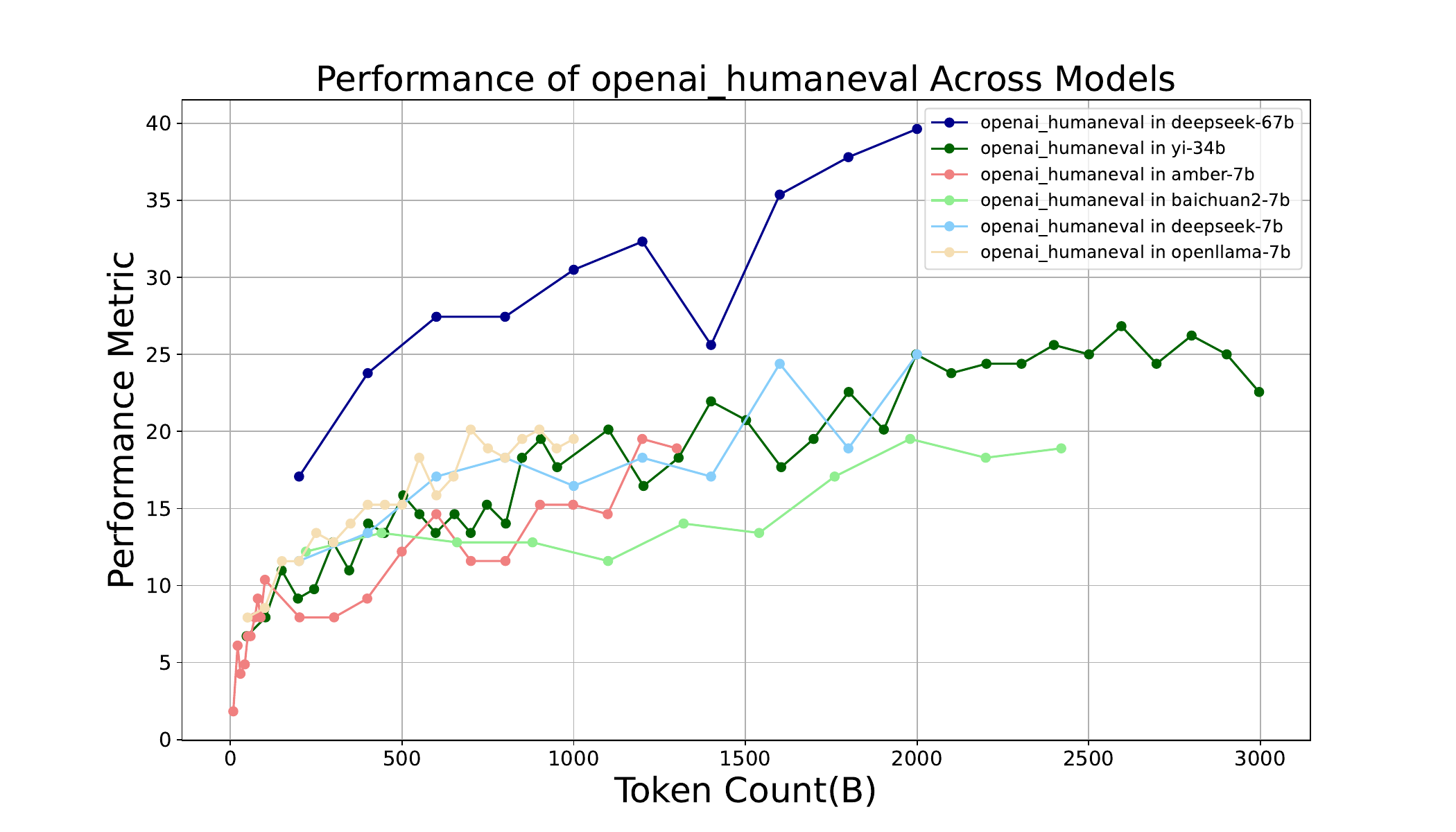}
        \caption{HumanEval performance}
        \label{fig:humaneval across 7b}
    \end{subfigure}
    \hfill
    \begin{subfigure}[b]{0.45\textwidth}
        \includegraphics[width=\textwidth]{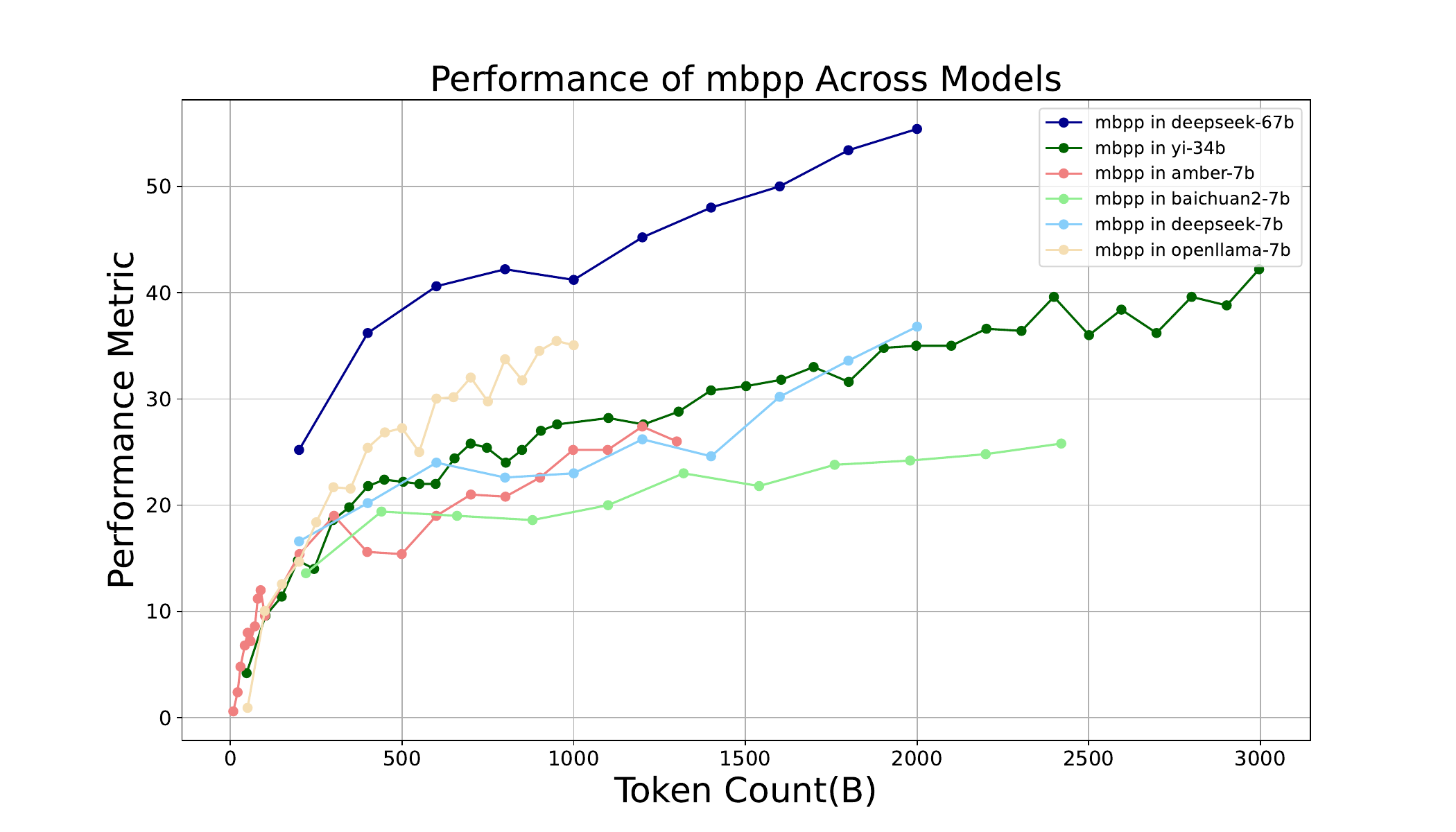}
        \caption{MBPP performance}
        \label{fig:mbpp across 7b}
    \end{subfigure}
    \caption{Code generation performance across Baichuan2-7b, DeepSeek-7b, OpenLLaMA-7b, and Amber-7b.}
    \label{fig:code generation across}
\end{figure}

As illustrated in Figure~\ref{fig:code generation across}, when analyzing the performance of HumanEval and MMBP in the code generation domain using 7b models, it becomes apparent that the initial performance is comparable across the various models. However, as the number of tokens increases, their performance curves diverge, showing different trends. This divergence is attributed to both the model architecture and differences in training strategies. Specifically for the Amber-7b model, there is a noticeable decline in capability in the 200b-300b token range, likely due to the training dataset.

\subsubsection{Analysis across scales}
\label{analysis across scales}

\begin{figure}[ht]
    \centering
    \begin{subfigure}[b]{0.45\textwidth}
        \includegraphics[width=\linewidth]{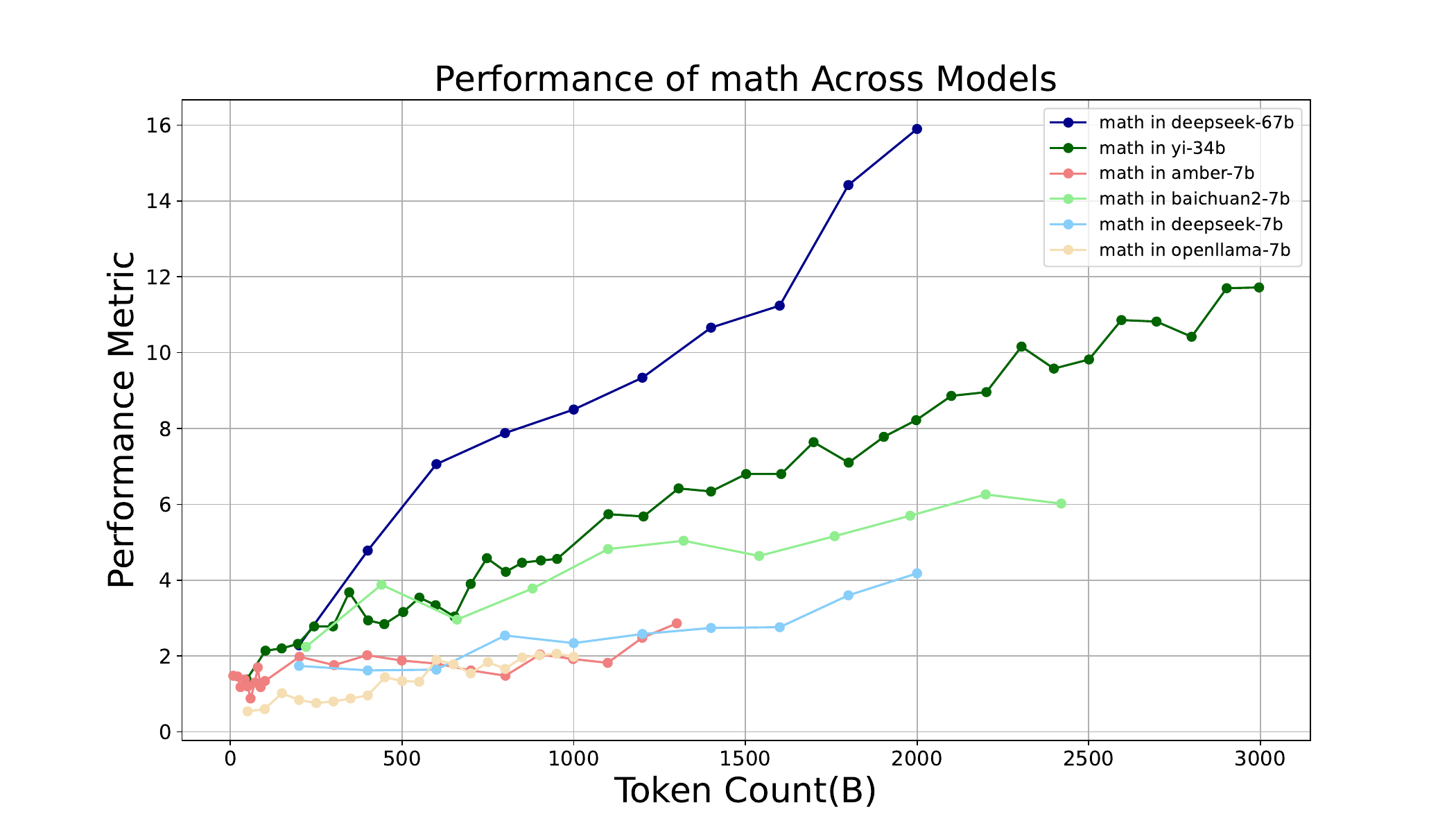}
        \caption{Math}
        \label{fig:math across scales}
    \end{subfigure}
    \hfill
    \begin{subfigure}[b]{0.45\textwidth}
        \includegraphics[width=\textwidth]{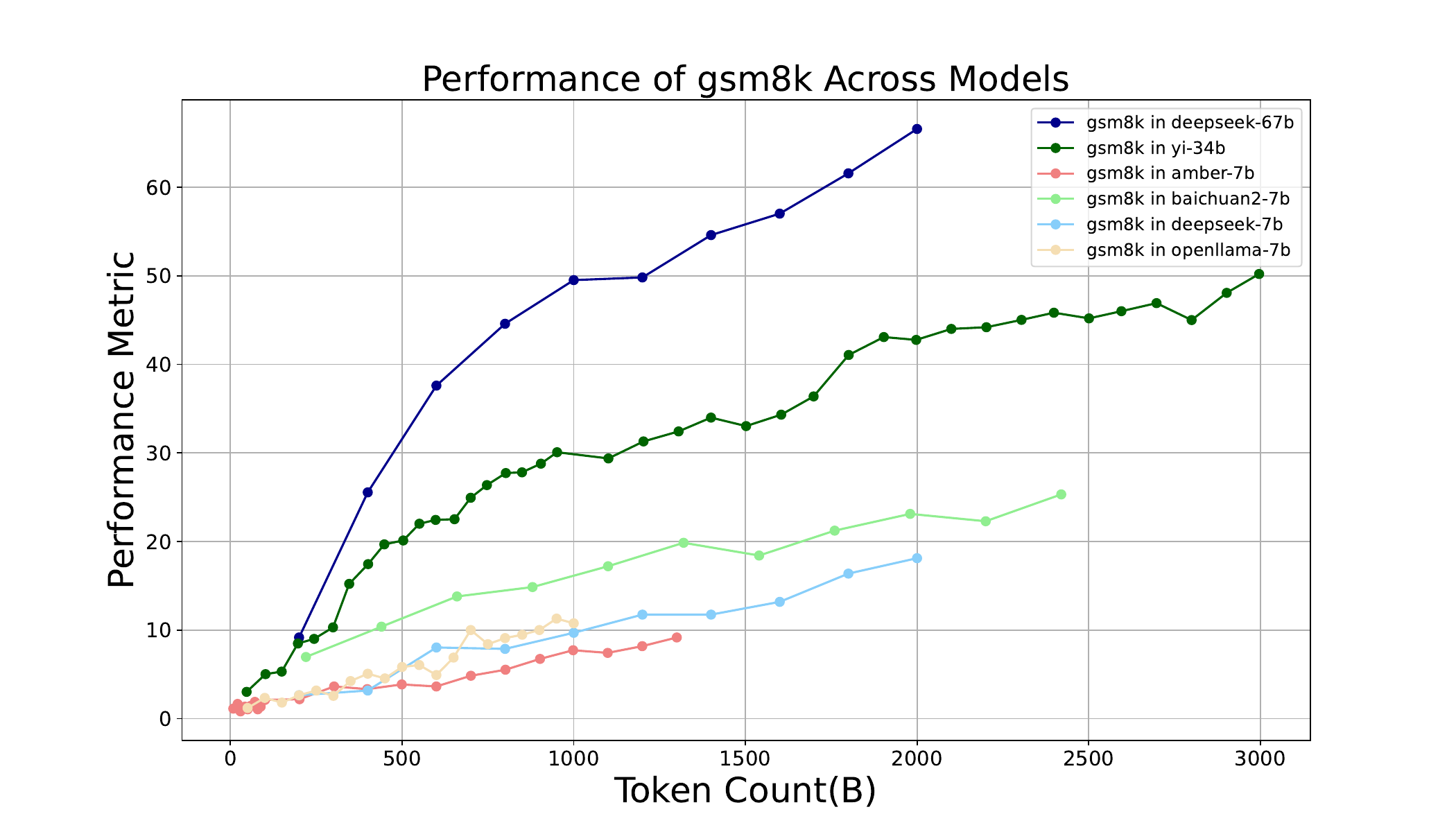}
        \caption{GSM8K}
        \label{fig:gsm8k across scales}
    \end{subfigure}
    \caption{MATH performance across 7b, 34b, 67b models.}
    \label{fig:math domain across scales}
\end{figure}

\begin{figure}[ht]
    \centering
    \begin{subfigure}[b]{0.32\textwidth}
        \includegraphics[width=\textwidth]{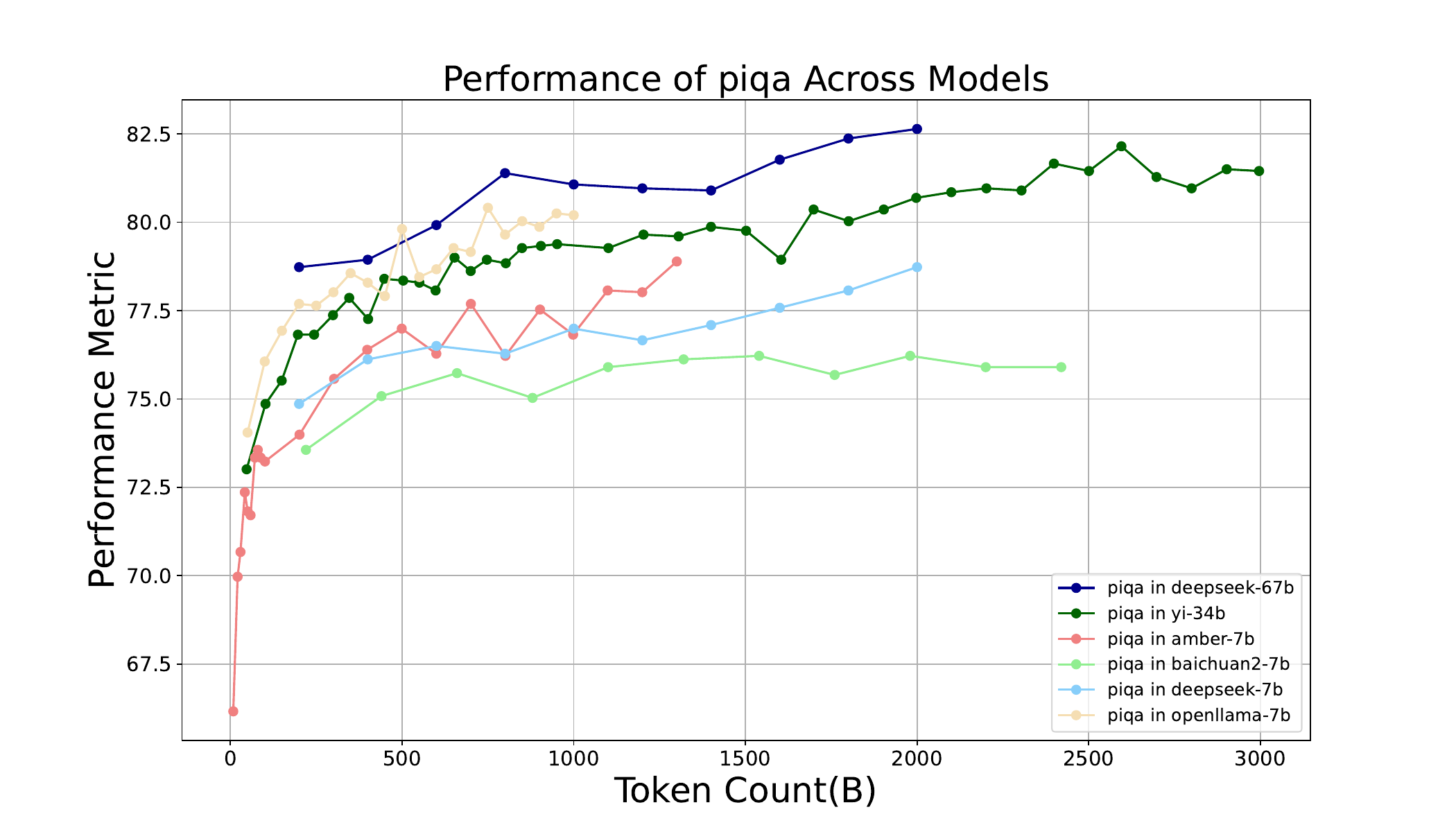}
        \caption{PIQA}
        \label{fig:piqa across scales}
    \end{subfigure}
    \hfill
    \begin{subfigure}[b]{0.32\textwidth}
        \includegraphics[width=\textwidth]{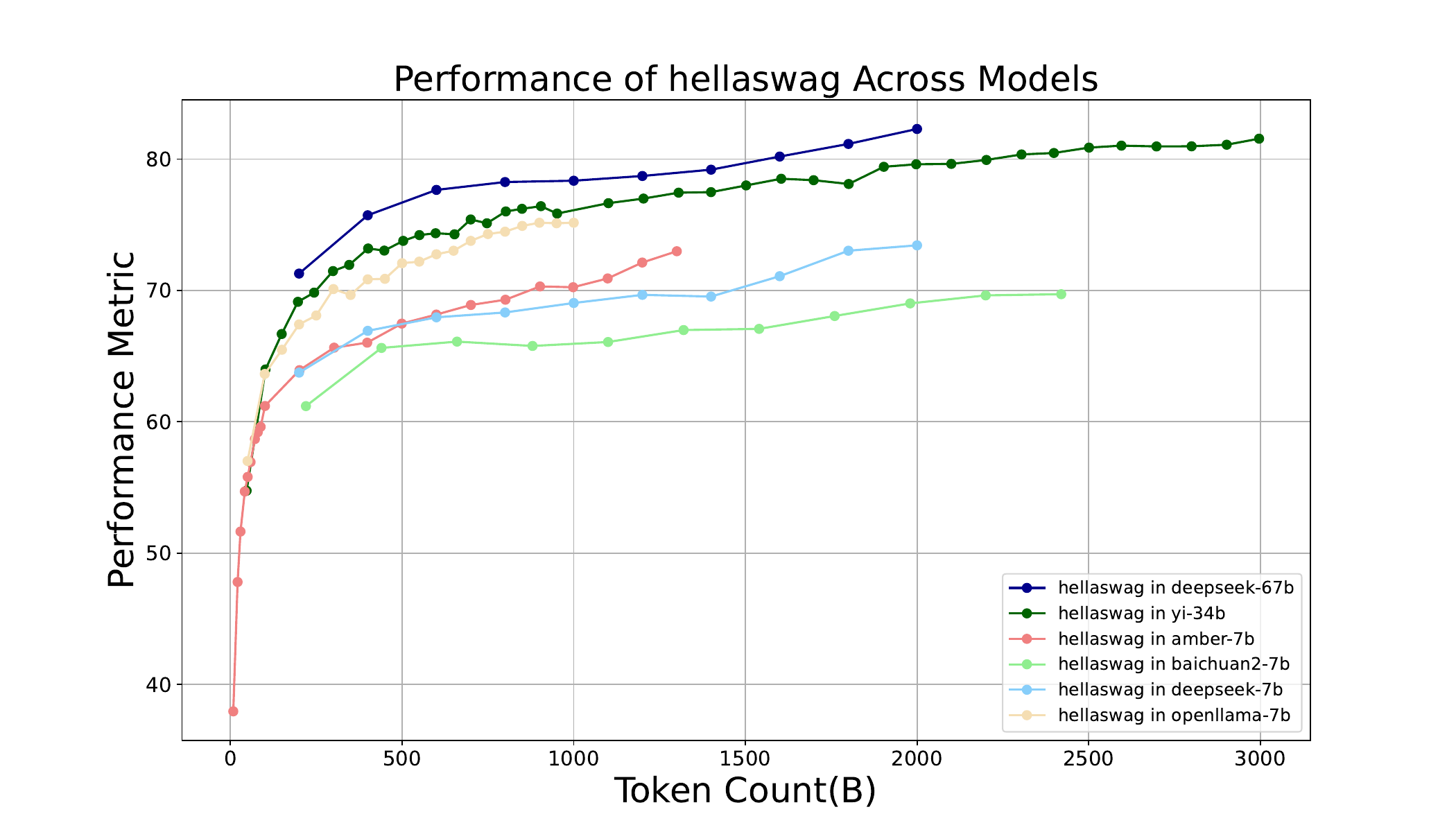}
        \caption{HellaSwag}
        \label{fig:hellaswag across scales}
    \end{subfigure}
    \hfill
    \begin{subfigure}[b]{0.32\textwidth}
        \includegraphics[width=\textwidth]{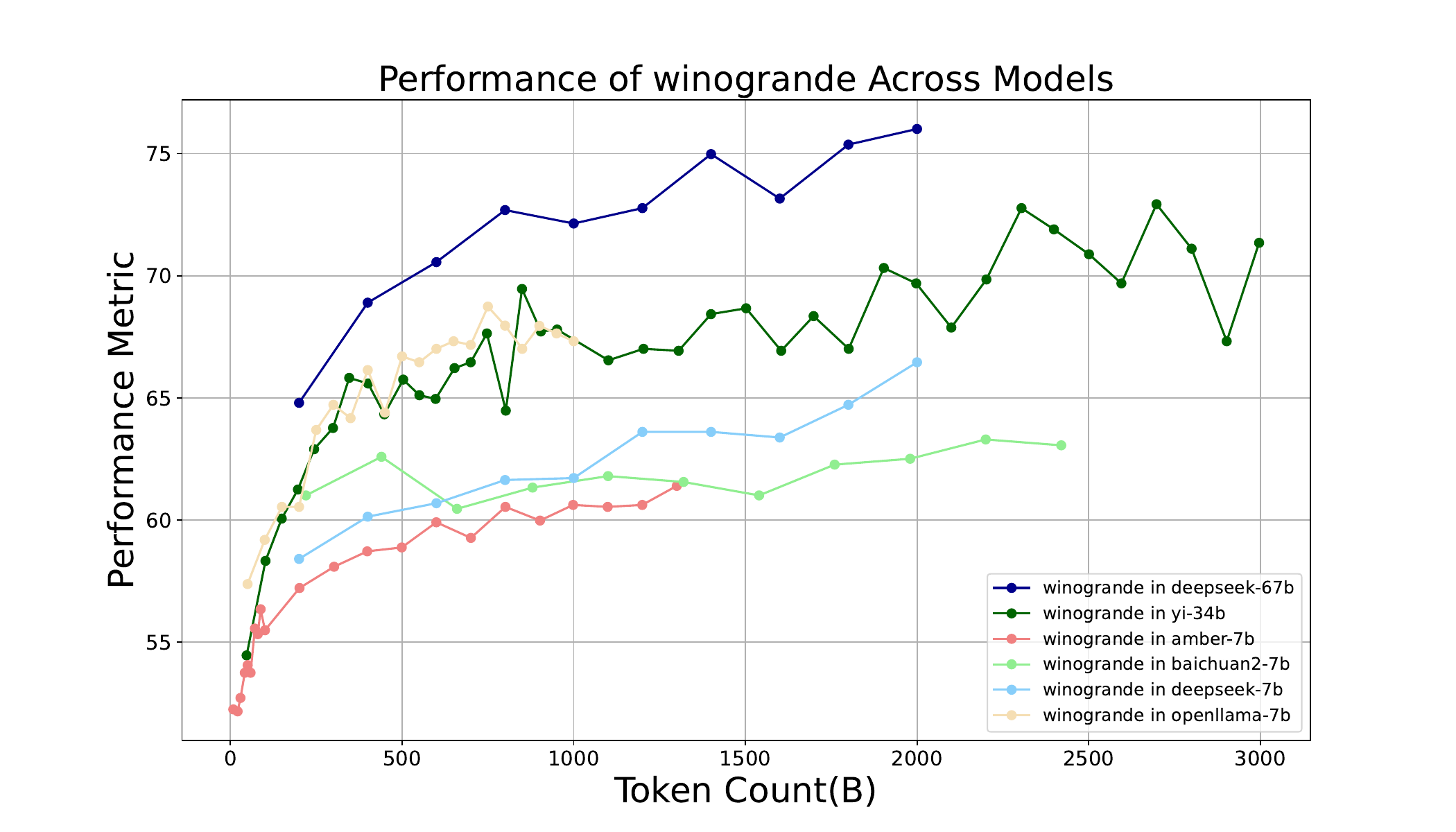}
        \caption{WinoGrande}
        \label{fig:winogrande across scales}
    \end{subfigure}
    \caption{Commonsense understanding performance across 7b, 34b, 67b models.}
    \label{fig:commensense across scales}
\end{figure}

The size and complexity of a model are key factors affecting its learning capabilities and performance in reasoning tasks, yet the application of specific techniques may allow smaller-scale models to match or even surpass the capabilities of larger models. As shown in Figure \ref{fig:math domain across scales} and Figure \ref{fig:commensense across scales}, by examining the performance on MATH, GSM8K, PIQA, HellaSwag, and WinoGrande across models of different sizes, it is observable that all models improve their abilities in tasks involving math, physical interaction understanding and commonsense reasoning in a relatively synchronized manner. Generally, the performance of 67b models surpasses that of 34b, which in turn exceeds that of 7b models at the same training phase in math-related datasets. However, exceptions like the OpenLLaMA-7b model approaching or even exceeding the capabilities of Yi-34b demonstrate that the model scale should be chosen based on the complexity of the task and the required reasoning capabilities. For tasks requiring extensive reasoning and deep understanding, larger models are more suitable; whereas for situations with limited resources or lower task complexity, smaller models may be a more economical choice.

\section{Scaling Law}
\label{sec: scaling law}
\subsection{Scaling law definition}

The Scaling Law posits a predictive enhancement in model performance through the augmentation of three critical dimensions: computational budget $C$, model size $N$, and data size $D$. This law articulates that when the model size $N$ is quantified by the count of model parameters, and the data size $D$ is measured in terms of the number of tokens, the computational budget $C$ can be approximated effectively by the formula $C = 6ND$. Consequently, an essential research focus within the ambit of the Scaling Law involves strategizing an optimal allocation between model size and data size during the amplification of the computational budget, with the ultimate goal of achieving the most efficacious model configuration.

Research into the optimal allocation strategies for model/data size enhancement \citep{hoffmann2022training, kaplan2020scaling} has led to varied conclusions, prompting skepticism about the universal applicability of the Scaling Law. Furthermore, structural differences among models, disparities in the quality of training data, and variations in hyperparameter configurations are often primary factors that contribute to the discrepancies observed in the Scale Law. Baichuan2 has conducted experiments employing the Scale Law as proposed by \citet{henighan2020scaling}. Meanwhile, the two models of Deepseek have adopted the IsoFLOP profile approach from chinchilla, and have implemented certain degrees of optimization to it.


In order to extrapolate more universally applicable insights, our investigation is predicated on the foundational principles of the chinchilla law. The formal mathematical representation of this law is encapsulated in the following expression:
\begin{equation}
  P = f(N, D, C) = a \cdot N^{\alpha} \cdot D^{\beta} \cdot C^{\gamma},
  \label{sclaing_law_equation}
\end{equation}
where $P$ denotes the performance metric of the model. Conventionally, in the ambit of prior scholarly inquiries, this metric is frequently quantified in terms of the model's loss, symbolized by $L$. And $a, \alpha, \beta, \gamma $ are constants that depend on the specific architecture and task. We explain the specific meaning of each parameter in detail in Appendix~\ref{app:explanation_scaling_law}.

\subsection{Scaling law evaluation}
\begin{figure}[ht]
    \centering
    \begin{subfigure}[b]{0.32\textwidth}
        \includegraphics[width=\textwidth]{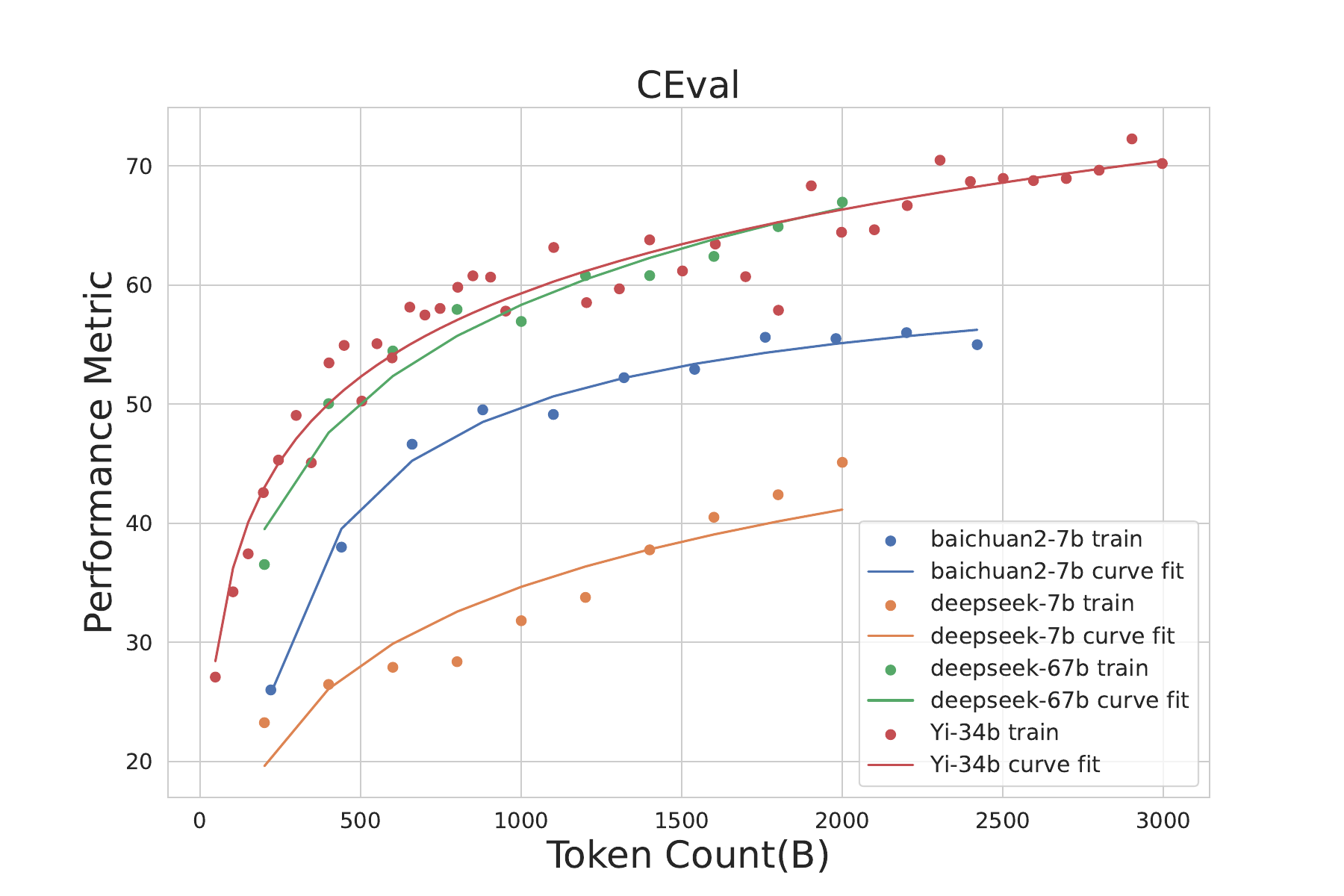}
        \caption{CEval Score Trends}
        \label{fig:scaling-law-ceval}
    \end{subfigure}
    \hfill
    \begin{subfigure}[b]{0.32\textwidth}
        \includegraphics[width=\textwidth]{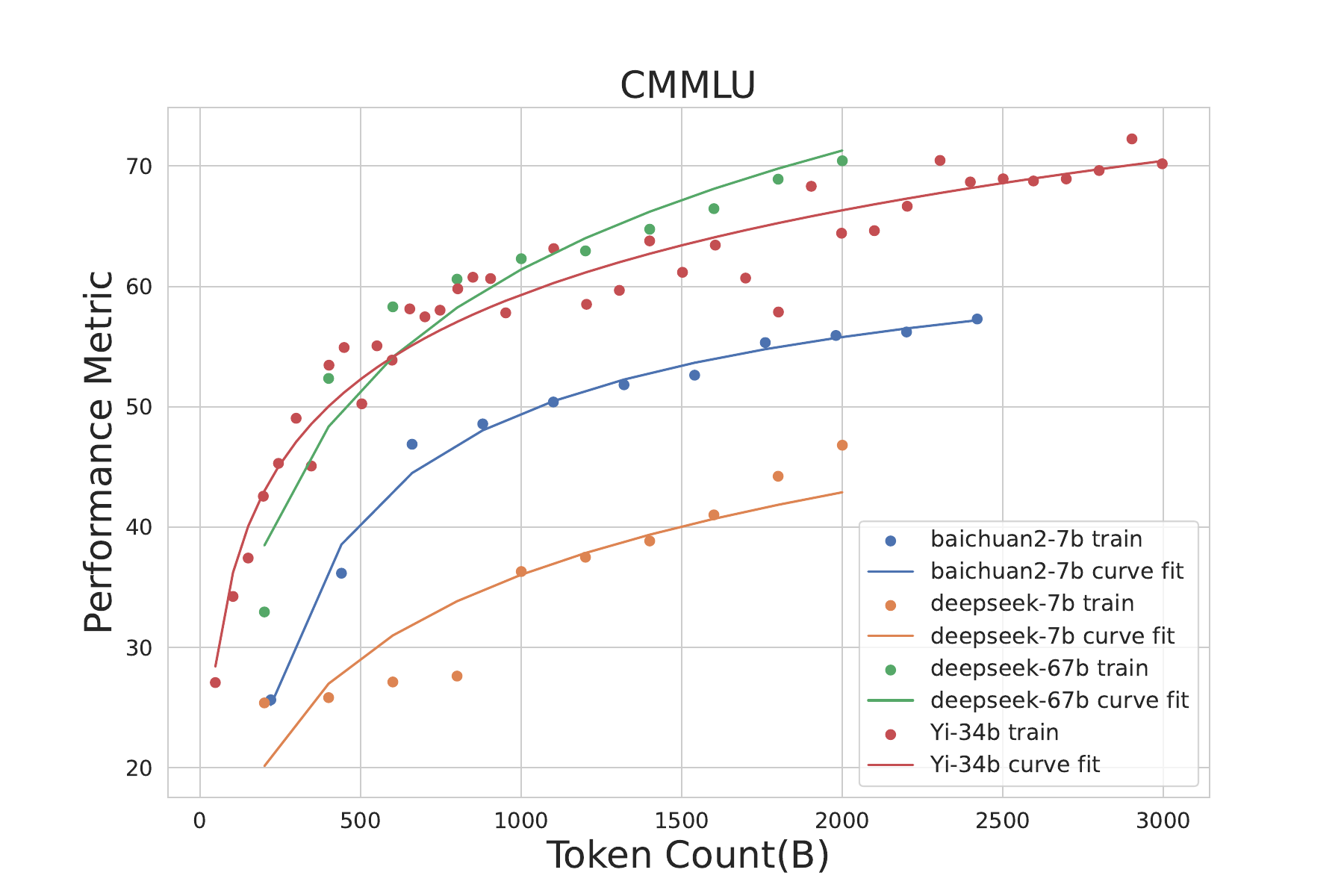}
        \caption{CMMLU Score Trends}
        \label{fig:scaling-law-cmmlu}
    \end{subfigure}
    \hfill
    \begin{subfigure}[b]{0.32\textwidth}
        \includegraphics[width=\textwidth]{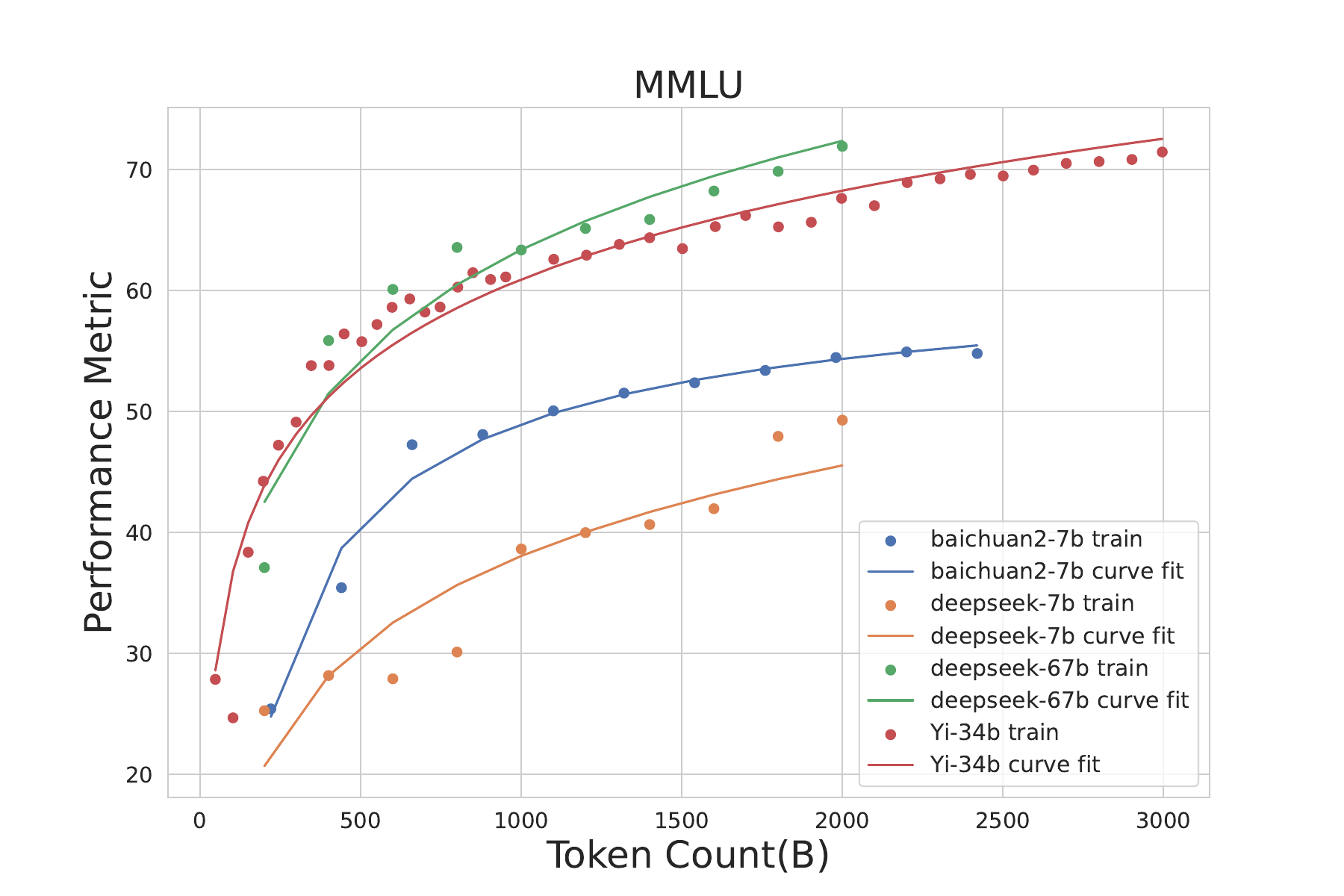}
        \caption{MMLU Score Trends}
        \label{fig:scaling-law-mmlu}
    \end{subfigure}
    \caption{Comparative performance trends of different language models over various training checkpoints, as fitted by the Eq.(\ref{sclaing_law_equation}). (a) We present the CEval Score Trends, showing a consistent increase in performance with larger training data sizes. (b) We depict the CMMLU Score Trends, with a similar positive correlation between score and training size.  (c) We illustrate the MMLU Score Trends, with all models demonstrating performance enhancements correlating with the scale of training data. Each graph plots individual data points for different versions of the models and overlays a curve fit to represent the scaling law's predictive performance scaling.}
    \label{fig:scaling-law}
\end{figure}

Figure~\ref{fig:scaling-law} indicates a clear relationship between the size of the training data and the performance scores across the CEval, CMMLU, and MMLU benchmarks. 
All three subfigures exhibit a monotonically increasing trend, where the scores improve as the number of training checkpoints increases. This suggests that larger training datasets enable the models to learn more effectively, leading to better performance on evaluation tasks.
The rate of score increase appears to be diminishing as the number of checkpoints grows. Initially, there is a steep improvement in scores, which gradually levels off. This could indicate that while larger training datasets initially provide substantial new information for the models to learn from, the incremental benefit decreases as the model approaches an asymptotic performance ceiling.

The scaling law's curve fits seem to be closely aligned with the actual data points in all three graphs. This indicates that Eq.(\ref{sclaing_law_equation}) effectively models the relationship between training data size and model performance in these cases.

Observing the provided graphs, it can be noted that the scaling law fits the performance data of the Baichuan2 model more accurately compared to the two DeepSeek models. This is evident from the closer alignment of the Baichuan2 data points to its trend line across all three subfigures (CEval, CMMLU, and MMLU score trends), suggesting that Eq.(\ref{sclaing_law_equation}) may be more predictive for the training and performance characteristics of the Baichuan2 model.

The deviation of the DeepSeek models' performance from the predictions of Eq.(\ref{sclaing_law_equation}) could indeed stem from the adoption of a new model scale representation. By utilizing non-embedding FLOPs per token $M$ instead of the overall model parameters $N$, the scaling law that was originally designed or calibrated for the parameter might not accurately capture the performance nuances associated with computational complexity per token.

The effectiveness of the fit may also differ between the early and late stages of training. For example, if the curve fits well initially but diverges as training progresses, it might suggest that Eq.(\ref{sclaing_law_equation}) more accurately predicts performance improvements at earlier stages of training.

Some models (like Baichuan2-7B and DeepSeek-67B) might show a better fit at later stages, possibly due to a more stable learning process as the model converges towards its peak performance.

The Yi-34B model seems to exhibit a superior fit to the Scaling Law. This enhanced alignment with the Scaling Law suggests that Yi-34B may have architectural or algorithmic advantages that enable it to leverage data more effectively. Despite Yi-34B having fewer parameters than Deepseek-67B, it demonstrates that with an increase in data volume, Yi-34B can achieve comparable results post-training. This observation suggests that the quantity of model parameters is not the sole determinant of performance enhancement when sufficient data is available. In expanding the training datasets, Yi-34B appears to utilize its parameters more efficiently, potentially indicating a better generalization capacity or optimization that allows for superior learning from additional data. This could encourage the development of more data-centric approaches in the design of algorithms, focusing on how to better utilize the information available to enhance learning outcomes, even with a comparatively smaller set of parameters.

Upon analyzing the graphs, it is evident that while the trend of increasing performance with larger datasets is present, the actual scores for each model at various training checkpoints do not precisely align with the expected trajectory of the scaling law. This discrepancy may stem from the fact that prior research often used loss as a measure of model performance. In contrast, our experiments employ specific downstream task performance metrics. Given that scoring methods vary and dataset distributions differ, there will be a certain degree of divergence from the original scaling law.

\section{Conclusion}

This investigation into large language models (LLMs) enhances our understanding of model training complexities and scaling laws. Analyzing model dynamics across various tasks and training stages, we have gleaned insights to improve training and optimization strategies.

Our study reveals the predictive power of task dynamics within a domain for unseen tasks, suggesting the benefit of adaptable training protocols. Drawing parallels between AI learning and human cognition, we see potential in applying cross-domain insights for better training outcomes.

Key findings highlight the influence of training strategies, dataset quality, and architecture on LLM efficiency and effectiveness. While model size impacts learning outcomes, innovative methods enable smaller models to rival larger ones.

In summary, our analysis of scaling laws, particularly through the scaling law, offers insights into training data size and model performance. We note improved performance with larger datasets, yet with diminishing returns. This indicates the importance of dataset expansion, alongside architectural and computational optimization, for maximizing data utility. Variations in scaling law efficacy across models suggest scaling behaviors are nuanced, necessitating tailored model development approaches. These insights hint at a future where refined scaling laws and optimization techniques significantly boost LLM capabilities, marking significant progress in AI.

\bibliography{colm2024_conference}
\bibliographystyle{colm2024_conference}

\appendix
\section{Model Description, Dataset Performances, and Detailed Experiment Prompts}
\subsection{Selected models and settings}
\label{a. models introduction}
\begin{itemize}
    \item \textbf{Baichuan-7B}: The Baichuan2-7B is an advanced large language model developed by the Baichuan AI team. This model is an enhanced version of the previous Baichuan models, featuring 7 billion parameters and optimized for improved performance in various natural language processing tasks. Baichuan2-7B is designed to handle complex language understanding and generation tasks with greater accuracy and efficiency, building upon the foundation of its predecessor while incorporating new architectural improvements and training techniques. 
    
    \item \textbf{Deepseek-7B}: The Deepseek-7B model is a large language model that is part of the Deepseek series, focusing on language understanding and generation tasks. It features a self-regressive Transformer decoder architecture and utilizes technologies like multi-head attention (MHA) and grouped query attention (GQA) to enhance performance and efficiency. With two versions available (70 billion and 670 billion parameters), Deepseek-7B is designed to cater to different application scenarios and performance needs.
    
    \item \textbf{Deepseek-67B}: The Deepseek-67B is a more powerful version within the DeepSeek series, boasting 670 billion parameters. It also adopts the self-regressive Transformer decoder architecture and benefits from the MHA and GQA technologies, similar to the Deepseek-7B, but with a larger parameter scale for enhanced capabilities.
    
    \item \textbf{Amber-7B}: 
Amber-7B represents a pioneering open-source model characterized by complete data transparency. It stands out as the first model to open-source all its training data and code, ensuring bitwise reproducibility. This initiative aims to foster collaboration among researchers by providing a fully transparent framework for model development and replication.

    \item \textbf{OpenLLaMA-7B} :
    The OpenLLaMA-7B model \citep{openlm2023openllama} exemplifies an open-source effort to replicate the Llama2-7B architecture, mirroring its exact structure while utilizing open-source datasets for training. To replicate the intermediate checkpoints of this model, we employed the Megatron-LM framework in conjunction with the Redpajama dataset \citep{together2023redpajama}, applying falcon data filtering strategies to curate a training corpus of 1 trillion tokens. This approach underscores our commitment to transparency and accessibility in the development of large-scale language models.

    \item \textbf{Yi-34B}: The Yi-34B model, introduced by 01.AI, is a part of the Yi model family that demonstrates strong multi-dimensional capabilities. Based on 6B and 34B pretrained language models, Yi-34B extends to chat models, long context models, depth-upscaled models, and vision-language models. The base models achieve strong performance on benchmarks like MMLU, and the finetuned chat models deliver strong human preference rates on platforms like AlpacaEval and Chatbot Arena. With a focus on data quality and engineering efforts, Yi-34B is designed to be a next-generation computational platform, offering capabilities close to GPT-3.5.
\end{itemize}

\subsubsection{Amber-7b settings}
We replicated the Amber model utilizing the Llama2-7B configuration. The tokenized data from Amber, originally processed with a sequence length of 2048, was further detokenized and re-encoded to a sequence length of 4096 as per the preprocessing strategy of Megatron-LM \citep{shoeybi2020megatronlm}. In aligning with the Llama2 model's architecture and hyperparameters, we adopted a sequence length of 4096 and a batch size of 4 million tokens. In order to increase the training efficiency, we incorporate GQA technologies, utilizing a configuration with 32 attention heads and a group size of 4.

\subsubsection{OpenLLaMA-7b settings}
The OpenLLaMA 7Bv2 model, trained on a composite dataset comprising Falcon refined-web and starcoder datasets, augmented by contributions from Wikipedia, arXiv, books, and Stack Exchange data curated by RedPajama. The training utilizes a max learning rate of 3e-4 and min learning rate of 3e-5 with a batch size of 4 million tokens. For the learning rate schedular, this configuration closely aligns with the setup employed in Llama2.
\subsection{Complete results on benchmark datasets}
\label{a. full models performance}

\begin{figure}[H]
    \centering
    \begin{subfigure}[b]{0.3\textwidth}
        \includegraphics[width=\textwidth]{Exam_Charts/DeepSeek-67b_-_MMLU_Details_-_Performance.pdf}
        \caption{DeepSeek-67B MMLU}
        \label{fig:a-deepseek67b-mmlu}
    \end{subfigure}
    \hfill
    \begin{subfigure}[b]{0.3\textwidth}
        \includegraphics[width=\textwidth]{Exam_Charts/DeepSeek-67b_-_CEVAL_Details_-_Performance.pdf}
        \caption{DeepSeek-67B CEval}
        \label{fig:a-deepseek67b-ceval}
    \end{subfigure}
    \hfill
    \begin{subfigure}[b]{0.3\textwidth}
        \includegraphics[width=\textwidth]{Exam_Charts/DeepSeek-67b_-_CMMLU_Details_-_Performance.pdf}
        \caption{DeepSeek-67B CMMLU}
        \label{fig:a-deepseek67b-cmmlu}
    \end{subfigure}
    
    \begin{subfigure}[b]{0.3\textwidth}
        \includegraphics[width=\textwidth]{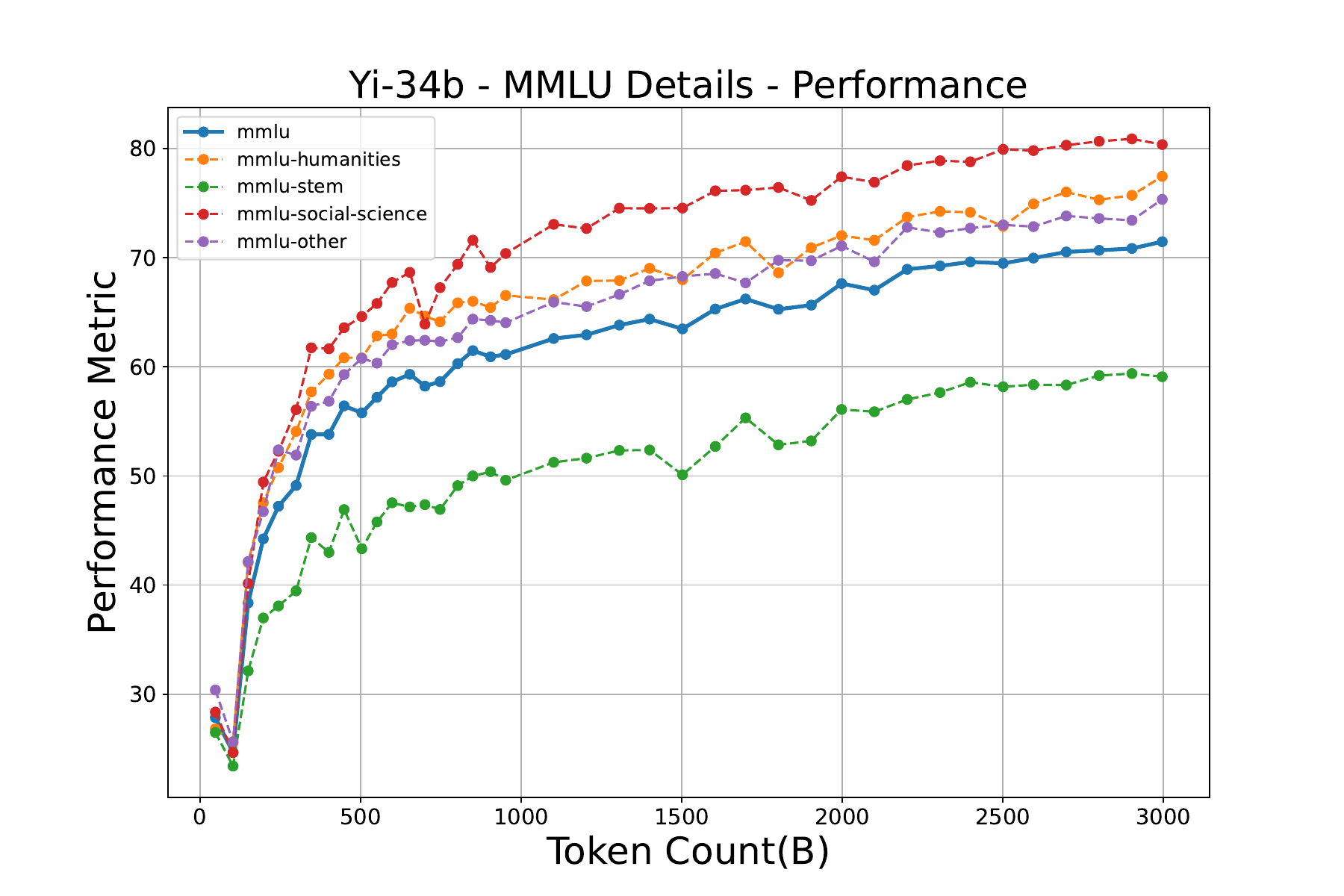}
        \caption{Yi-34B MMLU}
        \label{fig:a-yi34b-mmlu}
    \end{subfigure}
    \hfill
    \begin{subfigure}[b]{0.3\textwidth}
        \includegraphics[width=\textwidth]{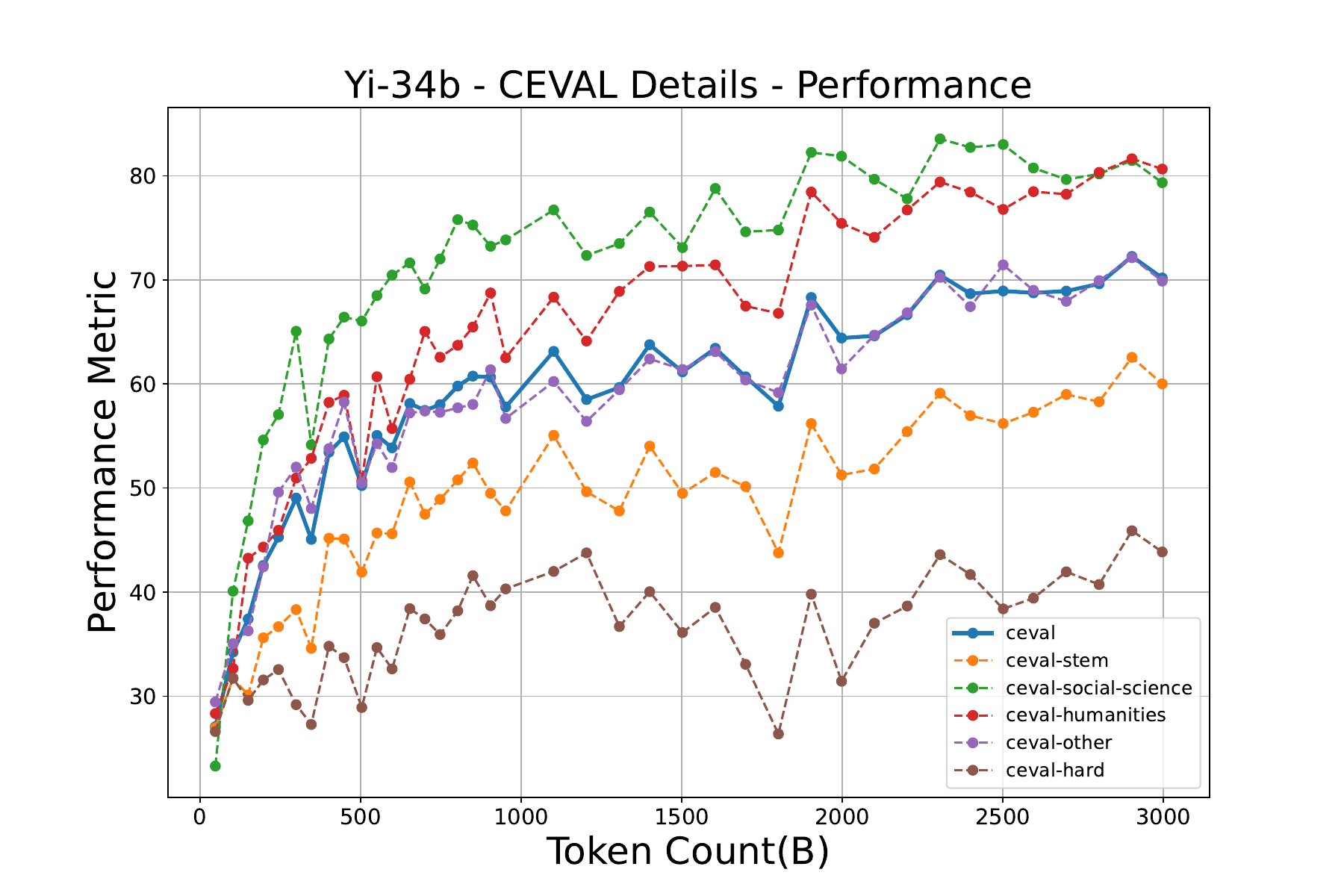}
        \caption{Yi-34B CEval}
        \label{fig:a-yi34b-ceval}
    \end{subfigure}
    \hfill
    \begin{subfigure}[b]{0.3\textwidth}
        \includegraphics[width=\textwidth]{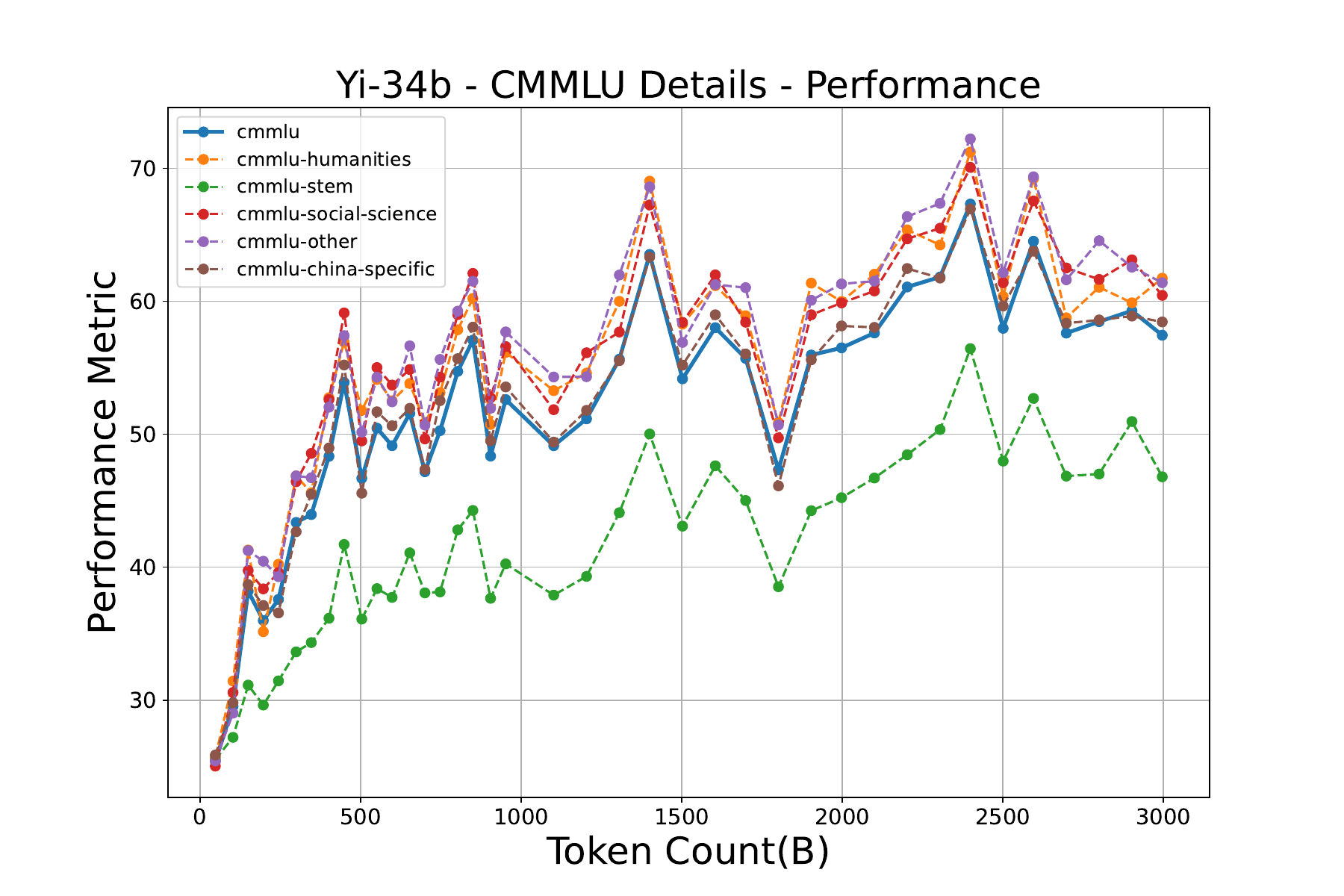}
        \caption{Yi-34B CMMLU}
        \label{fig:a-yi34b-cmmlu}
    \end{subfigure}
    
    \begin{subfigure}[b]{0.3\textwidth}
        \includegraphics[width=\textwidth]{Exam_Charts/Baichuan2-7b_-_MMLU_Details_-_Performance.pdf}
        \caption{Baichuan-7B MMLU}
        \label{fig:a-baichuan7b-mmlu}
    \end{subfigure}
    \hfill
    \begin{subfigure}[b]{0.3\textwidth}
        \includegraphics[width=\textwidth]{Exam_Charts/Baichuan2-7b_-_CEVAL_Details_-_Performance.pdf}
        \caption{Baichuan-7B CEval}
        \label{fig:a-baichuan7b-ceval}
    \end{subfigure}
    \hfill
    \begin{subfigure}[b]{0.3\textwidth}
        \includegraphics[width=\textwidth]{Exam_Charts/Baichuan2-7b_-_CMMLU_Details_-_Performance.pdf}
        \caption{Baichuan-7B CMMLU}
        \label{fig:a-baichuan7b-cmmlu}
    \end{subfigure}
    
    \begin{subfigure}[b]{0.3\textwidth}
        \includegraphics[width=\textwidth]{Exam_Charts/DeepSeek-7b_-_MMLU_Details_-_Performance.pdf}
        \caption{DeepSeek-7B MMLU}
        \label{fig:a-deepseek7b-mmlu}
    \end{subfigure}
    \hfill
    \begin{subfigure}[b]{0.3\textwidth}
        \includegraphics[width=\textwidth]{Exam_Charts/DeepSeek-7b_-_CEVAL_Details_-_Performance.pdf}
        \caption{DeepSeek-7B CEval}
        \label{fig:a-deepseek7b-ceval}
    \end{subfigure}
    \hfill
    \begin{subfigure}[b]{0.3\textwidth}
        \includegraphics[width=\textwidth]{Exam_Charts/DeepSeek-7b_-_CMMLU_Details_-_Performance.pdf}
        \caption{DeepSeek-7B CMMLU}
        \label{fig:a-deepseek7b-cmmlu}
    \end{subfigure}
    
    \begin{subfigure}[b]{0.3\textwidth}
        \includegraphics[width=\textwidth]{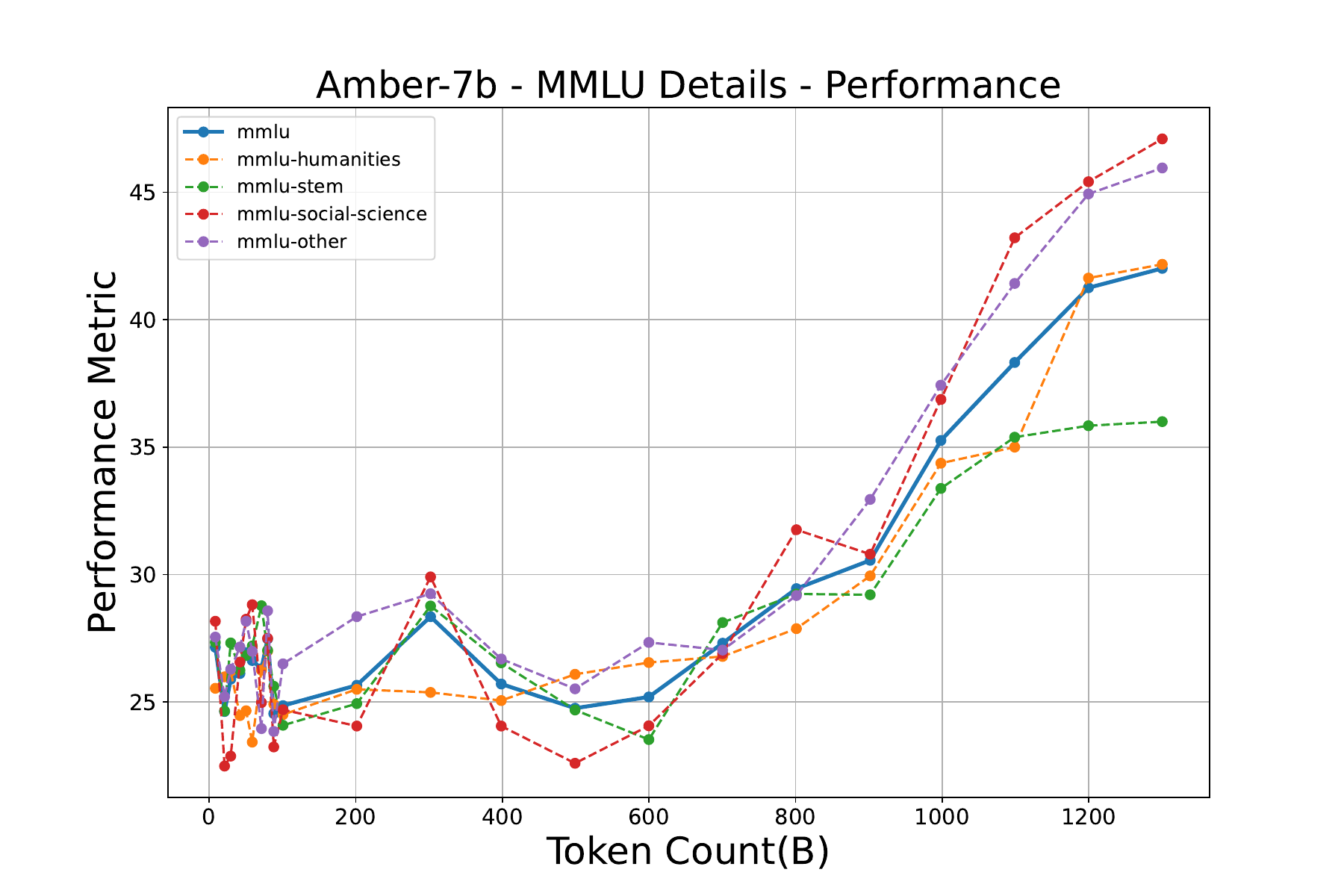}
        \caption{Amber-7B MMLU}
        \label{fig:a-amber7b-mmlu}
    \end{subfigure}
    \hfill
    \begin{subfigure}[b]{0.3\textwidth}
        \includegraphics[width=\textwidth]{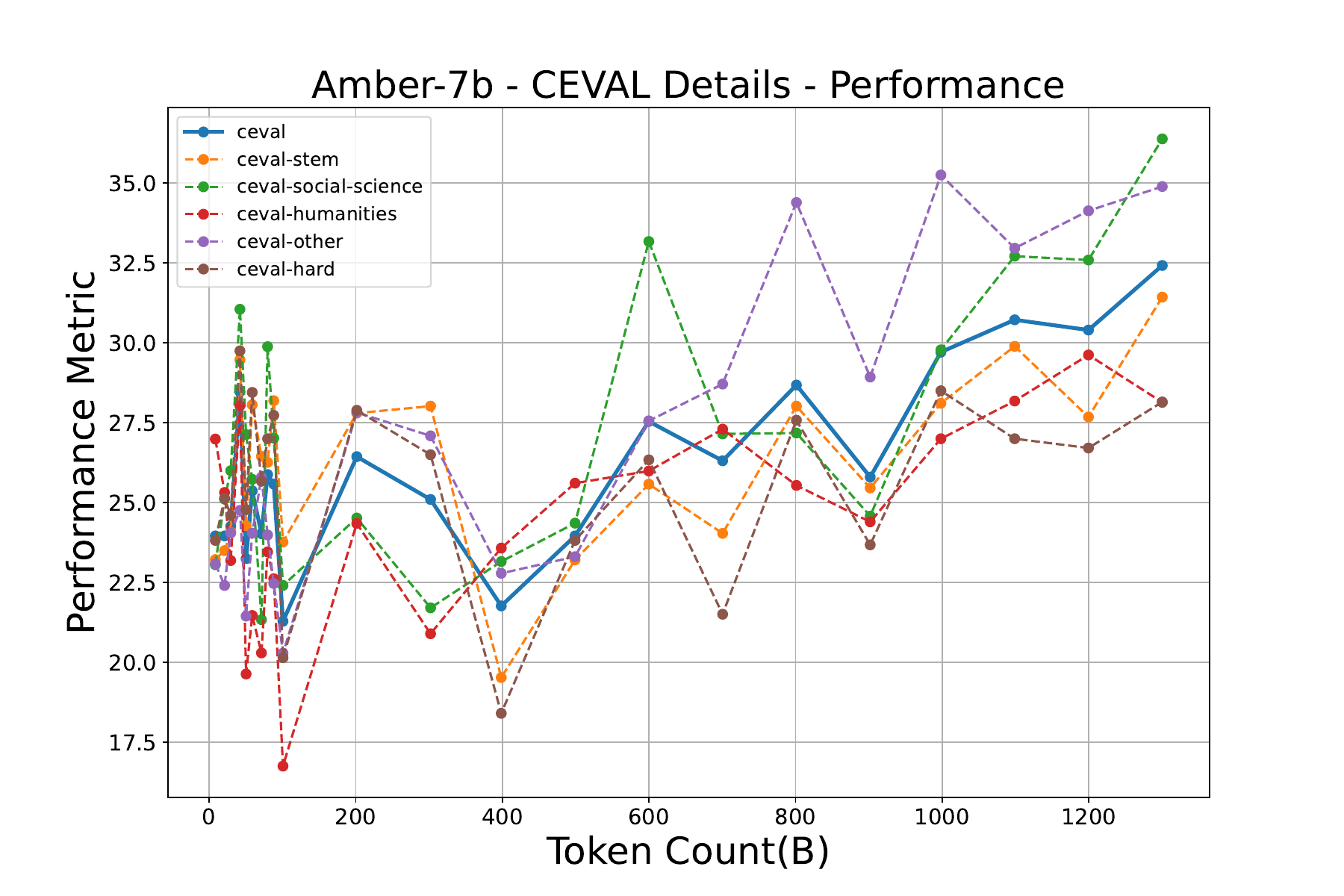}
        \caption{Amber-7B CEval}
        \label{fig:a-amber7b-ceval}
    \end{subfigure}
    \hfill
    \begin{subfigure}[b]{0.3\textwidth}
        \includegraphics[width=\textwidth]{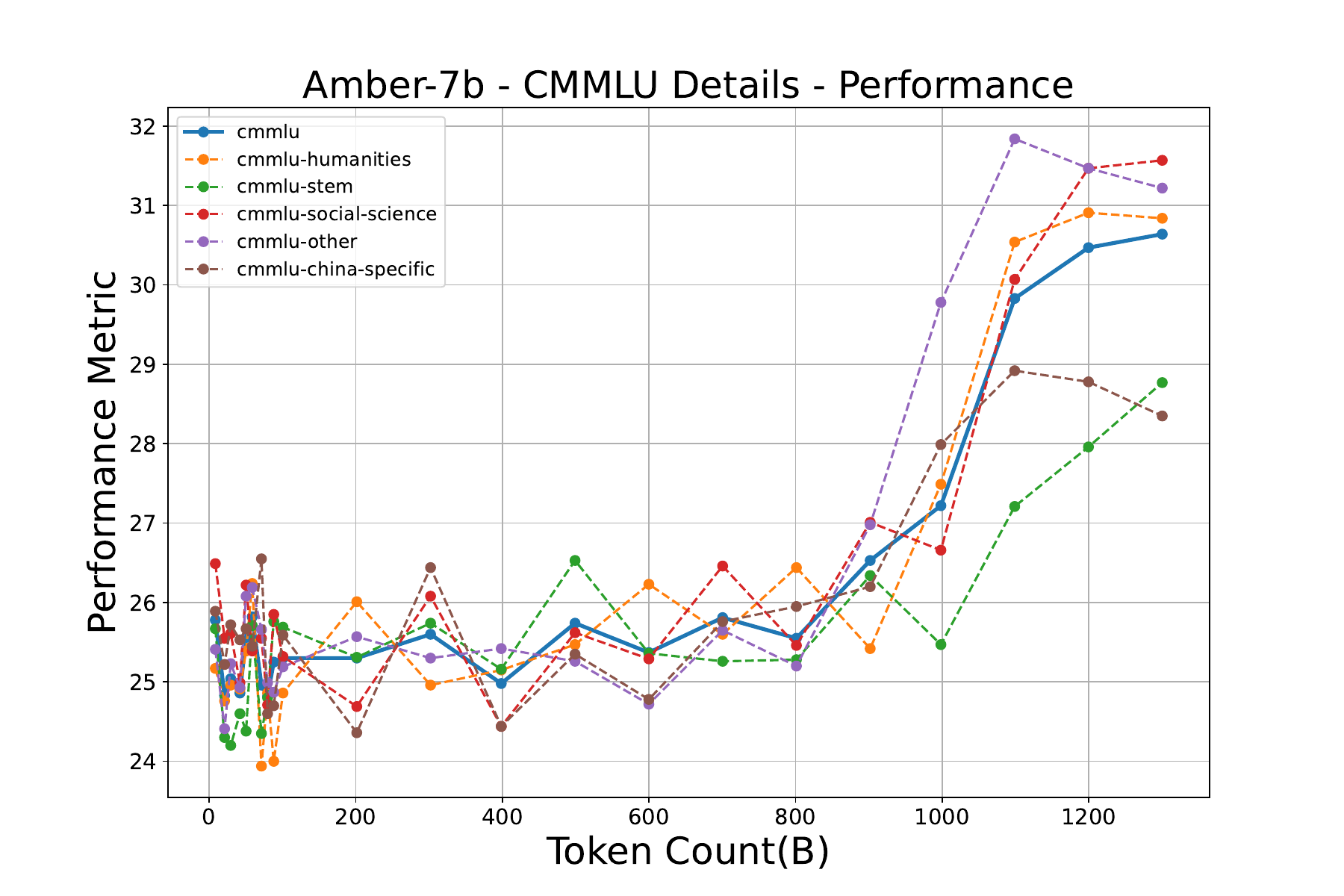}
        \caption{Amber-7B CMMLU}
        \label{fig:a-amber7b-cmmlu}
    \end{subfigure}
    
    \begin{subfigure}[b]{0.3\textwidth}
        \includegraphics[width=\textwidth]{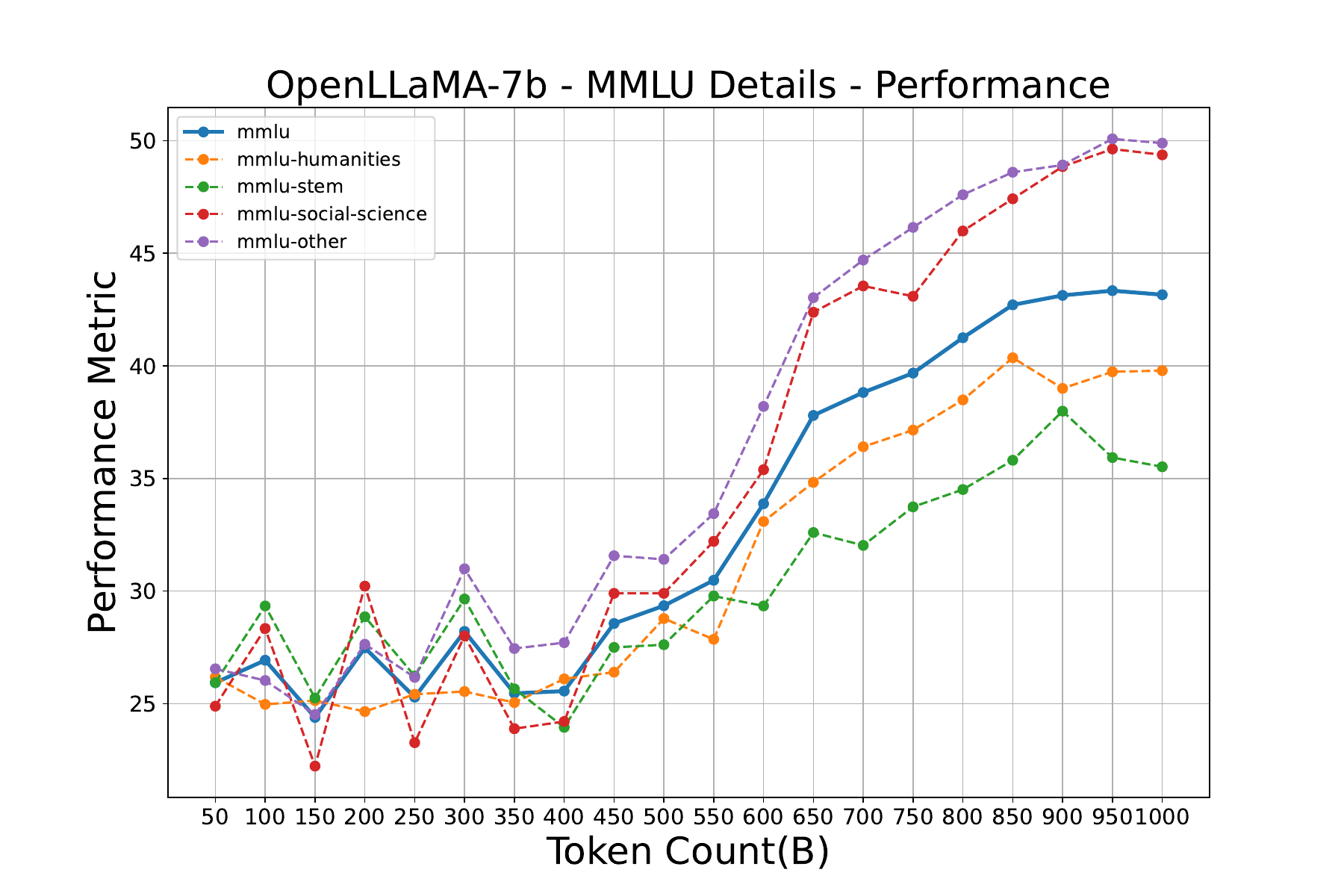}
        \caption{OpenLLaMA-7B MMLU}
        \label{fig:a-openllama7b-mmlu}
    \end{subfigure}
    \hfill
    \begin{subfigure}[b]{0.3\textwidth}
        \includegraphics[width=\textwidth]{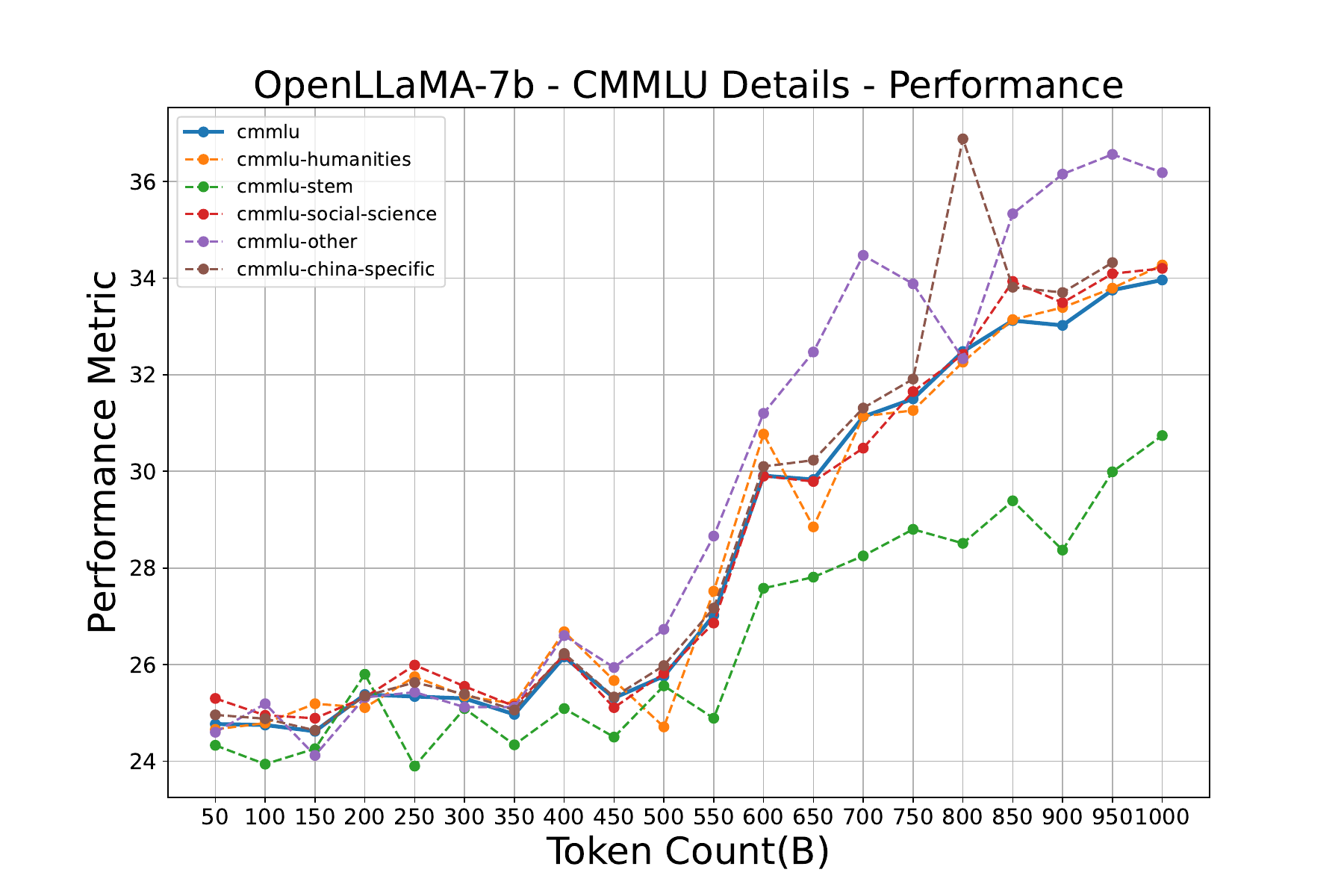}
        \caption{OpenLLaMA-7B CMMLU}
        \label{fig:a-openllama7b-cmmlu}
    \end{subfigure}

    \caption{Performance of DeepSeek-67B, Yi-34B, Baichuan-7B, DeepSeek-7B, Amber-7B, and OpenLLaMA-7B models across MMLU, CMMLU, and CEval benchmarks.}
    \label{fig:full-models-comparison-exam}
\end{figure}

\begin{figure}[htbp]
    \centering
    \begin{subfigure}[b]{0.3\textwidth}
        \includegraphics[width=\textwidth]{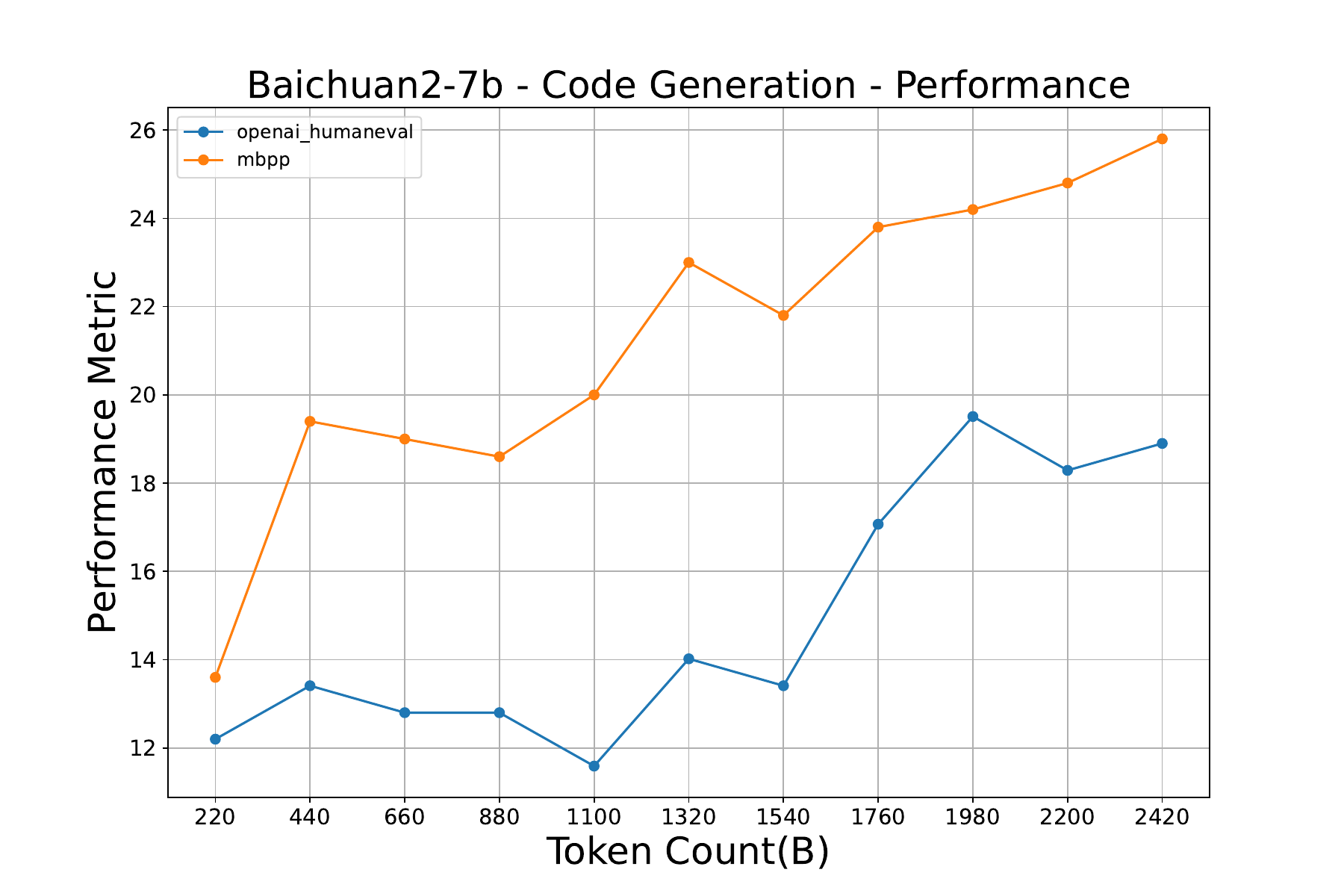}
        \caption{Baichuan-7B}
        \label{fig:Baichuan7b-Code-benchmarks}
    \end{subfigure}
    \hfill
    \begin{subfigure}[b]{0.3\textwidth}
        \includegraphics[width=\textwidth]{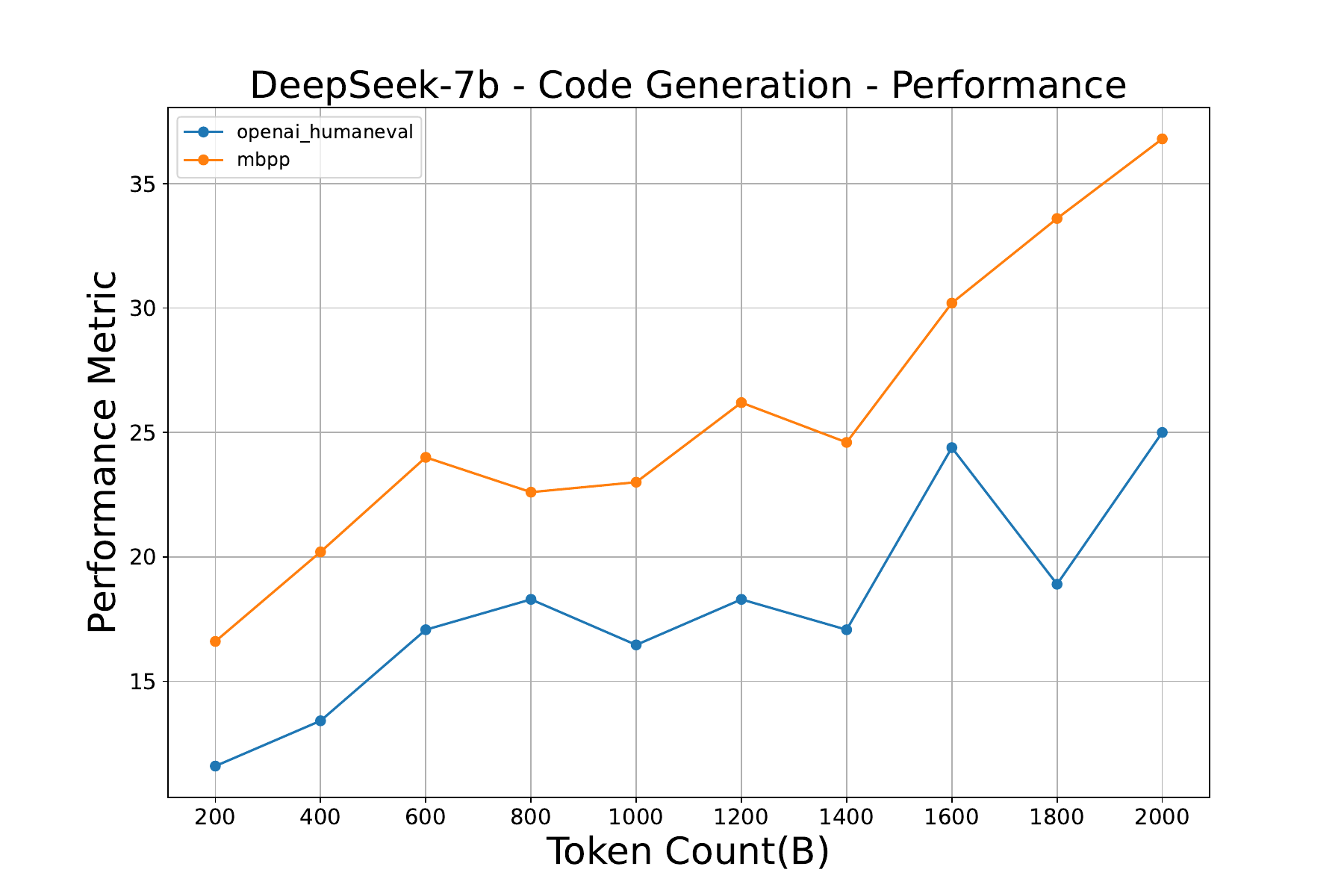}
        \caption{DeepSeek-7B}
        \label{fig:deepseek7b-Code-Benchmarks}
    \end{subfigure}
    \hfill
    \begin{subfigure}[b]{0.3\textwidth}
        \includegraphics[width=\textwidth]{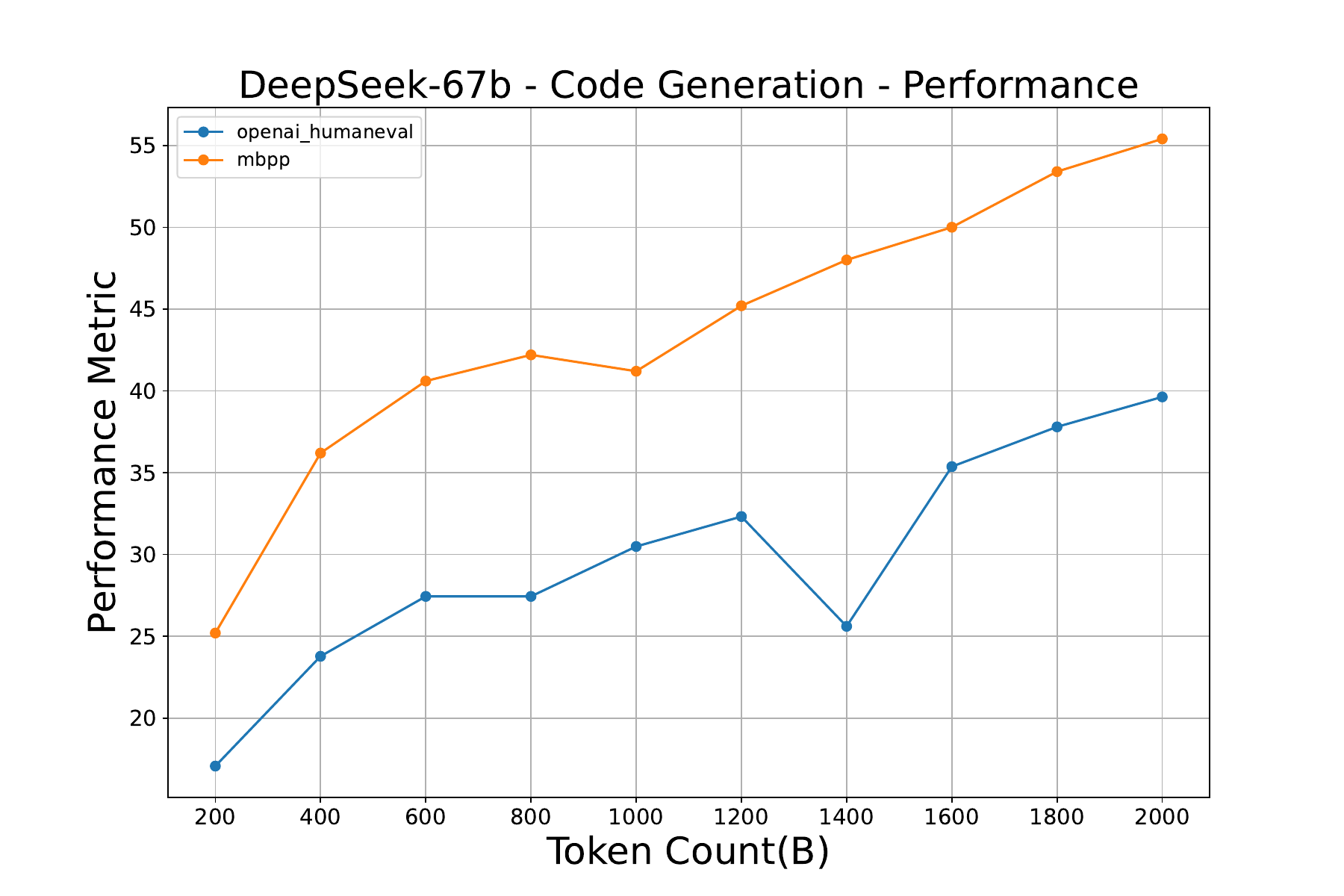}
        \caption{DeepSeek-67B}
        \label{fig:deepseek67b-Code-Benchmarks}
    \end{subfigure}
    
    \begin{subfigure}[b]{0.3\textwidth}
        \includegraphics[width=\textwidth]{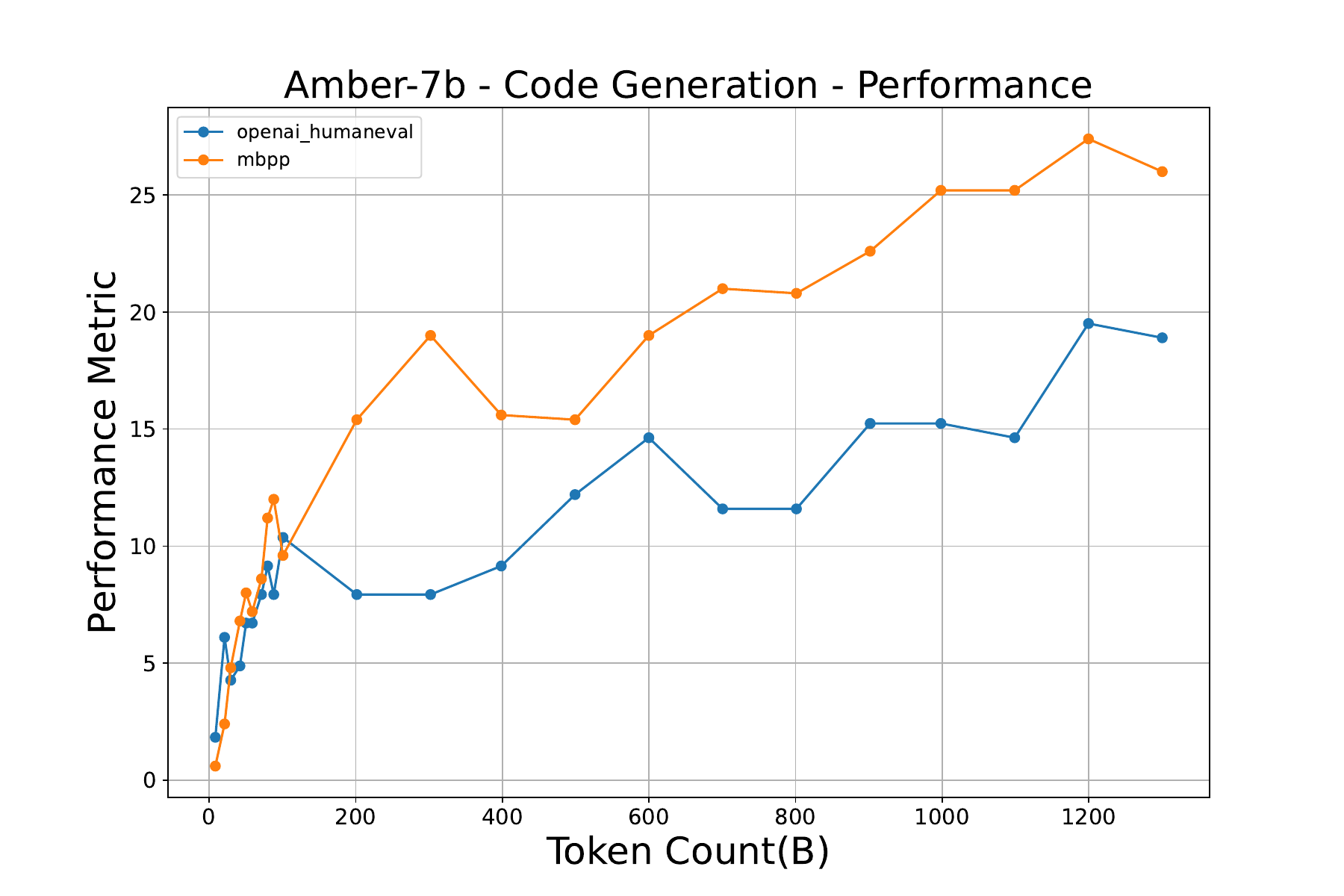}
        \caption{Amber-7B}
        \label{fig:amber7B-Code-Benchmarks}
    \end{subfigure}
    \hfill
    \begin{subfigure}[b]{0.3\textwidth}
        \includegraphics[width=\textwidth]{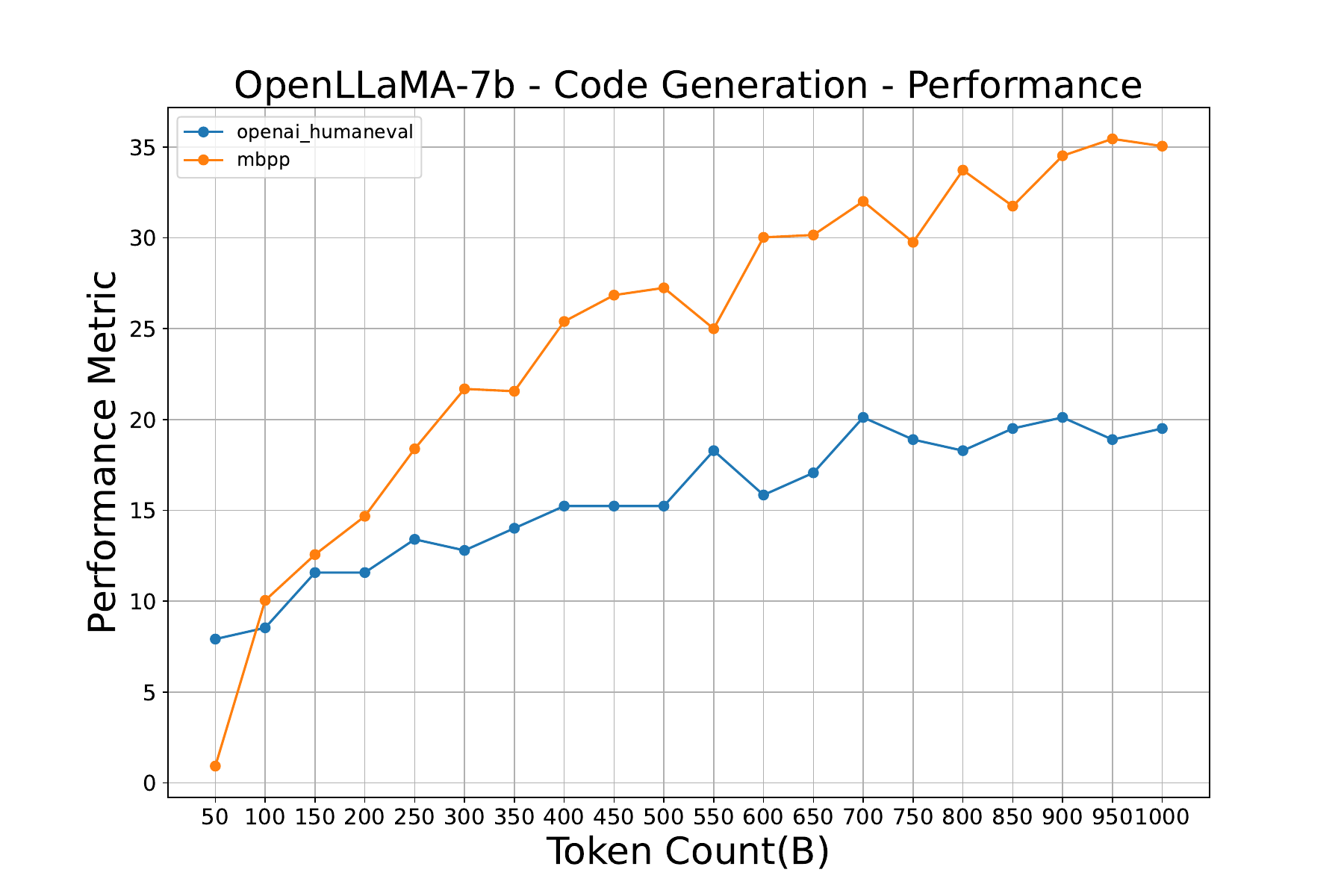}
        \caption{OpenLLaMA-7B}
        \label{fig:openllama7b-Code-Benchmarks}
    \end{subfigure}
    \hfill
    \begin{subfigure}[b]{0.3\textwidth}
        \includegraphics[width=\textwidth]{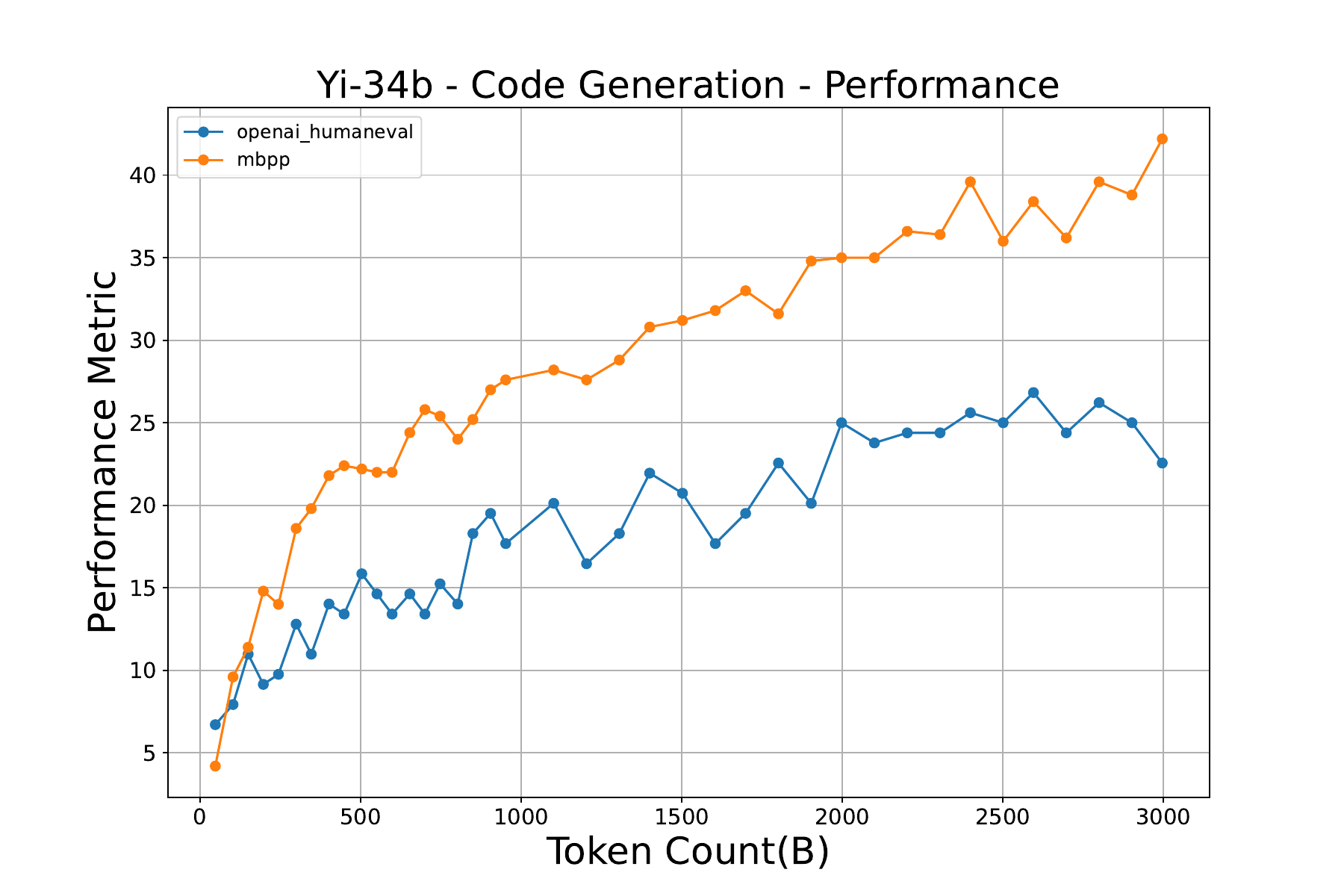}
        \caption{Yi-34B}
        \label{fig:yi34b-Code-Benchmarks}
    \end{subfigure}
    
    \caption{Performance of Baichuan-7B, DeepSeek-7B, DeepSeek-67B, Amber-7B, OpenLLaMA-7B, and Yi-34B models across code benchmarks.}
    \label{fig:models-comparison-code-benckmarks}
\end{figure}

\begin{figure}[htbp]
    \centering
    \begin{subfigure}[b]{0.3\textwidth}
        \includegraphics[width=\textwidth]{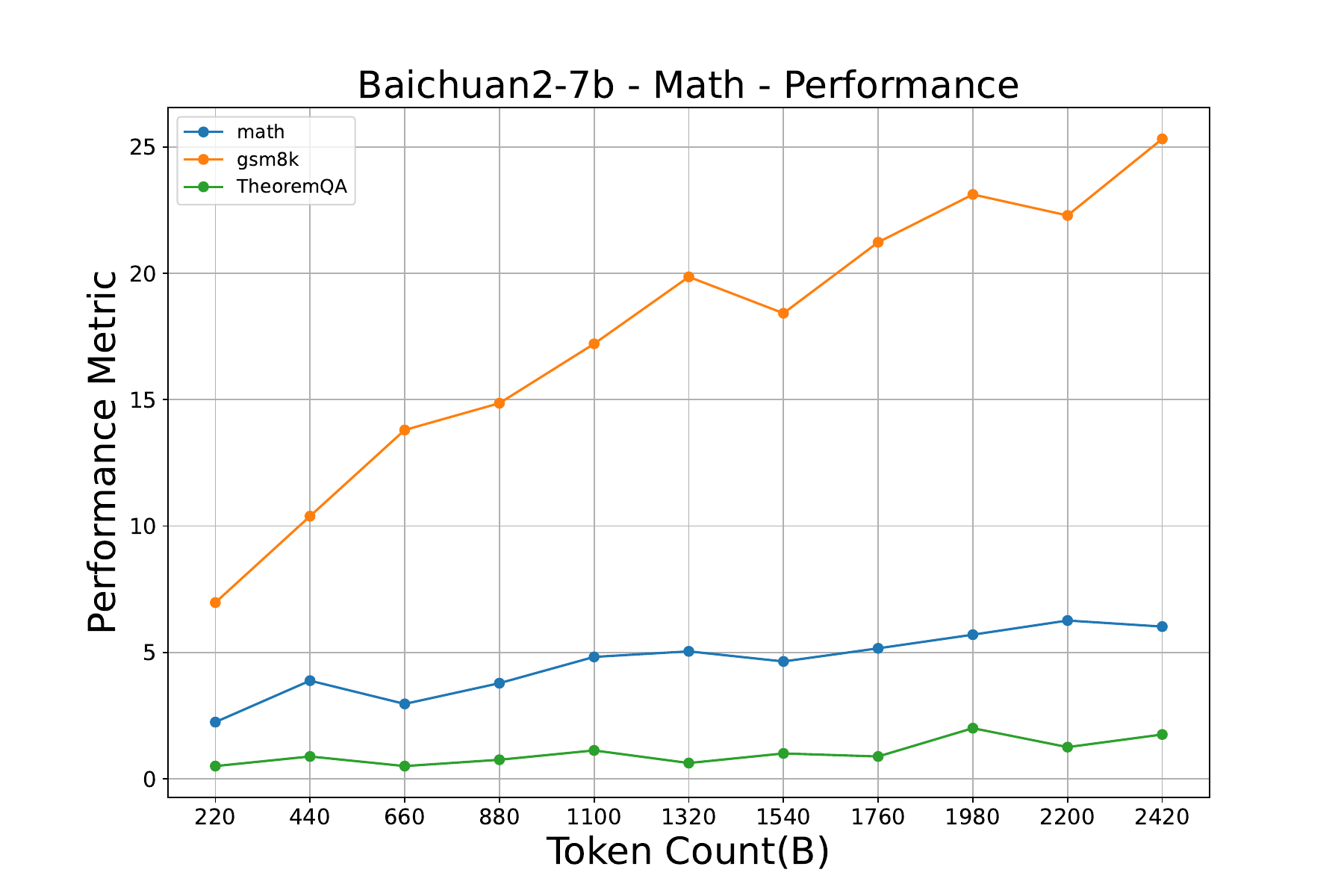}
        \caption{Baichuan-7B}
        \label{fig:Baichuan7b-Math-benchmarks}
    \end{subfigure}
    \hfill
    \begin{subfigure}[b]{0.3\textwidth}
        \includegraphics[width=\textwidth]{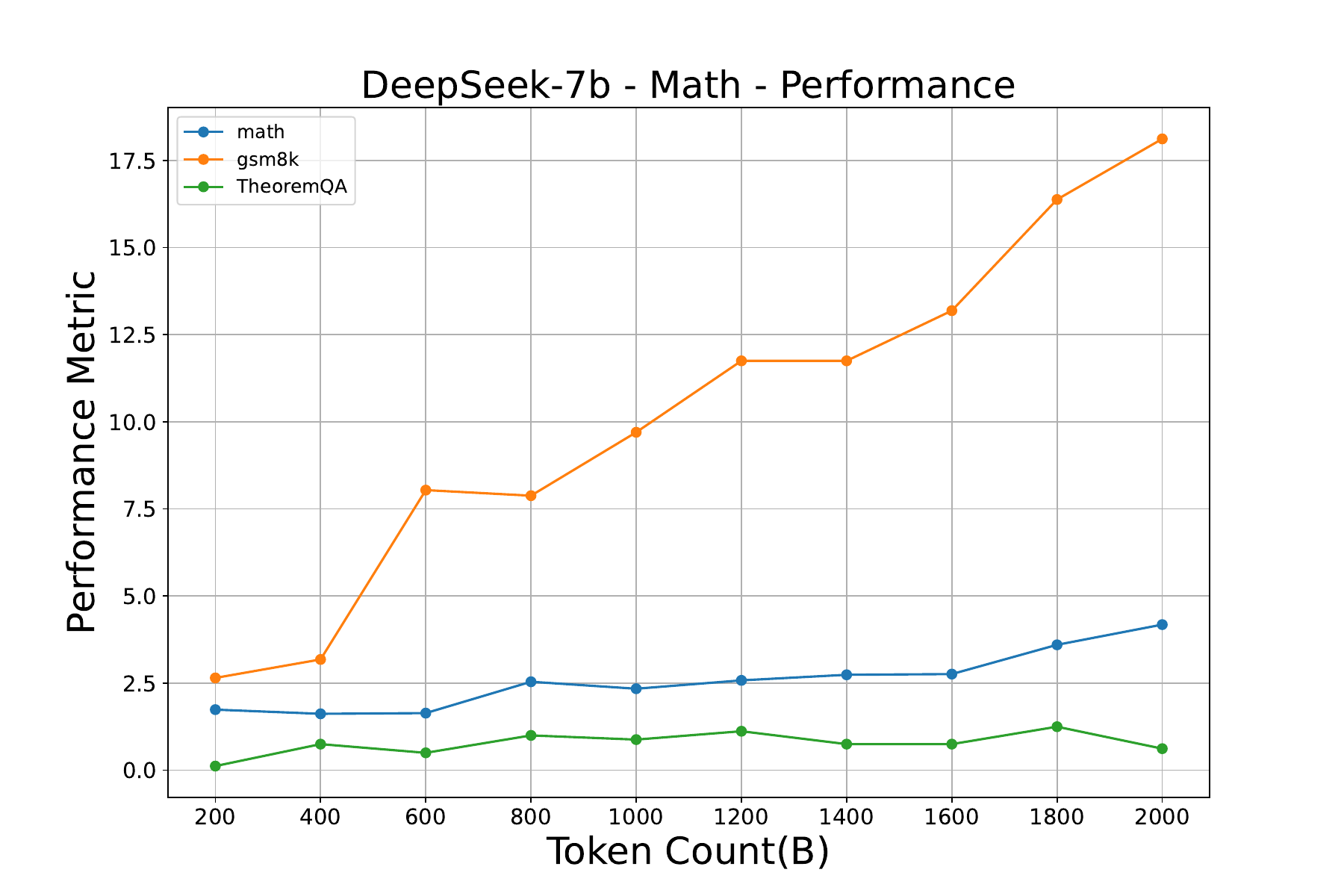}
        \caption{DeepSeek-7B}
        \label{fig:deepseek7b-Math-Benchmarks}
    \end{subfigure}
    \hfill
    \begin{subfigure}[b]{0.3\textwidth}
        \includegraphics[width=\textwidth]{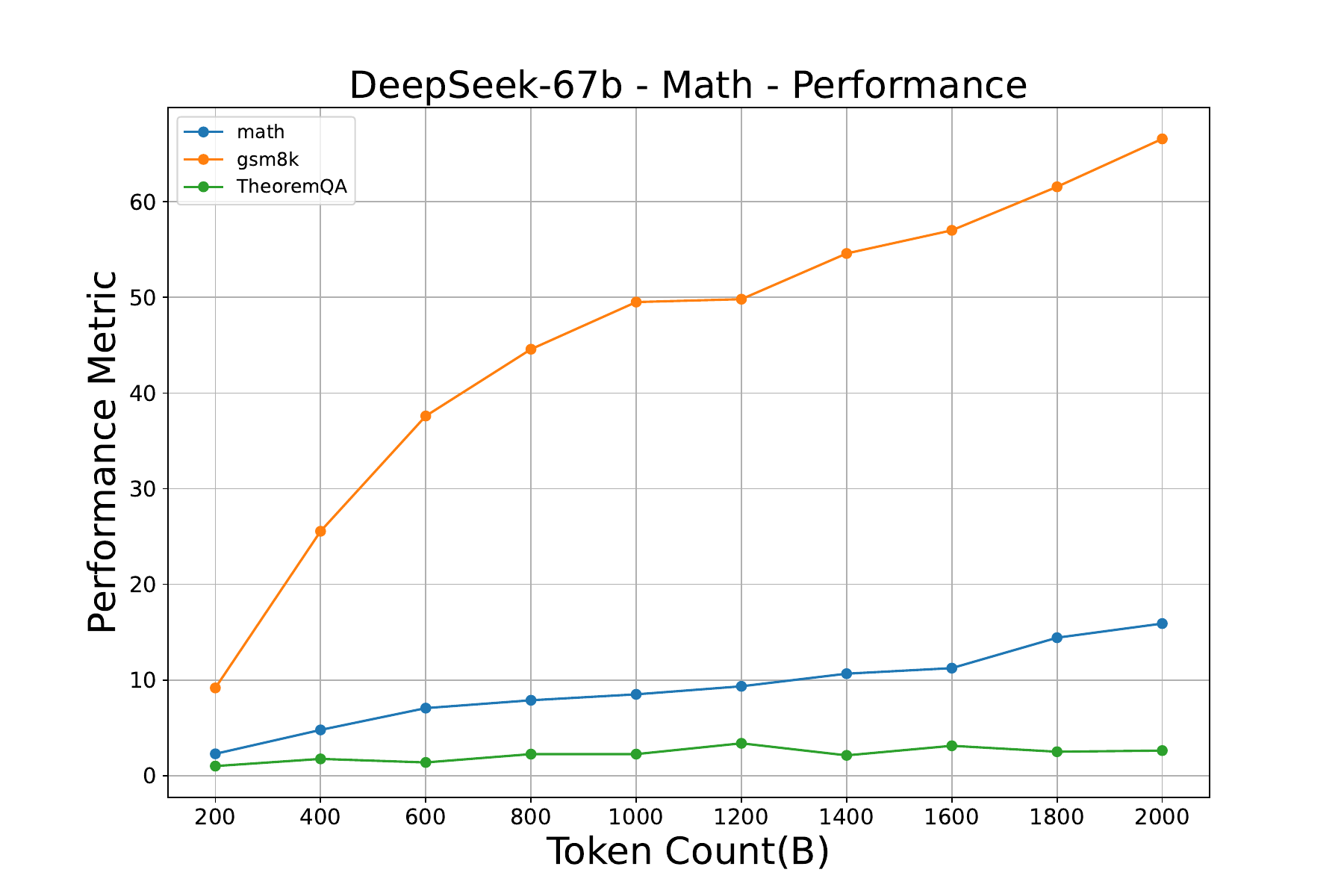}
        \caption{DeepSeek-67B}
        \label{fig:deepseek67b-Math-Benchmarks}
    \end{subfigure}
    
    \begin{subfigure}[b]{0.3\textwidth}
        \includegraphics[width=\textwidth]{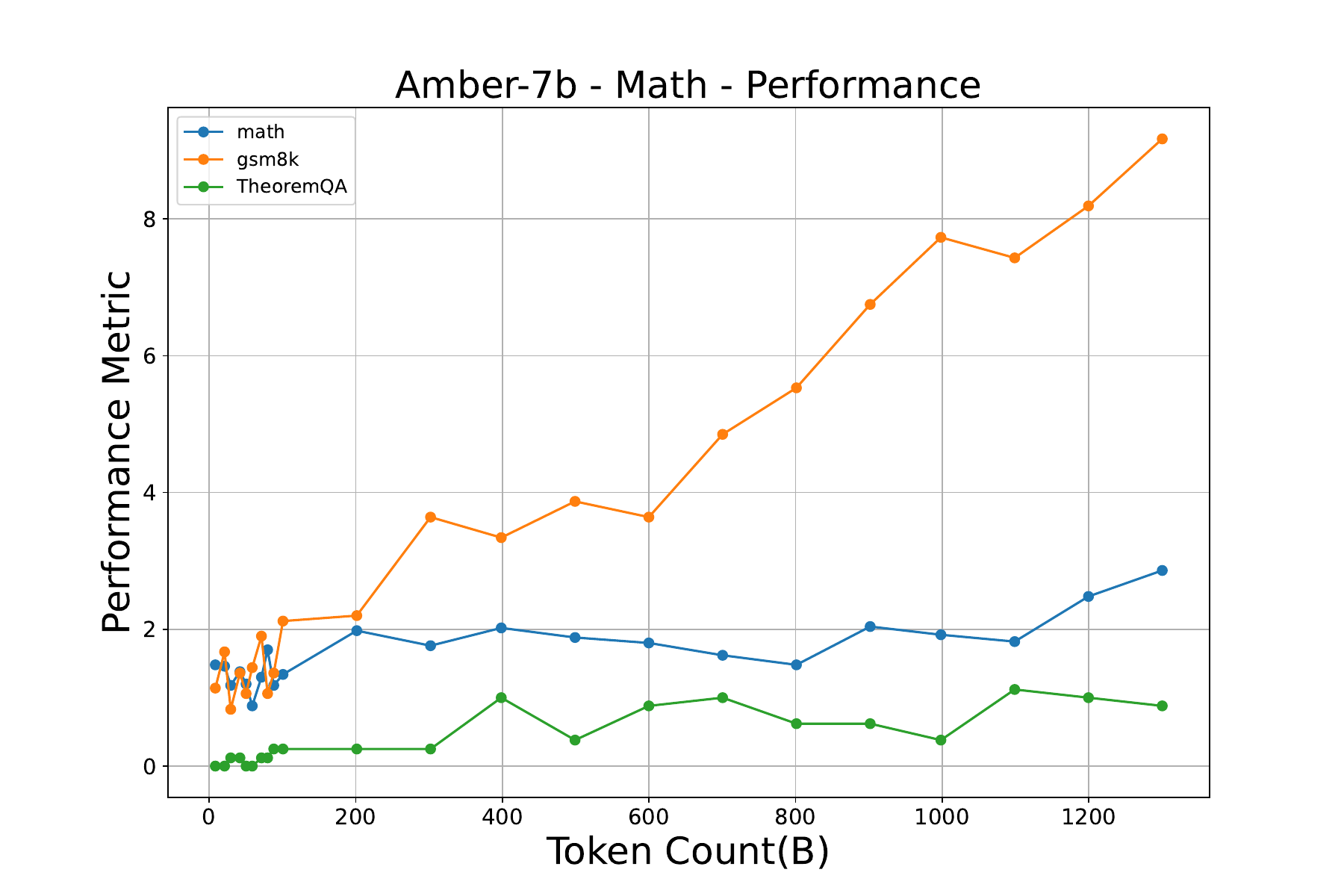}
        \caption{Amber-7B}
        \label{fig:amber7B-Math-Benchmarks}
    \end{subfigure}
    \hfill
    \begin{subfigure}[b]{0.3\textwidth}
        \includegraphics[width=\textwidth]{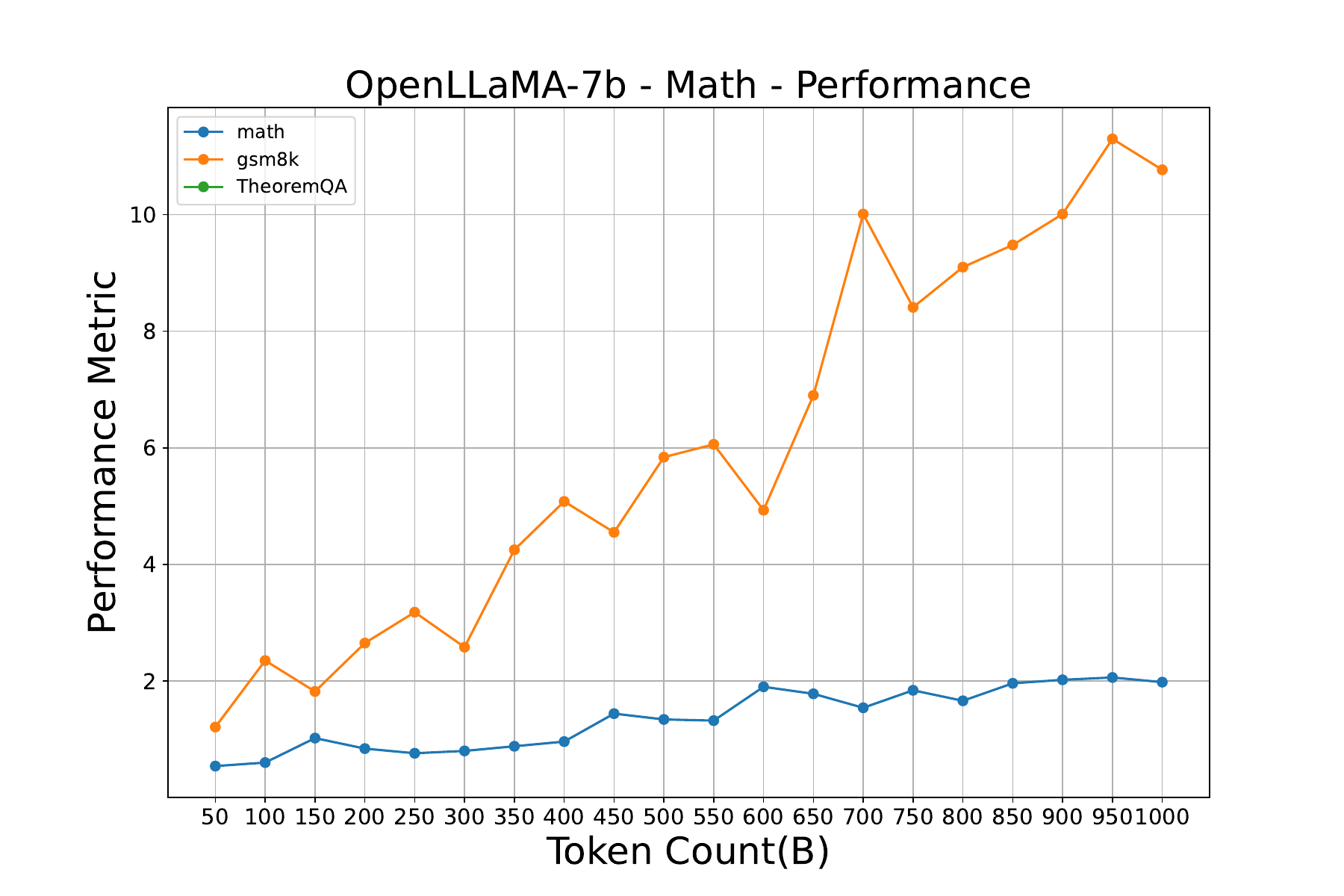}
        \caption{OpenLLaMA-7B}
        \label{fig:openllama7b-Math-Benchmarks}
    \end{subfigure}
    \hfill
    \begin{subfigure}[b]{0.3\textwidth}
        \includegraphics[width=\textwidth]{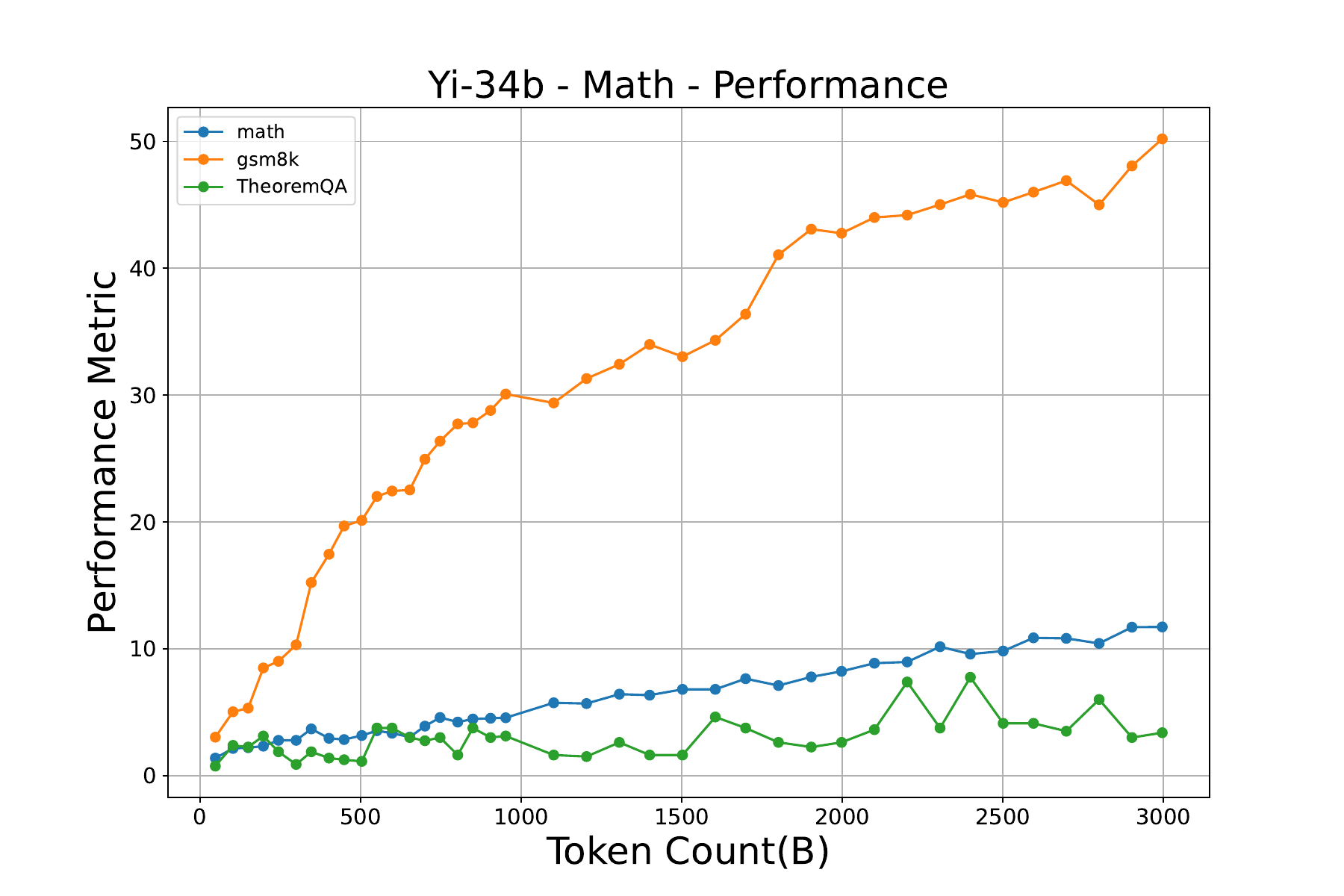}
        \caption{Yi-34B Math Benchmarks}
        \label{fig:yi34b-Math-Benchmarks}
    \end{subfigure}
    
    \caption{Performance of Baichuan-7B, DeepSeek-7B, DeepSeek-67B, Amber-7B, OpenLLaMA-7B, and Yi-34B models across math benchmarks.}
    \label{fig:models-comparison-math-benckmarks}
\end{figure}

\begin{figure}[htbp]
    \centering
    \begin{subfigure}[b]{0.3\textwidth}
        \includegraphics[width=\textwidth]{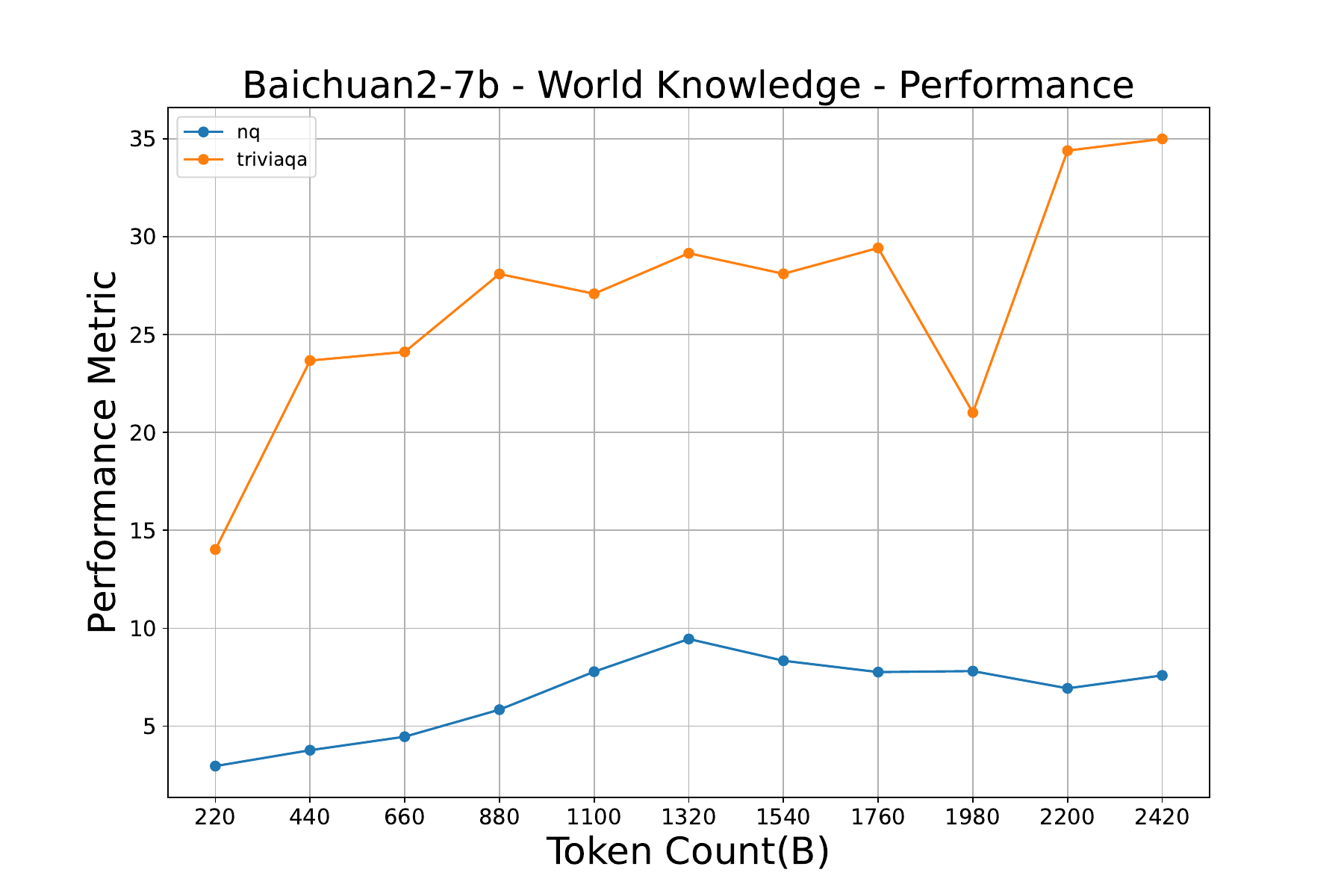}
        \caption{Baichuan-7B}
        \label{fig:Baichuan7b-World-benchmarks}
    \end{subfigure}
    \hfill
    \begin{subfigure}[b]{0.3\textwidth}
        \includegraphics[width=\textwidth]{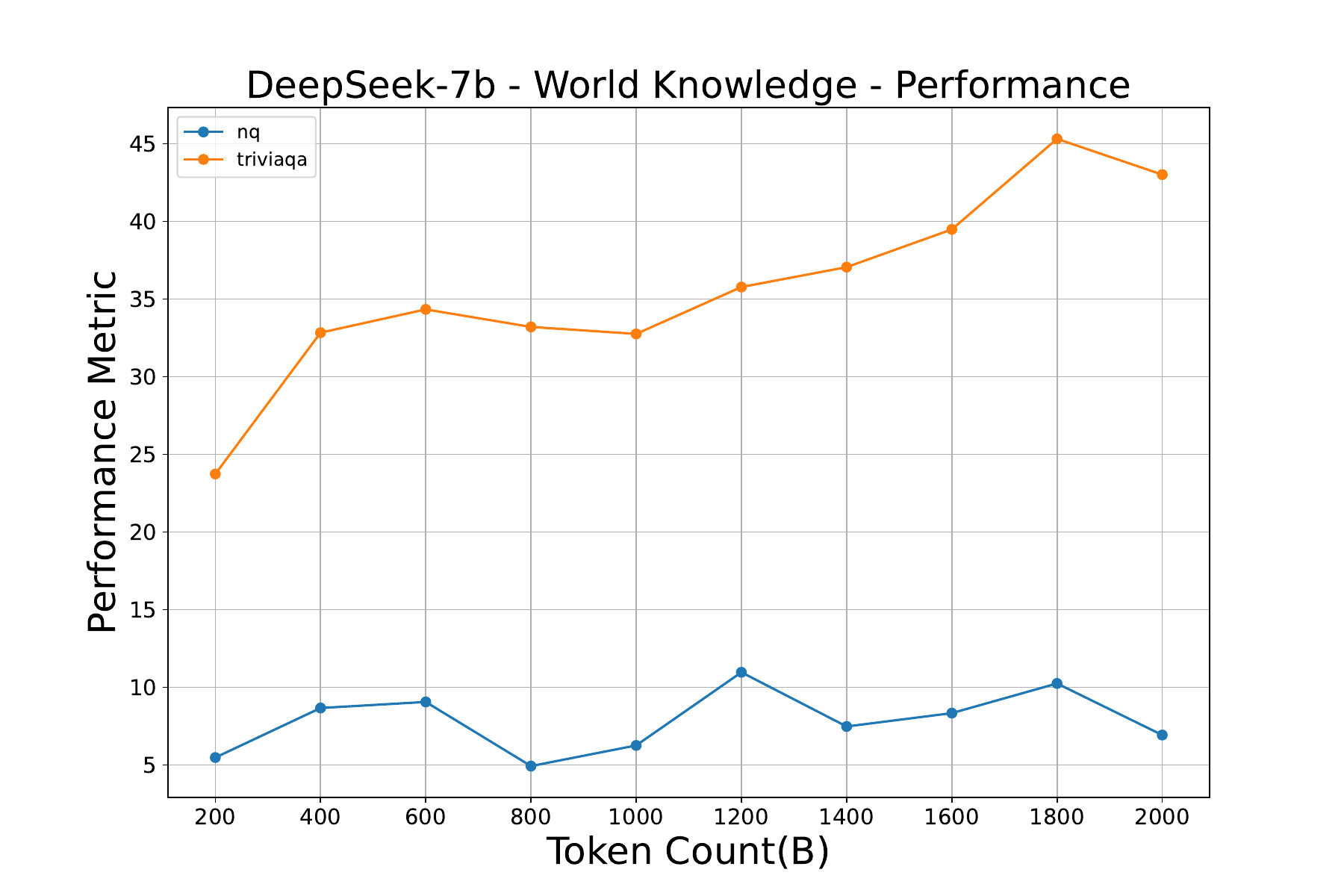}
        \caption{DeepSeek-7B}
        \label{fig:deepseek7b-World-Benchmarks}
    \end{subfigure}
    \hfill
    \begin{subfigure}[b]{0.3\textwidth}
        \includegraphics[width=\textwidth]{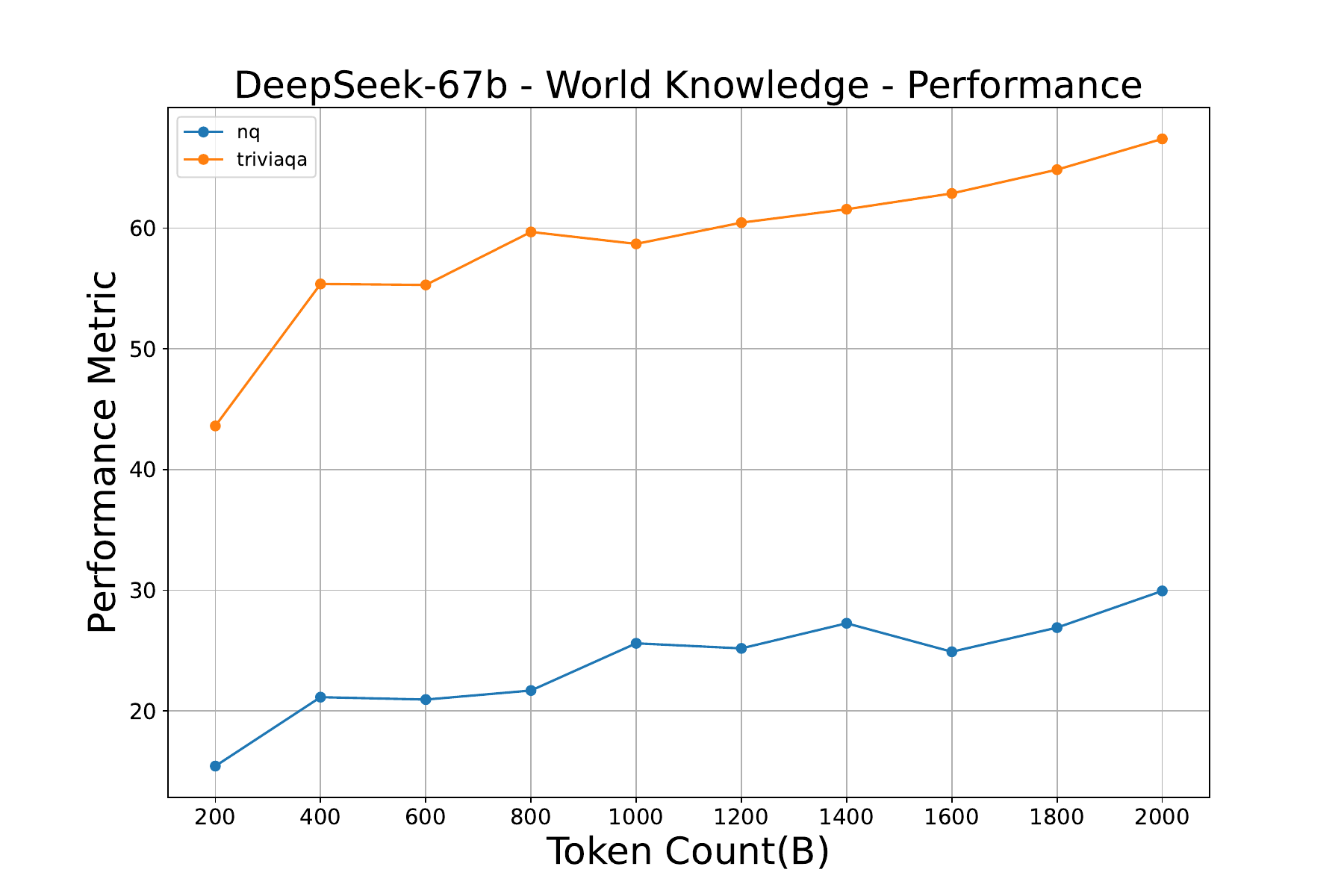}
        \caption{DeepSeek-67B}
        \label{fig:deepseek67b-World-Benchmarks}
    \end{subfigure}
    
    \begin{subfigure}[b]{0.3\textwidth}
        \includegraphics[width=\textwidth]{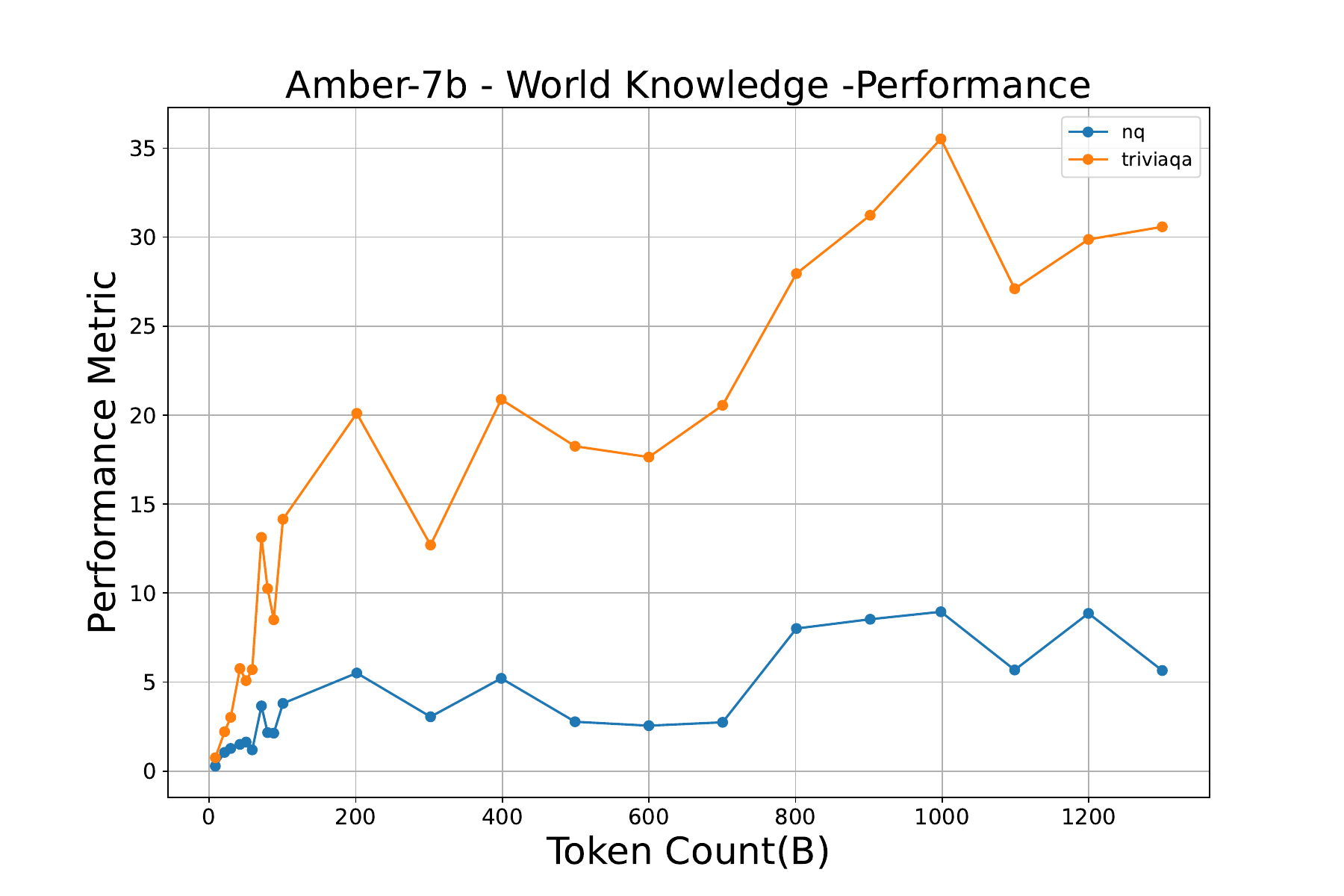}
        \caption{Amber-7B}
        \label{fig:amber7B-World-Benchmarks}
    \end{subfigure}
    \hfill
    \begin{subfigure}[b]{0.3\textwidth}
        \includegraphics[width=\textwidth]{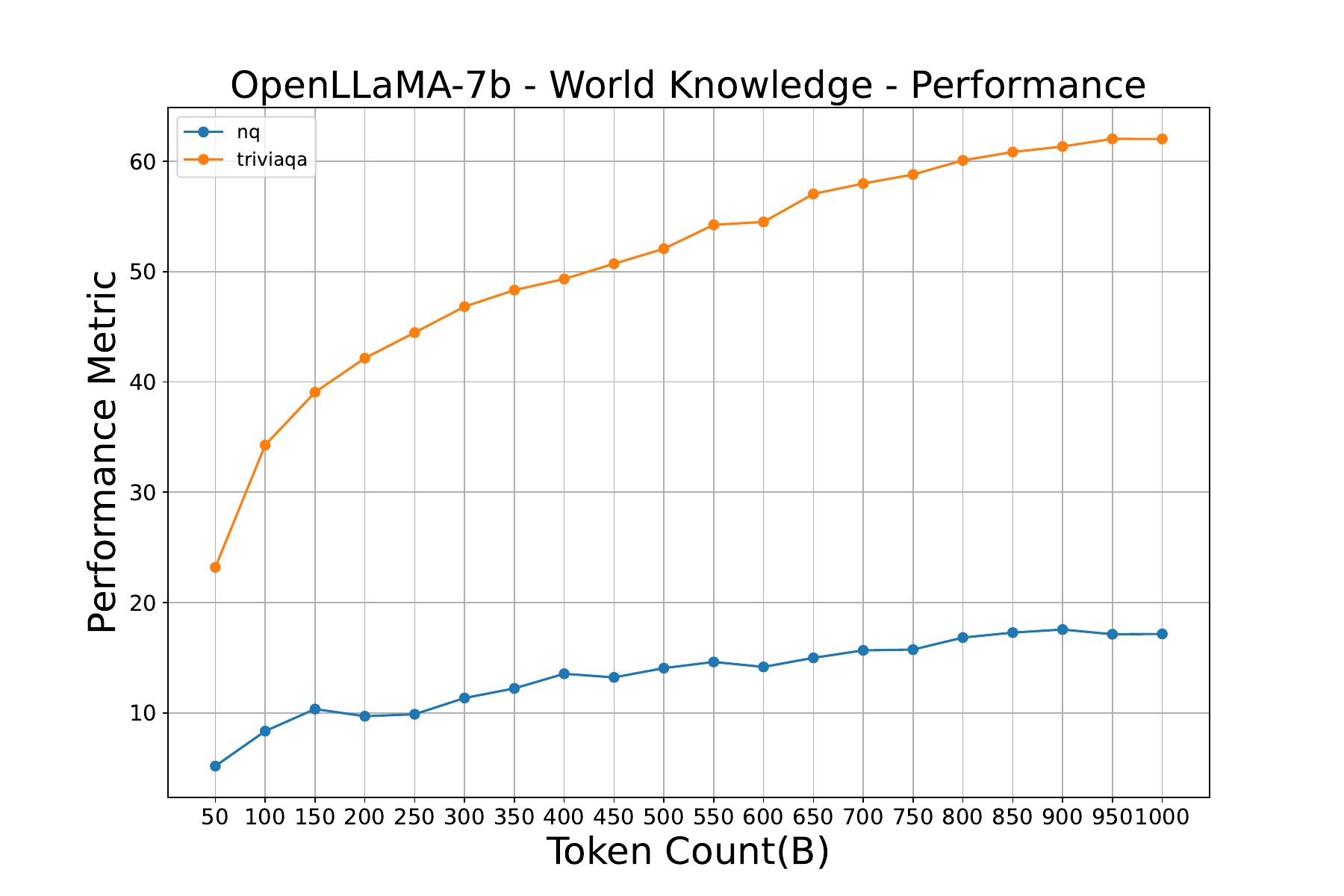}
        \caption{OpenLLaMA-7B}
        \label{fig:openllama7b-World-Benchmarks}
    \end{subfigure}
    \hfill
    \begin{subfigure}[b]{0.3\textwidth}
        \includegraphics[width=\textwidth]{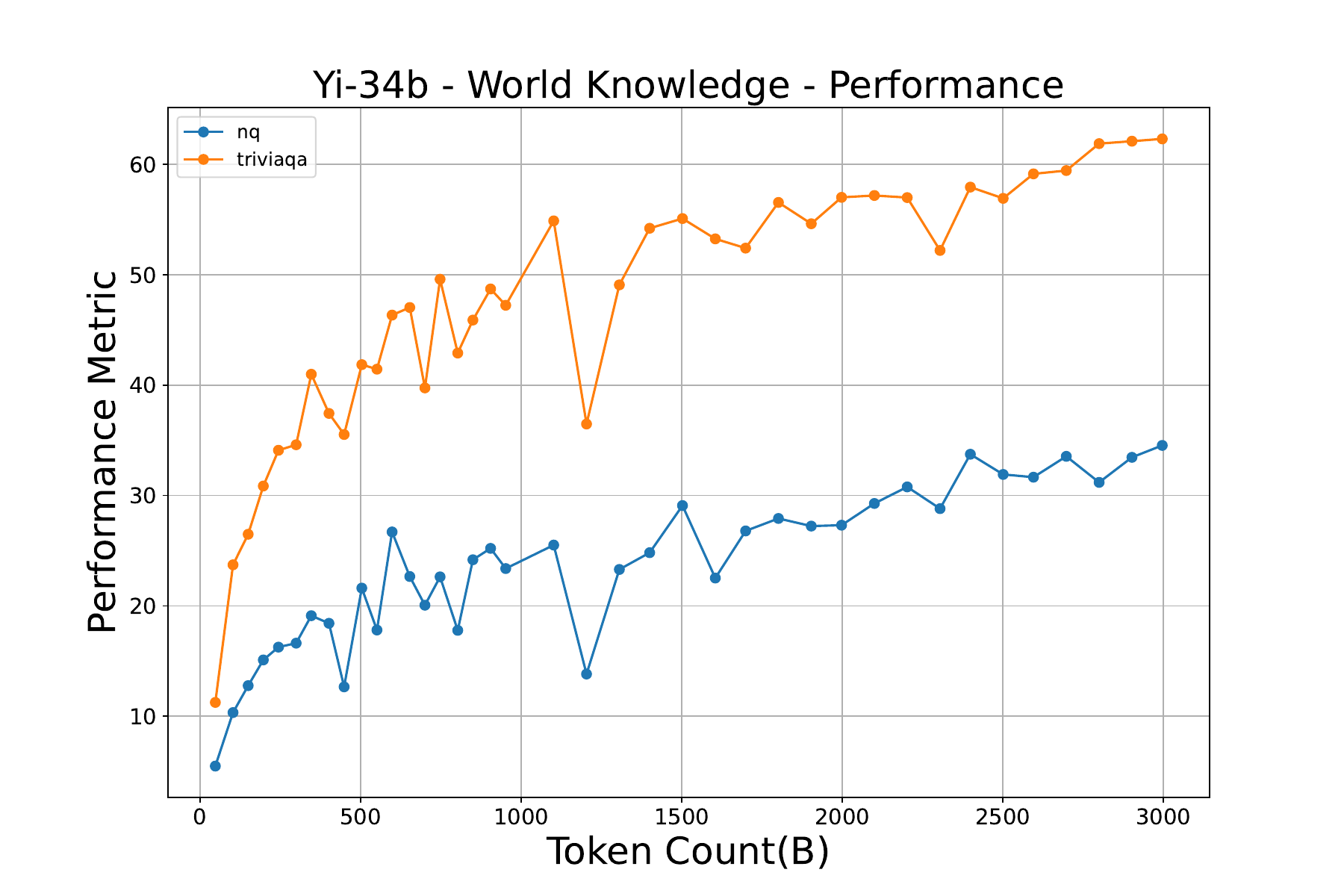}
        \caption{Yi-34B}
        \label{fig:yi34b-World-Benchmarks}
    \end{subfigure}
    
    \caption{Performance of Baichuan-7B, DeepSeek-7B, DeepSeek-67B, Amber-7B, OpenLLaMA-7B, and Yi-34B models across world knowledge benchmarks.}
    \label{fig:models-comparison-world-benckmarks}
\end{figure}

\begin{figure}[htbp]
    \centering
    \begin{subfigure}[b]{0.3\textwidth}
        \includegraphics[width=\textwidth]{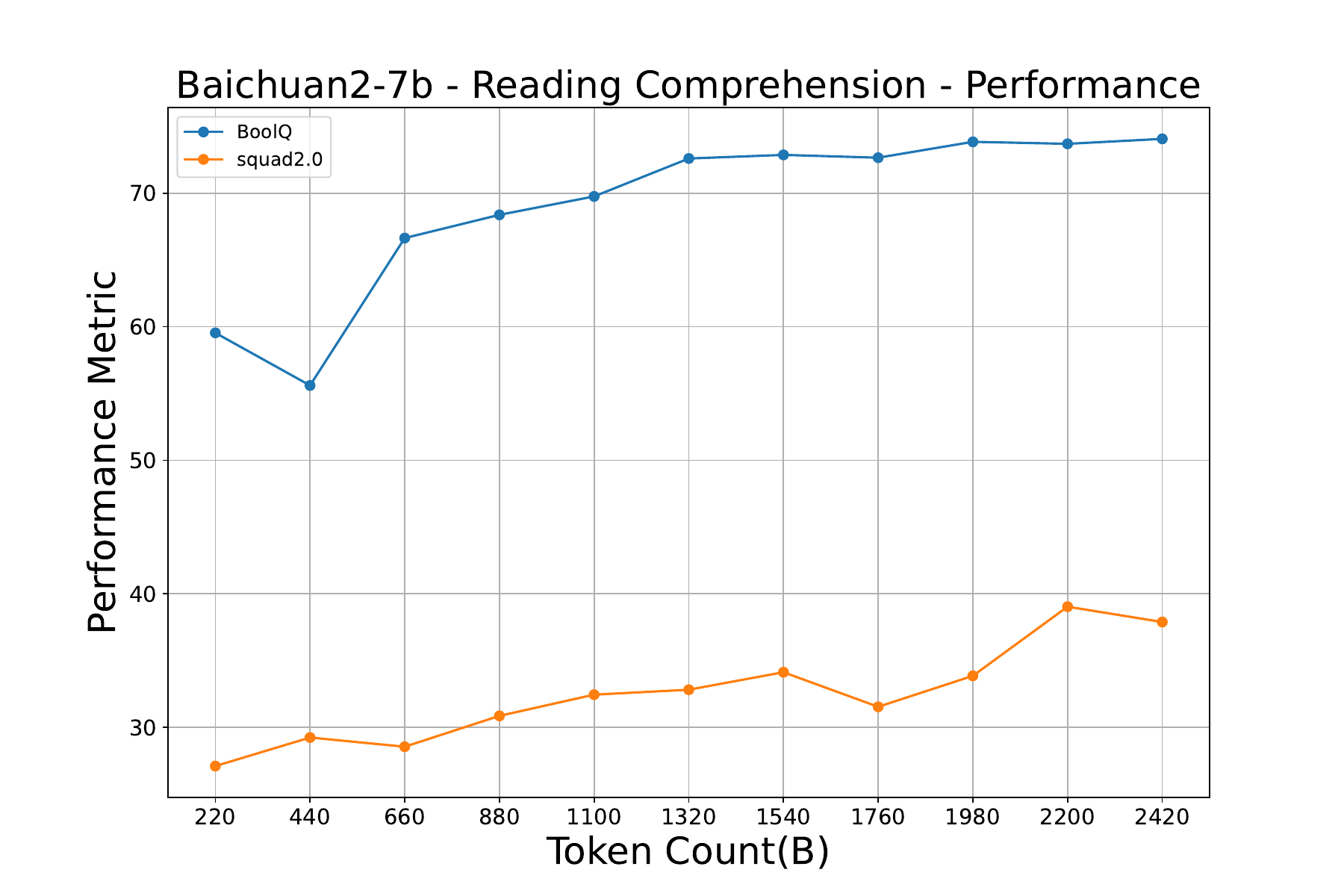}
        \caption{Baichuan-7B}
        \label{fig:Baichuan7b-Read-benchmarks}
    \end{subfigure}
    \hfill
    \begin{subfigure}[b]{0.3\textwidth}
        \includegraphics[width=\textwidth]{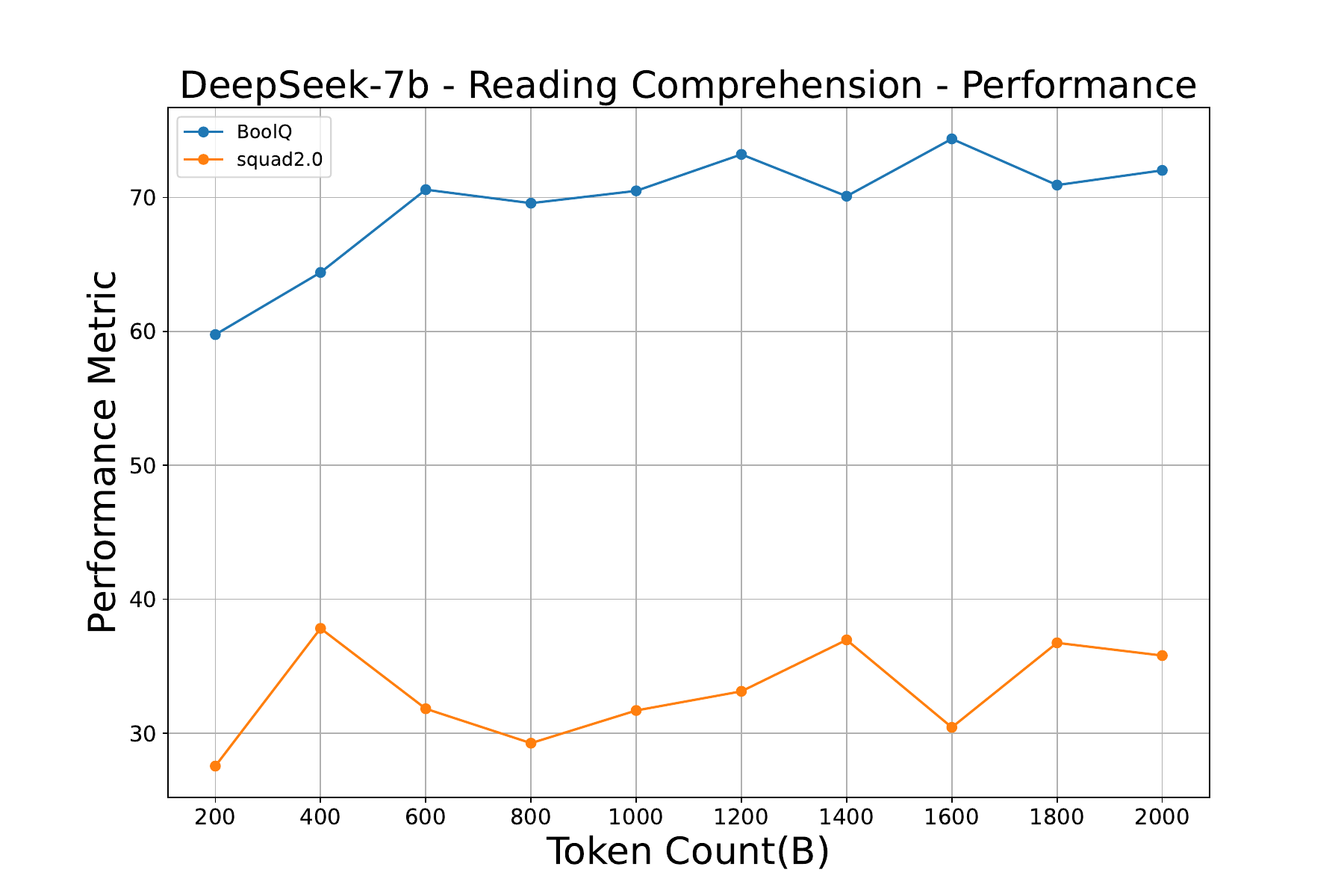}
        \caption{DeepSeek-7B}
        \label{fig:deepseek7b-Read-Benchmarks}
    \end{subfigure}
    \hfill
    \begin{subfigure}[b]{0.3\textwidth}
        \includegraphics[width=\textwidth]{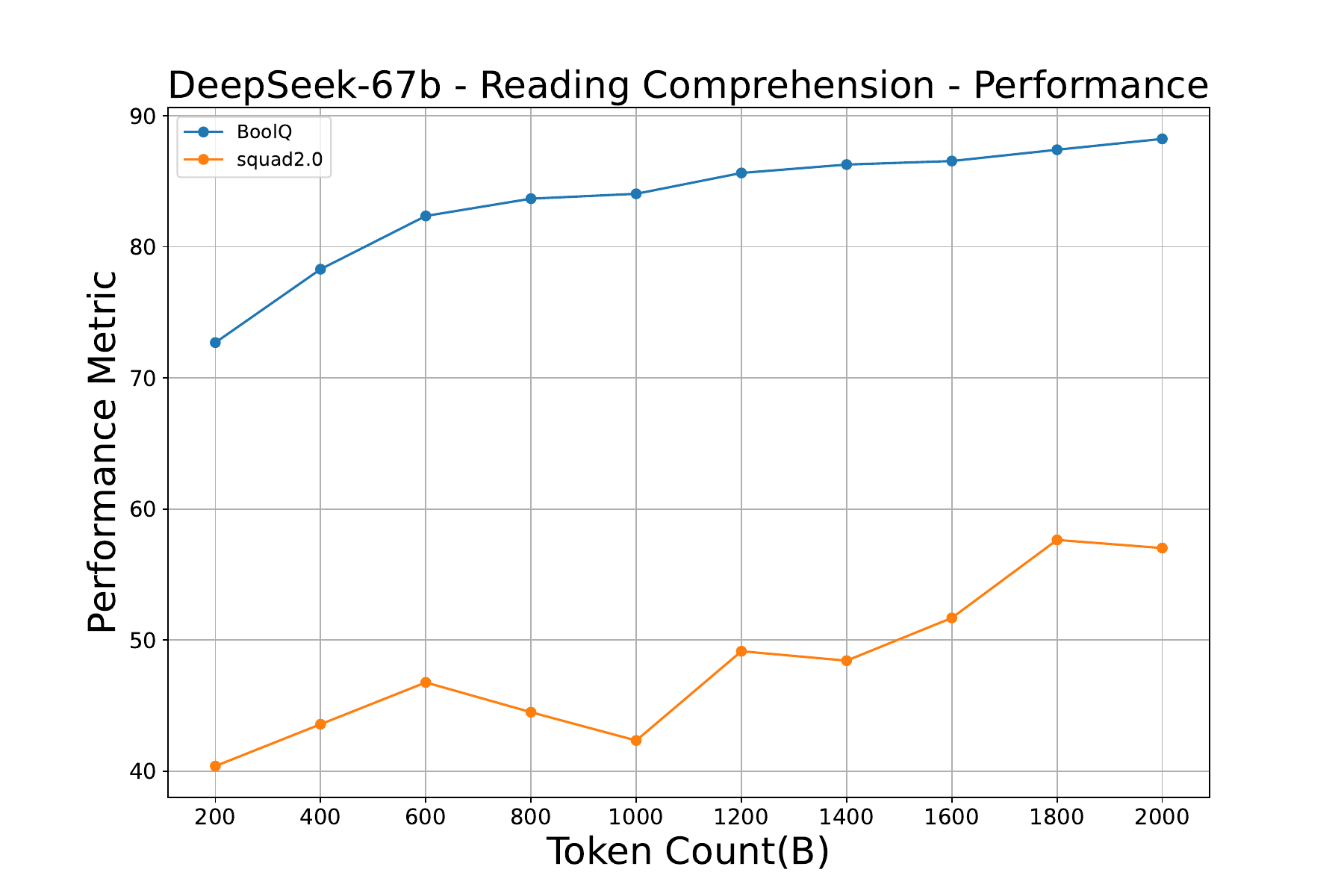}
        \caption{DeepSeek-67B}
        \label{fig:deepseek67b-Read-Benchmarks}
    \end{subfigure}
    
    \begin{subfigure}[b]{0.3\textwidth}
        \includegraphics[width=\textwidth]{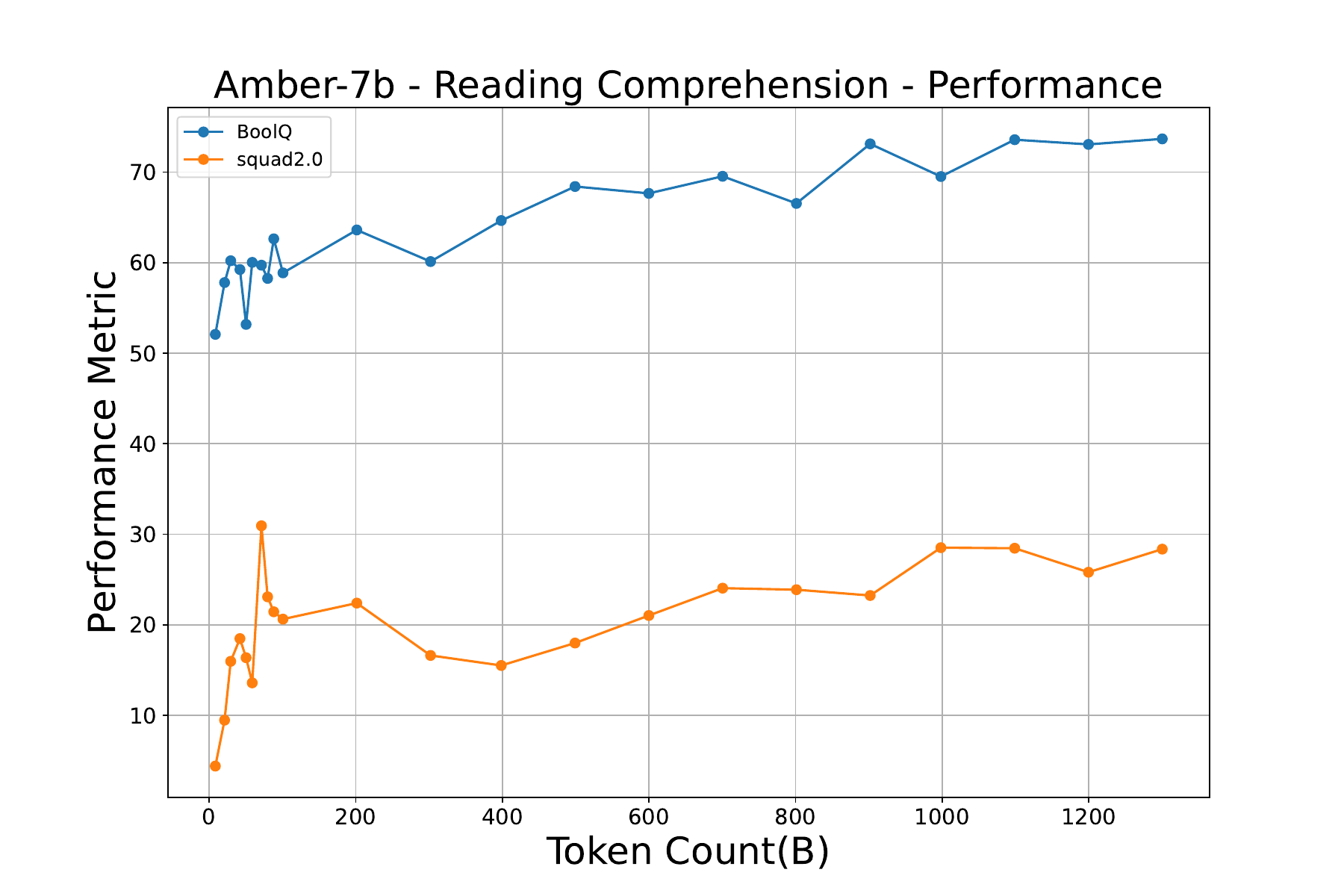}
        \caption{Amber-7B}
        \label{fig:amber7B-Read-Benchmarks}
    \end{subfigure}
    \hfill
    \begin{subfigure}[b]{0.3\textwidth}
        \includegraphics[width=\textwidth]{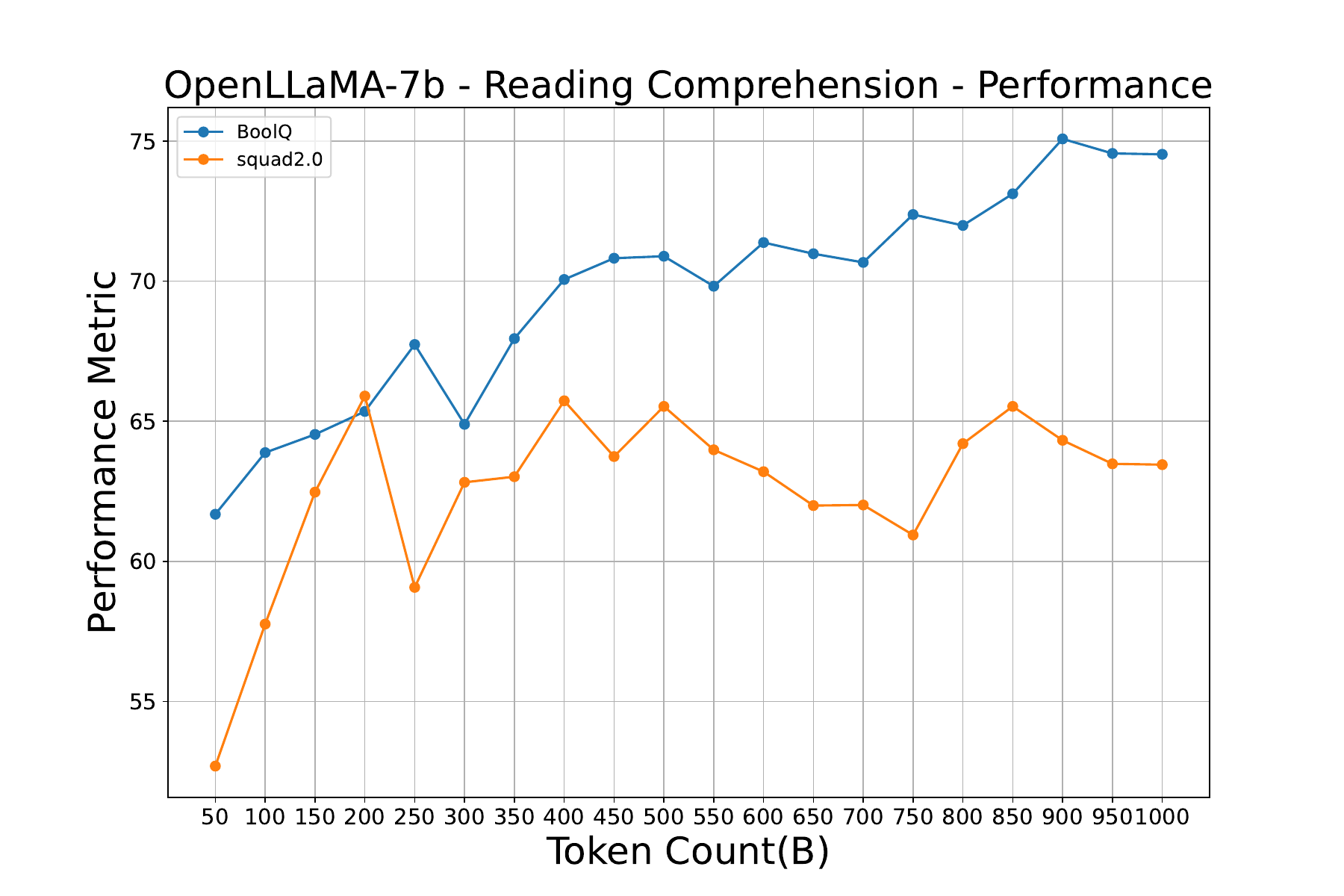}
        \caption{OpenLLaMA-7B}
        \label{fig:openllama7b-Read-Benchmarks}
    \end{subfigure}
    \hfill
    \begin{subfigure}[b]{0.3\textwidth}
        \includegraphics[width=\textwidth]{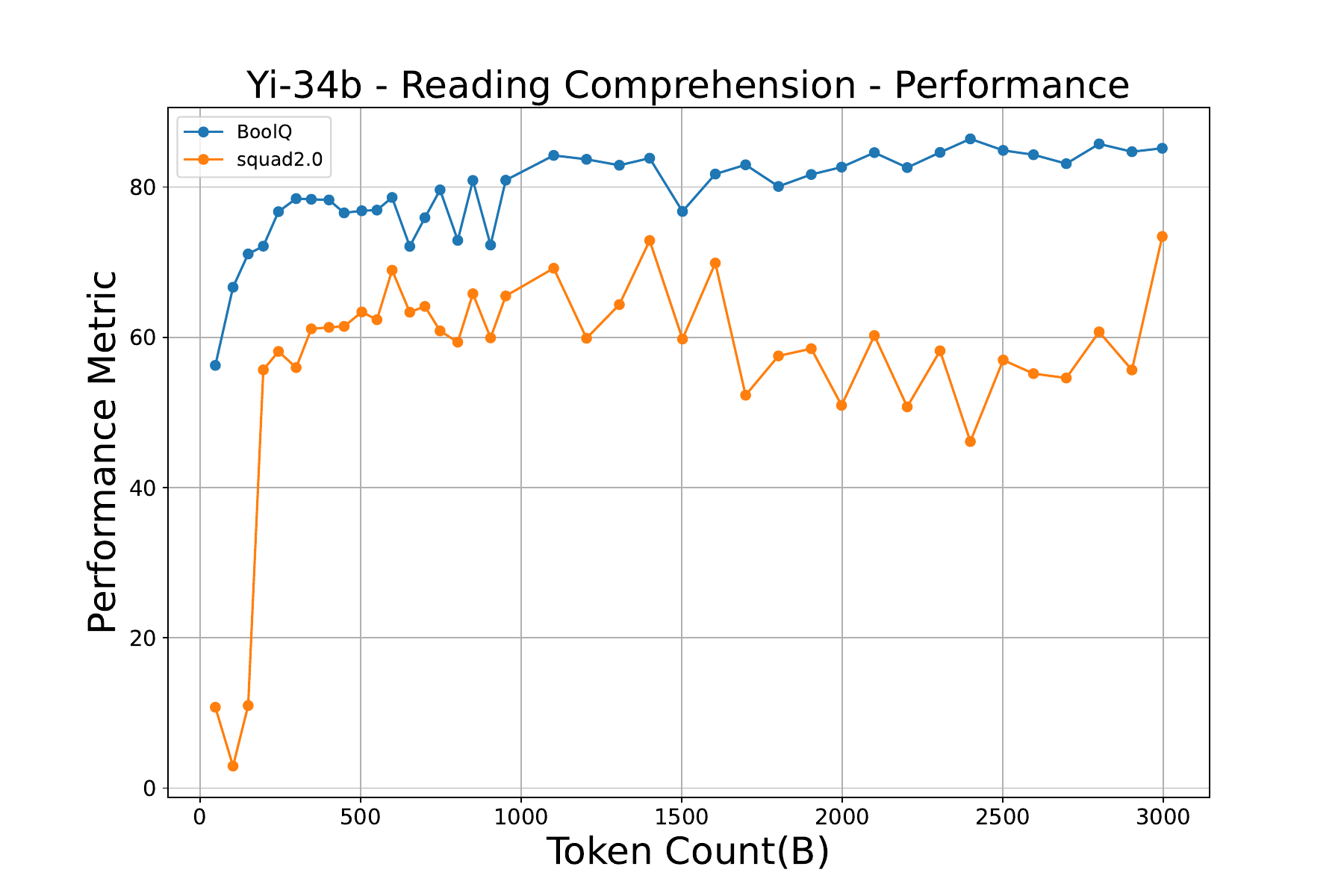}
        \caption{Yi-34B}
        \label{fig:yi34b-Read-Benchmarks}
    \end{subfigure}
    
    \caption{Performance of Baichuan-7B, DeepSeek-7B, DeepSeek-67B, Amber-7B, OpenLLaMA-7B, and Yi-34B models across reading comprehension benchmarks.}
    \label{fig:models-comparison-read-benckmarks}
\end{figure}

\subsection{Experimental results}

Our evaluation experiments are based on the OpenCompass~\citep{2023opencompass}, with some modifications and developments made.

\begin{table}[H]
    \resizebox{\columnwidth}{!}{%
    \begin{tabular}{lllllllllllllll}
        \toprule
Dataset                       & 220B & 440B & 660B & 880B & 1100B & 1320B & 1540B & 1760B & 1980B & 2200B & 2420B \\
\midrule
------- MMLU details ------- & - & - & - & - & - & - & - & - & - & - & - \\
mmlu-humanities              & 25.87 & 36.3 & 51.53 & 51.22 & 53.74 & 56.34 & 56.27 & 59.25 & 58.56 & 59.74 & 59.35 \\
mmlu-stem                    & 24.23 & 31.22 & 39.78 & 40.25 & 41.59 & 42.36 & 43.48 & 43.37 & 44.85 & 45.26 & 45.46 \\
mmlu-social-science          & 24.94 & 39.28 & 52.17 & 54.54 & 57.92 & 58.7 & 59.4 & 61.7 & 62.53 & 63.05 & 63.1 \\
mmlu-other                   & 27.16 & 37.2 & 49.39 & 50.52 & 51.51 & 53.52 & 55.03 & 54.57 & 57.01 & 56.76 & 56.24 \\
mmlu                         & 25.42 & 35.44 & 47.26 & 48.1 & 50.06 & 51.53 & 52.38 & 53.4 & 54.47 & 54.93 & 54.8 \\
---- Standard Benchmarks --- & - & - & - & - & - & - & - & - & - & - & - \\
BoolQ                        & 59.54 & 55.6 & 66.64 & 68.38 & 69.76 & 72.6 & 72.87 & 72.66 & 73.85 & 73.7 & 74.07 \\
piqa                         & 73.56 & 75.08 & 75.73 & 75.03 & 75.9 & 76.12 & 76.22 & 75.68 & 76.22 & 75.9 & 75.9 \\
siqa                         & 34.6 & 37.62 & 45.19 & 47.49 & 46.42 & 48.57 & 50.56 & 50.77 & 51.69 & 50.2 & 50.05 \\
hellaswag                    & 61.19 & 65.62 & 66.1 & 65.77 & 66.07 & 66.98 & 67.07 & 68.05 & 69.01 & 69.62 & 69.71 \\
winogrande                   & 61.01 & 62.59 & 60.46 & 61.33 & 61.8 & 61.56 & 61.01 & 62.27 & 62.51 & 63.3 & 63.06 \\
ARC-e                        & 26.81 & 31.04 & 61.2 & 65.96 & 70.9 & 73.72 & 74.07 & 79.01 & 77.95 & 76.01 & 76.01 \\
ARC-c                        & 21.36 & 27.12 & 37.97 & 44.07 & 49.15 & 55.25 & 54.24 & 56.61 & 56.61 & 58.31 & 56.95 \\
openbookqa\_fact             & 28.4 & 42.4 & 46 & 59.8 & 65.4 & 67.2 & 69.2 & 74 & 70.8 & 71.8 & 75.6 \\
commonsense\_qa              & 58.89 & 61.83 & 63.23 & 59.46 & 59.21 & 60.77 & 65.52 & 66.34 & 67.08 & 67.49 & 67.73 \\
mmlu                         & 25.42 & 35.44 & 47.26 & 48.1 & 50.06 & 51.53 & 52.38 & 53.4 & 54.47 & 54.93 & 54.8 \\
------ Code Generation ----- & - & - & - & - & - & - & - & - & - & - & - \\
openai\_humaneval            & 12.2 & 13.41 & 12.8 & 12.8 & 11.59 & 14.02 & 13.41 & 17.07 & 19.51 & 18.29 & 18.9 \\
mbpp                         & 13.6 & 19.4 & 19 & 18.6 & 20 & 23 & 21.8 & 23.8 & 24.2 & 24.8 & 25.8 \\
------ World Knowledge ----- & - & - & - & - & - & - & - & - & - & - & - \\
nq                           & 2.96 & 3.77 & 4.46 & 5.84 & 7.78 & 9.45 & 8.34 & 7.76 & 7.81 & 6.93 & 7.59 \\
triviaqa                     & 14.02 & 23.67 & 24.11 & 28.09 & 27.08 & 29.15 & 28.1 & 29.42 & 21.01 & 34.39 & 34.99 \\
--- Reading Comprehension -- & - & - & - & - & - & - & - & - & - & - & - \\
squad2.0                     & 27.09 & 29.23 & 28.54 & 30.85 & 32.44 & 32.81 & 34.12 & 31.53 & 33.85 & 39.03 & 37.88 \\
---------- Math -----------  & - & - & - & - & - & - & - & - & - & - & - \\
math                         & 2.24 & 3.88 & 2.96 & 3.78 & 4.82 & 5.04 & 4.64 & 5.16 & 5.7 & 6.26 & 6.02 \\
gsm8k                        & 6.97 & 10.39 & 13.8 & 14.86 & 17.21 & 19.86 & 18.42 & 21.23 & 23.12 & 22.29 & 25.32 \\
TheoremQA                    & 0.5 & 0.88 & 0.5 & 0.75 & 1.12 & 0.62 & 1 & 0.88 & 2 & 1.25 & 1.75 \\
--------- Chinese ---------- & - & - & - & - & - & - & - & - & - & - & - \\
ceval                        & 25.99 & 37.98 & 46.62 & 49.51 & 49.12 & 52.21 & 52.91 & 55.6 & 55.49 & 55.99 & 54.98 \\
ceval-stem                   & 21.89 & 31.97 & 39.1 & 43.17 & 41.19 & 45.31 & 44.97 & 48.29 & 43.75 & 45.59 & 46.3 \\
ceval-social-science         & 28.1 & 47.21 & 54.24 & 57.7 & 58.68 & 61.29 & 64.94 & 63.48 & 70.13 & 67.43 & 67.24 \\
ceval-humanities             & 28.85 & 39.28 & 51.82 & 53.8 & 53.64 & 56.5 & 56.9 & 60.89 & 62.27 & 64.31 & 59.6 \\
ceval-other                  & 28.68 & 39.21 & 48.17 & 49.33 & 50.35 & 52.21 & 52.41 & 56.43 & 56.73 & 56.17 & 54.99 \\
ceval-hard                   & 18.97 & 28.98 & 32.58 & 35.13 & 33.54 & 35.93 & 40.29 & 40.9 & 33.9 & 36.69 & 36.92 \\
cmmlu                        & 25.64 & 36.16 & 46.88 & 48.57 & 50.39 & 51.82 & 52.62 & 55.33 & 55.92 & 56.21 & 57.29 \\
cmmlu-humanities             & 24.67 & 37.88 & 49.89 & 52.11 & 53.66 & 55.44 & 56.64 & 59.06 & 59.49 & 60.16 & 61.58 \\
cmmlu-stem                   & 25.73 & 29.01 & 35.75 & 38.28 & 38.5 & 40.83 & 41.02 & 44.09 & 44.12 & 43.8 & 44.19 \\
cmmlu-social-science         & 25.42 & 38.79 & 51.39 & 52.82 & 55.29 & 55.97 & 57.29 & 59.27 & 60.67 & 60.69 & 62.17 \\
cmmlu-other                  & 26.7 & 38.9 & 50.27 & 50.92 & 53.85 & 55.05 & 55.45 & 59.05 & 59.22 & 60.29 & 61.28 \\
cmmlu-china-specific         & 25.32 & 37.51 & 47.9 & 49.68 & 52.06 & 52.76 & 53.4 & 56.44 & 56.99 & 57.23 & 59.07 \\

        \bottomrule
    \end{tabular}%
    }
    \caption{Summary of Baichuan2-7B checkpoints.}
    \label{tab:Baichuan2-7B summary}
\end{table}

\begin{table}[t]
    \resizebox{\columnwidth}{!}{%
    \begin{tabular}{lllllllllllllll}
        \toprule
Dataset                       & 200B & 400B & 600B & 800B & 1000B & 1200B & 1400B & 1600B & 1800B & 2000B \\
\midrule
------- MMLU details ------- & - & - & - & - & - & - & - & - & - & - \\
mmlu-humanities              & 26.79 & 28.32 & 27.75 & 30.63 & 38.38 & 41.94 & 42.68 & 44.73 & 51.78 & 53.53 \\
mmlu-stem                    & 24.47 & 29.12 & 26.65 & 28.21 & 35.94 & 34.35 & 34.31 & 36.65 & 40.62 & 40.31 \\
mmlu-social-science          & 24.24 & 29.14 & 28.88 & 31.64 & 41.77 & 45.72 & 45.61 & 46.12 & 54.13 & 56.7 \\
mmlu-other                   & 25.84 & 25.79 & 29.04 & 31.01 & 39.95 & 40.99 & 43.37 & 43.16 & 49.13 & 51.35 \\
mmlu                         & 25.27 & 28.18 & 27.91 & 30.12 & 38.64 & 39.99 & 40.66 & 41.97 & 47.95 & 49.29 \\
---- Standard Benchmarks --- & - & - & - & - & - & - & - & - & - & - \\
BoolQ                        & 59.76 & 64.4 & 70.58 & 69.57 & 70.49 & 73.21 & 70.09 & 74.37 & 70.92 & 72.02 \\
piqa                         & 74.86 & 76.12 & 76.5 & 76.28 & 76.99 & 76.66 & 77.09 & 77.58 & 78.07 & 78.73 \\
siqa                         & 33.57 & 33.93 & 33.62 & 37.82 & 39.1 & 37.67 & 46.88 & 48.98 & 49.39 & 48.82 \\
hellaswag                    & 63.74 & 66.92 & 67.95 & 68.32 & 69.04 & 69.66 & 69.53 & 71.08 & 73.02 & 73.43 \\
winogrande                   & 58.41 & 60.14 & 60.69 & 61.64 & 61.72 & 63.61 & 63.61 & 63.38 & 64.72 & 66.46 \\
ARC-e                        & 28.92 & 25.22 & 26.63 & 29.98 & 37.74 & 35.1 & 52.56 & 55.38 & 68.43 & 68.43 \\
ARC-c                        & 26.78 & 27.46 & 23.73 & 28.81 & 32.54 & 29.15 & 33.9 & 36.95 & 45.42 & 47.8 \\
openbookqa\_fact             & 23.8 & 22.2 & 29.6 & 32.0 & 36.6 & 45.2 & 57.4 & 57.4 & 64.4 & 63.8 \\
commonsense\_qa              & 59.71 & 62.57 & 66.01 & 66.75 & 61.75 & 62.41 & 63.31 & 63.96 & 69.37 & 69.45 \\
mmlu                         & 25.27 & 28.18 & 27.91 & 30.12 & 38.64 & 39.99 & 40.66 & 41.97 & 47.95 & 49.29 \\
------ Code Generation ----- & - & - & - & - & - & - & - & - & - & - \\
openai\_humaneval            & 11.59 & 13.41 & 17.07 & 18.29 & 16.46 & 18.29 & 17.07 & 24.39 & 18.9 & 25.0 \\
mbpp                         & 16.6 & 20.2 & 24.0 & 22.6 & 23 & 26.2 & 24.6 & 30.2 & 33.6 & 36.8 \\
------ World Knowledge ----- & - & - & - & - & - & - & - & - & - & - \\
nq                           & 5.48 & 8.67 & 9.06 & 4.93 & 6.26 & 10.97 & 7.48 & 8.34 & 10.25 & 6.93 \\
triviaqa                     & 23.74 & 32.84 & 34.34 & 33.21 & 32.76 & 35.78 & 37.06 & 39.49 & 45.32 & 43.02 \\
--- Reading Comprehension -- & - & - & - & - & - & - & - & - & - & - \\
squad2.0                     & 27.55 & 37.83 & 31.83 & 29.26 & 31.7 & 33.13 & 36.97 & 30.44 & 36.75 & 35.8 \\
---------- Math -----------  & - & - & - & - & - & - & - & - & - & - \\
math                         & 1.74 & 1.62 & 1.64 & 2.54 & 2.34 & 2.58 & 2.74 & 2.76 & 3.6 & 4.18 \\
gsm8k                        & 2.65 & 3.18 & 8.04 & 7.88 & 9.7 & 11.75 & 11.75 & 13.19 & 16.38 & 18.12 \\
TheoremQA                    & 0.12 & 0.75 & 0.5 & 1.0 & 0.88 & 1.12 & 0.75 & 0.75 & 1.25 & 0.62 \\
--------- Chinese ---------- & - & - & - & - & - & - & - & - & - & - \\
ceval                        & 23.24 & 26.45 & 27.89 & 28.36 & 31.8 & 33.76 & 37.75 & 40.49 & 42.38 & 45.1 \\
ceval-stem                   & 23.28 & 24.48 & 25.81 & 26.66 & 28.85 & 29.99 & 32.97 & 35.72 & 36.69 & 38.81 \\
ceval-social-science         & 25.08 & 26.58 & 27.53 & 28.27 & 36.1 & 39.77 & 42.18 & 48.05 & 49.33 & 56.5 \\
ceval-humanities             & 23.0 & 26.47 & 29.21 & 26.9 & 30.76 & 30.85 & 41.79 & 44.41 & 47.6 & 47.58 \\
ceval-other                  & 21.73 & 29.9 & 30.7 & 32.98 & 34.28 & 38.03 & 38.37 & 38.38 & 41.17 & 43.67 \\
ceval-hard                   & 24.27 & 24.78 & 26.56 & 24.83 & 28.43 & 27 & 29.53 & 32.27 & 33.37 & 31.94 \\
cmmlu                        & 25.38 & 25.82 & 27.12 & 27.61 & 36.3 & 37.49 & 38.84 & 41.01 & 44.22 & 46.8 \\
cmmlu-humanities             & 25.76 & 27.06 & 26.89 & 28.07 & 39.91 & 39.91 & 41.81 & 45 & 49.54 & 52.83 \\
cmmlu-stem                   & 24.49 & 24.75 & 25.73 & 24.56 & 28.85 & 30.03 & 31.05 & 31.65 & 32.37 & 34.72 \\
cmmlu-social-science         & 25.3 & 26.22 & 27.41 & 28.29 & 38.22 & 40.74 & 42.01 & 44.32 & 48.35 & 50.81 \\
cmmlu-other                  & 26.19 & 25.38 & 28.47 & 29.67 & 38.79 & 39.09 & 40.43 & 43.31 & 46.95 & 49.38 \\
cmmlu-china-specific         & 25.93 & 26.27 & 27.12 & 27.79 & 37.52 & 39.11 & 39.41 & 42.13 & 45.68 & 48.56 \\

        \bottomrule
    \end{tabular}%
    }
    \caption{Summary of DeepSeek-7B checkpoints.}
    \label{tab:DeepSeek-7B summary}
\end{table}

\begin{table}[t]
    \resizebox{\columnwidth}{!}{%
    \begin{tabular}{llllllllllllllllllllllllll}
        \toprule
Dataset                       & 8.39B & 20.97B & 29.36B & 41.94B & 50.33B & 58.72B & 71.30B & 79.69B & 88.08B \\
\midrule
------- MMLU details ------- & - & - & - & - & - & - & - & - & - \\
mmlu-humanities              & 25.54 & 25.99 & 26.21 & 24.47 & 24.66 & 23.43 & 26.25 & 27 & 24.92 \\
mmlu-stem                    & 27.34 & 24.65 & 27.32 & 26.25 & 26.83 & 27.2 & 28.78 & 27.03 & 25.62 \\
mmlu-social-science          & 28.17 & 22.49 & 22.88 & 26.57 & 28.25 & 28.82 & 24.98 & 27.5 & 23.24 \\
mmlu-other                   & 27.55 & 25.21 & 26.29 & 27.16 & 28.17 & 26.99 & 23.96 & 28.58 & 23.85 \\
mmlu                         & 27.15 & 24.63 & 25.9 & 26.12 & 26.94 & 26.63 & 26.31 & 27.48 & 24.55 \\
---- Standard Benchmarks --- & - & - & - & - & - & - & - & - & - \\
BoolQ                        & 52.08 & 57.8 & 60.21 & 59.24 & 53.18 & 60.03 & 59.72 & 58.26 & 62.63 \\
piqa                         & 66.16 & 69.97 & 70.67 & 72.36 & 71.82 & 71.71 & 73.34 & 73.56 & 73.34 \\
siqa                         & 33.83 & 33.88 & 33.21 & 33.21 & 34.95 & 32.55 & 33.67 & 33.62 & 34.39 \\
hellaswag                    & 37.95 & 47.8 & 51.64 & 54.69 & 55.8 & 56.93 & 58.68 & 59.2 & 59.61 \\
winogrande                   & 52.25 & 52.17 & 52.72 & 53.75 & 54.06 & 53.75 & 55.56 & 55.33 & 56.35 \\
ARC-e                        & 25.93 & 27.69 & 27.87 & 24.16 & 22.05 & 23.1 & 24.34 & 26.98 & 26.63 \\
ARC-c                        & 28.81 & 24.41 & 20.34 & 23.05 & 26.1 & 28.47 & 29.49 & 28.47 & 24.41 \\
openbookqa\_fact             & 24.2 & 24.6 & 25.4 & 22.8 & 25.8 & 20.4 & 25 & 23 & 24.4 \\
commonsense\_qa              & 36.77 & 45.21 & 49.39 & 49.63 & 53.73 & 54.14 & 55.77 & 55.86 & 54.3 \\
mmlu                         & 27.15 & 24.63 & 25.9 & 26.12 & 26.94 & 26.63 & 26.31 & 27.48 & 24.55 \\
------ Code Generation ----- & - & - & - & - & - & - & - & - & - \\
openai\_humaneval            & 1.83 & 6.1 & 4.27 & 4.88 & 6.71 & 6.71 & 7.93 & 9.15 & 7.93 \\
mbpp                         & 0.6 & 2.4 & 4.8 & 6.8 & 8 & 7.2 & 8.6 & 11.2 & 12 \\
------ World Knowledge ----- & - & - & - & - & - & - & - & - & - \\
nq                           & 0.28 & 1.05 & 1.27 & 1.5 & 1.63 & 1.19 & 3.66 & 2.16 & 2.13 \\
triviaqa                     & 0.75 & 2.21 & 3.02 & 5.76 & 5.08 & 5.7 & 13.13 & 10.25 & 8.5 \\
--- Reading Comprehension -- & - & - & - & - & - & - & - & - & - \\
squad2.0                     & 4.4 & 9.47 & 15.97 & 18.48 & 16.37 & 13.59 & 30.94 & 23.09 & 21.44 \\
---------- Math -----------  & - & - & - & - & - & - & - & - & - \\
math                         & 1.48 & 1.46 & 1.18 & 1.38 & 1.2 & 0.88 & 1.3 & 1.7 & 1.18 \\
gsm8k                        & 1.14 & 1.67 & 0.83 & 1.36 & 1.06 & 1.44 & 1.9 & 1.06 & 1.36 \\
TheoremQA                    & 0 & 0 & 0.12 & 0.12 & 0 & 0 & 0.12 & 0.12 & 0.25 \\
--------- Chinese ---------- & - & - & - & - & - & - & - & - & - \\
ceval                        & 23.96 & 23.96 & 24.26 & 28.47 & 23.25 & 25.37 & 24.03 & 25.88 & 25.58 \\
ceval-stem                   & 23.22 & 23.5 & 24.09 & 29.47 & 24.28 & 28.06 & 26.45 & 26.26 & 28.19 \\
ceval-social-science         & 23.06 & 25.11 & 26 & 31.05 & 27.13 & 25.73 & 21.34 & 29.88 & 27.02 \\
ceval-humanities             & 26.99 & 25.32 & 23.19 & 28.03 & 19.64 & 21.47 & 20.3 & 23.46 & 22.62 \\
ceval-other                  & 23.09 & 22.41 & 24.05 & 24.76 & 21.45 & 24.05 & 25.8 & 23.99 & 22.47 \\
ceval-hard                   & 23.82 & 25.14 & 24.6 & 29.75 & 24.78 & 28.45 & 25.67 & 27 & 27.73 \\
cmmlu                        & 25.78 & 24.83 & 25.04 & 24.86 & 25.56 & 25.82 & 24.96 & 24.82 & 25.25 \\
cmmlu-humanities             & 25.17 & 24.76 & 24.96 & 24.91 & 25.39 & 26.24 & 23.94 & 24.86 & 24 \\
cmmlu-stem                   & 25.67 & 24.3 & 24.2 & 24.6 & 24.38 & 25.71 & 24.35 & 24.8 & 25.76 \\
cmmlu-social-science         & 26.49 & 25.55 & 25.61 & 24.99 & 26.22 & 25.39 & 25.55 & 24.71 & 25.85 \\
cmmlu-other                  & 25.41 & 24.41 & 25.23 & 24.93 & 26.08 & 26.19 & 25.66 & 24.99 & 24.87 \\
cmmlu-china-specific         & 25.89 & 25.22 & 25.72 & 25.53 & 25.67 & 25.45 & 26.55 & 24.6 & 24.7 \\

        \bottomrule
    \end{tabular}%
    }
    \caption{Summary of Amber-7B checkpoints for the first 100 billion tokens.}
    \label{tab:Amber-7B summary before 100}
\end{table}

\begin{table}[t]
    \resizebox{\columnwidth}{!}{%
    \begin{tabular}{llllllllllllllllllllllllll}
        \toprule
Dataset                       & 100.66B & 201.33B & 301.99B & 398.46B & 499.12B & 599.79B & 700.45B & 801.11B & 901.78B & 998.24B & 1098.91B & 1199.57B & 1300.23B \\
\midrule
------- MMLU details ------- & - & - & - & - & - & - & - & - & - & - & - & - & - \\
mmlu-humanities              & 24.51 & 25.5 & 25.38 & 25.06 & 26.09 & 26.55 & 26.79 & 27.88 & 29.95 & 34.37 & 35 & 41.63 & 42.17 \\
mmlu-stem                    & 24.09 & 24.94 & 28.77 & 26.54 & 24.69 & 23.53 & 28.12 & 29.24 & 29.21 & 33.38 & 35.39 & 35.84 & 36 \\
mmlu-social-science          & 24.7 & 24.06 & 29.91 & 24.06 & 22.6 & 24.07 & 26.9 & 31.76 & 30.8 & 36.87 & 43.21 & 45.41 & 47.09 \\
mmlu-other                   & 26.5 & 28.35 & 29.25 & 26.69 & 25.52 & 27.34 & 27.04 & 29.18 & 32.95 & 37.43 & 41.42 & 44.93 & 45.95 \\
mmlu                         & 24.86 & 25.66 & 28.35 & 25.71 & 24.76 & 25.2 & 27.31 & 29.45 & 30.56 & 35.26 & 38.32 & 41.25 & 42.01 \\
---- Standard Benchmarks --- & - & - & - & - & - & - & - & - & - & - & - & - & - \\
BoolQ                        & 58.87 & 63.61 & 60.12 & 64.65 & 68.41 & 67.65 & 69.54 & 66.54 & 73.12 & 69.51 & 73.58 & 73.06 & 73.67 \\
piqa                         & 73.23 & 73.99 & 75.57 & 76.39 & 76.99 & 76.28 & 77.69 & 76.22 & 77.53 & 76.82 & 78.07 & 78.02 & 78.89 \\
siqa                         & 35.62 & 33.42 & 33.01 & 33.47 & 34.85 & 36.9 & 35.11 & 35.62 & 36.54 & 36.54 & 38.74 & 43.45 & 47.8 \\
hellaswag                    & 61.2 & 63.93 & 65.64 & 66.02 & 67.47 & 68.16 & 68.89 & 69.29 & 70.3 & 70.24 & 70.91 & 72.12 & 72.98 \\
winogrande                   & 55.49 & 57.22 & 58.09 & 58.72 & 58.88 & 59.91 & 59.27 & 60.54 & 59.98 & 60.62 & 60.54 & 60.62 & 61.4 \\
ARC-e                        & 26.81 & 29.98 & 27.16 & 23.28 & 24.16 & 28.75 & 26.46 & 28.75 & 27.69 & 33.51 & 32.8 & 44.97 & 49.38 \\
ARC-c                        & 28.14 & 27.12 & 29.15 & 29.83 & 29.83 & 30.17 & 27.8 & 28.14 & 29.15 & 28.14 & 31.19 & 35.59 & 33.22 \\
openbookqa\_fact             & 24.6 & 22 & 23 & 24.8 & 21.8 & 31 & 25 & 30.8 & 30.6 & 30.8 & 43.6 & 49 & 53.8 \\
commonsense\_qa              & 56.27 & 58.23 & 60.03 & 60.52 & 60.03 & 62.33 & 63.39 & 62.98 & 63.06 & 64.13 & 61.59 & 64.29 & 64.13 \\
mmlu                         & 24.86 & 25.66 & 28.35 & 25.71 & 24.76 & 25.2 & 27.31 & 29.45 & 30.56 & 35.26 & 38.32 & 41.25 & 42.01 \\
------ Code Generation ----- & - & - & - & - & - & - & - & - & - & - & - & - & - \\
openai\_humaneval            & 10.37 & 7.93 & 7.93 & 9.15 & 12.2 & 14.63 & 11.59 & 11.59 & 15.24 & 15.24 & 14.63 & 19.51 & 18.9 \\
mbpp                         & 9.6 & 15.4 & 19 & 15.6 & 15.4 & 19 & 21 & 20.8 & 22.6 & 25.2 & 25.2 & 27.4 & 26 \\
------ World Knowledge ----- & - & - & - & - & - & - & - & - & - & - & - & - & - \\
nq                           & 3.8 & 5.51 & 3.05 & 5.21 & 2.77 & 2.55 & 2.74 & 8.01 & 8.53 & 8.95 & 5.68 & 8.86 & 5.65 \\
triviaqa                     & 14.15 & 20.1 & 12.7 & 20.88 & 18.25 & 17.64 & 20.55 & 27.95 & 31.23 & 35.52 & 27.1 & 29.87 & 30.58 \\
--- Reading Comprehension -- & - & - & - & - & - & - & - & - & - & - & - & - & - \\
squad2.0                     & 20.63 & 22.4 & 16.62 & 15.51 & 17.99 & 21.03 & 24.05 & 23.88 & 23.24 & 28.52 & 28.46 & 25.81 & 28.36 \\
---------- Math -----------  & - & - & - & - & - & - & - & - & - & - & - & - & - \\
math                         & 1.34 & 1.98 & 1.76 & 2.02 & 1.88 & 1.8 & 1.62 & 1.48 & 2.04 & 1.92 & 1.82 & 2.48 & 2.86 \\
gsm8k                        & 2.12 & 2.2 & 3.64 & 3.34 & 3.87 & 3.64 & 4.85 & 5.53 & 6.75 & 7.73 & 7.43 & 8.19 & 9.17 \\
TheoremQA                    & 0.25 & 0.25 & 0.25 & 1 & 0.38 & 0.88 & 1 & 0.62 & 0.62 & 0.38 & 1.12 & 1 & 0.88 \\
--------- Chinese ---------- & - & - & - & - & - & - & - & - & - & - & - & - & - \\
ceval                        & 21.29 & 26.44 & 25.1 & 21.77 & 23.96 & 27.54 & 26.31 & 28.68 & 25.8 & 29.71 & 30.72 & 30.4 & 32.42 \\
ceval-stem                   & 23.77 & 27.8 & 28.02 & 19.53 & 23.21 & 25.58 & 24.04 & 28.02 & 25.46 & 28.11 & 29.89 & 27.68 & 31.43 \\
ceval-social-science         & 22.41 & 24.52 & 21.71 & 23.16 & 24.36 & 33.17 & 27.15 & 27.18 & 24.58 & 29.78 & 32.71 & 32.59 & 36.38 \\
ceval-humanities             & 16.76 & 24.35 & 20.9 & 23.58 & 25.61 & 25.99 & 27.3 & 25.54 & 24.4 & 27 & 28.18 & 29.62 & 28.14 \\
ceval-other                  & 20.29 & 27.82 & 27.09 & 22.79 & 23.31 & 27.56 & 28.71 & 34.39 & 28.93 & 35.25 & 32.96 & 34.13 & 34.89 \\
ceval-hard                   & 20.15 & 27.89 & 26.5 & 18.41 & 23.82 & 26.34 & 21.51 & 27.58 & 23.68 & 28.5 & 27 & 26.71 & 28.15 \\
cmmlu                        & 25.3 & 25.3 & 25.6 & 24.98 & 25.74 & 25.37 & 25.81 & 25.55 & 26.53 & 27.22 & 29.83 & 30.47 & 30.64 \\
cmmlu-humanities             & 24.86 & 26.01 & 24.96 & 25.15 & 25.47 & 26.23 & 25.6 & 26.44 & 25.42 & 27.49 & 30.54 & 30.91 & 30.84 \\
cmmlu-stem                   & 25.69 & 25.31 & 25.74 & 25.16 & 26.53 & 25.36 & 25.26 & 25.28 & 26.34 & 25.47 & 27.21 & 27.96 & 28.77 \\
cmmlu-social-science         & 25.32 & 24.69 & 26.08 & 24.44 & 25.62 & 25.29 & 26.46 & 25.46 & 27.01 & 26.66 & 30.07 & 31.47 & 31.57 \\
cmmlu-other                  & 25.19 & 25.57 & 25.3 & 25.42 & 25.26 & 24.72 & 25.65 & 25.2 & 26.98 & 29.78 & 31.84 & 31.47 & 31.22 \\
cmmlu-china-specific         & 25.59 & 24.36 & 26.44 & 24.44 & 25.35 & 24.78 & 25.76 & 25.95 & 26.2 & 27.99 & 28.92 & 28.78 & 28.35 \\

        \bottomrule
    \end{tabular}%
    }
    \caption{Summary of Amber-7B checkpoints after the first 100 billion tokens.}
    \label{tab:Amber-7B summary after 100}
\end{table}

\begin{table}[t]
    \resizebox{\columnwidth}{!}{%
    \begin{tabular}{llllllllllllllllllllllll}
        \toprule
Dataset                       & 50B & 100B & 150B & 200B & 250B & 300B & 350B & 400B & 450B & 500B & 550B & 600B & 650B & 700B & 750B & 800B & 850B & 900B & 950B & 1000B \\
\midrule
------- MMLU details ------- & - & - & - & - & - & - & - & - & - & - & - & - & - & - & - & - & - & - & - & - \\
mmlu-humanities              & 26.18 & 24.97 & 25.14 & 24.65 & 25.42 & 25.54 & 25.05 & 26.1 & 26.4 & 28.78 & 27.86 & 33.09 & 34.83 & 36.41 & 37.15 & 38.49 & 40.36 & 39 & 39.74 & 39.79 \\
mmlu-stem                    & 25.94 & 29.34 & 25.25 & 28.86 & 26.23 & 29.65 & 25.66 & 23.95 & 27.5 & 27.62 & 29.78 & 29.34 & 32.6 & 32.03 & 33.74 & 34.51 & 35.81 & 37.99 & 35.93 & 35.52 \\
mmlu-social-science          & 24.89 & 28.34 & 22.23 & 30.22 & 23.27 & 28.01 & 23.89 & 24.21 & 29.9 & 29.9 & 32.21 & 35.39 & 42.38 & 43.55 & 43.09 & 45.99 & 47.42 & 48.85 & 49.63 & 49.37 \\
mmlu-other                   & 26.55 & 26.03 & 24.52 & 27.64 & 26.17 & 30.99 & 27.45 & 27.71 & 31.57 & 31.41 & 33.44 & 38.2 & 43.03 & 44.7 & 46.15 & 47.6 & 48.6 & 48.92 & 50.08 & 49.89 \\
mmlu                         & 25.93 & 26.93 & 24.39 & 27.48 & 25.29 & 28.21 & 25.46 & 25.56 & 28.56 & 29.35 & 30.48 & 33.88 & 37.8 & 38.82 & 39.68 & 41.25 & 42.71 & 43.13 & 43.34 & 43.16 \\
---- Standard Benchmarks --- & - & - & - & - & - & - & - & - & - & - & - & - & - & - & - & - & - & - & - & - \\
BoolQ                        & 61.68 & 63.88 & 64.53 & 65.35 & 67.74 & 64.89 & 67.95 & 70.06 & 70.82 & 70.89 & 69.82 & 71.38 & 70.98 & 70.67 & 72.38 & 71.99 & 73.12 & 75.08 & 74.56 & 74.53 \\
piqa                         & 74.05 & 76.06 & 76.93 & 77.69 & 77.64 & 78.02 & 78.56 & 78.29 & 77.91 & 79.81 & 78.45 & 78.67 & 79.27 & 79.16 & 80.41 & 79.65 & 80.03 & 79.87 & 80.25 & 80.2 \\
siqa                         & 33.32 & 33.21 & 34.13 & 32.85 & 33.21 & 33.11 & 34.34 & 33.62 & 34.75 & 34.44 & 35.57 & 36.28 & 35.31 & 38.23 & 37.2 & 38.38 & 39.46 & 38.22 & 38.48 & 38.79 \\
hellaswag                    & 57.01 & 63.64 & 65.48 & 67.41 & 68.09 & 70.1 & 69.65 & 70.84 & 70.88 & 72.07 & 72.19 & 72.75 & 73.02 & 73.77 & 74.29 & 74.47 & 74.91 & 75.15 & 75.11 & 75.15 \\
winogrande                   & 57.38 & 59.19 & 60.54 & 60.54 & 63.69 & 64.72 & 64.17 & 66.14 & 64.4 & 66.7 & 66.46 & 67.01 & 67.32 & 67.17 & 68.74 & 67.96 & 67.01 & 67.95 & 67.64 & 67.32 \\
ARC-e                        & 53.93 & 58.29 & 60.77 & 62.03 & 63.17 & 63.55 & 64.02 & 64.35 & 65.91 & 64.77 & 67.68 & 66.79 & 64.69 & 67.08 & 68.18 & 69.4 & 68.81 & 69.91 & 69.4 & 69.53 \\
ARC-c                        & 27.82 & 32.59 & 34.3 & 35.32 & 34.56 & 36.6 & 35.75 & 38.31 & 38.65 & 37.2 & 38.48 & 37.63 & 38.39 & 38.31 & 39.76 & 39.93 & 39.93 & 41.04 & 42.32 & 42.24 \\
openbookqa\_fact             & 28.6 & 22.2 & 25.4 & 20 & 28 & 23 & 27.8 & 28.8 & 28.8 & 27.8 & 37.4 & 40.6 & 44.6 & 51.2 & 50.4 & 53 & 58.4 & 53.8 & 55.6 & 56 \\
commonsense\_qa              & 18.92 & 19.57 & 19.74 & 19.58 & 19.57 & 20.31 & 21.38 & 19.9 & - & 22.19 & 24.24 & 35.54 & 43.73 & 45.13 & 43.41 & 46.76 & 52.09 & 54.14 & 55.45 & 55.61 \\
mmlu                         & 25.93 & 26.93 & 24.39 & 27.48 & 25.29 & 28.21 & 25.46 & 25.56 & 28.56 & 29.35 & 30.48 & 33.88 & 37.8 & 38.82 & 39.68 & 41.25 & 42.71 & 43.13 & 43.34 & 43.16 \\
------ Code Generation ----- & - & - & - & - & - & - & - & - & - & - & - & - & - & - & - & - & - & - & - & - \\
openai\_humaneval            & 7.92 & 8.54 & 11.58 & 11.58 & 13.41 & 12.8 & 14.02 & 15.24 & 15.24 & 15.24 & 18.29 & 15.85 & 17.07 & 20.12 & 18.9 & 18.29 & 19.51 & 20.12 & 18.9 & 19.51 \\
mbpp                         & 0.93 & 10.05 & 12.57 & 14.68 & 18.39 & 21.69 & 21.56 & 25.4 & 26.85 & 27.25 & 25 & 30.03 & 30.16 & 32.01 & 29.76 & 33.73 & 31.75 & 34.52 & 35.45 & 35.05 \\
------ World Knowledge ----- & - & - & - & - & - & - & - & - & - & - & - & - & - & - & - & - & - & - & - & - \\
nq                           & 5.18 & 8.35 & 10.35 & 9.7 & 9.88 & 11.35 & 12.22 & 13.55 & 13.22 & 14.06 & 14.62 & 14.17 & 14.99 & 15.67 & 15.74 & 16.83 & 17.28 & 17.56 & 17.13 & 17.16 \\
triviaqa                     & 23.2 & 34.28 & 39.07 & 42.16 & 44.48 & 46.83 & 48.33 & 49.34 & 50.72 & 52.08 & 54.25 & 54.51 & 57.05 & 57.99 & 58.81 & 60.09 & 60.85 & 61.35 & 62.04 & 62.03 \\
--- Reading Comprehension -- & - & - & - & - & - & - & - & - & - & - & - & - & - & - & - & - & - & - & - & - \\
squad2.0                     & 52.69 & 57.76 & 62.47 & 65.9 & 59.07 & 62.82 & 63.02 & 65.73 & 63.74 & 65.53 & 63.98 & 63.2 & 61.99 & 62.01 & 60.94 & 64.2 & 65.53 & 64.32 & 63.48 & 63.45 \\
---------- Math -----------  & - & - & - & - & - & - & - & - & - & - & - & - & - & - & - & - & - & - & - & - \\
math                         & 0.54 & 0.6 & 1.02 & 0.84 & 0.76 & 0.8 & 0.88 & 0.96 & 1.44 & 1.34 & 1.32 & 1.9 & 1.78 & 1.54 & 1.84 & 1.66 & 1.96 & 2.02 & 2.06 & 1.98 \\
gsm8k                        & 1.21 & 2.35 & 1.82 & 2.65 & 3.18 & 2.58 & 4.25 & 5.08 & 4.55 & 5.84 & 6.06 & 4.93 & 6.9 & 10.01 & 8.41 & 9.1 & 9.48 & 10.01 & 11.3 & 10.77 \\
TheoremQA                    & - & - & - & - & - & - & - & - & - & - & - & - & - & - & - & - & - & - & - & - \\
--------- Chinese ---------- & - & - & - & - & - & - & - & - & - & - & - & - & - & - & - & - & - & - & - & - \\
ceval                        & - & - & - & - & - & - & - & - & - & - & - & - & - & - & - & - & - & - & - & - \\
ceval-stem                   & - & - & - & - & - & - & - & - & - & - & - & - & - & - & - & - & - & - & - & - \\
ceval-social-science         & - & - & - & - & - & - & - & - & - & - & - & - & - & - & - & - & - & - & - & - \\
ceval-humanities             & - & - & - & - & - & - & - & - & - & - & - & - & - & - & - & - & - & - & - & - \\
ceval-other                  & - & - & - & - & - & - & - & - & - & - & - & - & - & - & - & - & - & - & - & - \\
ceval-hard                   & - & - & - & - & - & - & - & - & - & - & - & - & - & - & - & - & - & - & - & - \\
cmmlu                        & 24.77 & 24.75 & 24.62 & 25.38 & 25.34 & 25.3 & 24.97 & 26.16 & 25.31 & 25.76 & 27.02 & 29.91 & 29.83 & 31.13 & 31.5 & 32.48 & 33.12 & 33.02 & 33.75 & 33.96 \\
cmmlu-humanities             & 24.66 & 24.79 & 25.19 & 25.11 & 25.75 & 25.35 & 25.19 & 26.68 & 25.67 & 24.71 & 27.52 & 30.77 & 28.85 & 31.14 & 31.26 & 32.26 & 33.14 & 33.39 & 33.79 & 34.27 \\
cmmlu-stem                   & 24.33 & 23.94 & 24.26 & 25.8 & 23.9 & 25.09 & 24.34 & 25.09 & 24.5 & 25.56 & 24.89 & 27.58 & 27.81 & 28.25 & 28.8 & 28.51 & 29.39 & 28.37 & 29.99 & 30.74 \\
cmmlu-social-science         & 25.3 & 24.95 & 24.89 & 25.32 & 25.99 & 25.55 & 25.14 & 26.2 & 25.11 & 25.82 & 26.86 & 29.9 & 29.79 & 30.48 & 31.65 & 32.42 & 33.93 & 33.49 & 34.09 & 34.2 \\
cmmlu-other                  & 24.6 & 25.19 & 24.12 & 25.32 & 25.43 & 25.12 & 25.12 & 26.6 & 25.94 & 26.73 & 28.66 & 31.2 & 32.47 & 34.47 & 33.88 & 32.33 & 35.33 & 36.15 & 36.56 & 36.18 \\
cmmlu-china-specific         & 24.96 & 24.88 & 24.64 & 25.35 & 25.63 & 25.39 & 25.06 & 26.23 & 25.33 & 25.98 & 27.17 & 30.1 & 30.23 & 31.31 & 31.91 & 36.88 & 33.81 & 33.7 & 34.32 & - \\

        \bottomrule
    \end{tabular}%
    }
    \caption{Summary of OpenLLaMA-7B checkpoints.}
    \label{tab:OpenLLaMA-7B summary}
\end{table}

\begin{table}[t]
    \resizebox{\textwidth}{!}{%
    \begin{tabular}{llllllllllllllllllllllllllllllllllllllllll}
        \toprule
Dataset                       & 47.19B & 102.24B & 149.42B & 196.61B & 243.79B & 298.84B & 346.03B & 401.08B & 448.27B & 503.32B & 550.5B & 597.69B & 652.74B & 699.92B & 747.11B & 802.16B & 849.35B & 904.4B & 951.58B \\
\midrule
------- MMLU details ------- & - & - & - & - & - & - & - & - & - & - & - & - & - & - & - & - & - & - & - \\
mmlu-humanities              & 26.82 & 25.57 & 42.06 & 47.55 & 50.76 & 54.08 & 57.7 & 59.32 & 60.83 & 60.82 & 62.83 & 63 & 65.36 & 64.64 & 64.12 & 65.86 & 66 & 65.43 & 66.54 \\
mmlu-stem                    & 26.5 & 23.42 & 32.14 & 36.99 & 38.09 & 39.47 & 44.35 & 43 & 46.92 & 43.33 & 45.78 & 47.54 & 47.16 & 47.37 & 46.93 & 49.11 & 50 & 50.39 & 49.61 \\
mmlu-social-science          & 28.37 & 24.65 & 40.15 & 49.44 & 52.26 & 56.07 & 61.74 & 61.65 & 63.58 & 64.61 & 65.8 & 67.72 & 68.65 & 63.91 & 67.25 & 69.39 & 71.6 & 69.1 & 70.38 \\
mmlu-other                   & 30.39 & 25.65 & 42.16 & 46.74 & 52.39 & 51.91 & 56.38 & 56.84 & 59.29 & 60.78 & 60.33 & 62.02 & 62.4 & 62.43 & 62.31 & 62.67 & 64.37 & 64.25 & 64.04 \\
mmlu                         & 27.86 & 24.68 & 38.37 & 44.24 & 47.22 & 49.13 & 53.8 & 53.81 & 56.42 & 55.78 & 57.2 & 58.62 & 59.31 & 58.23 & 58.64 & 60.29 & 61.48 & 60.92 & 61.13 \\
---- Standard Benchmarks --- & - & - & - & - & - & - & - & - & - & - & - & - & - & - & - & - & - & - & - \\
BoolQ                        & 56.27 & 66.67 & 71.1 & 72.14 & 76.73 & 78.47 & 78.38 & 78.29 & 76.57 & 76.85 & 76.94 & 78.62 & 72.11 & 75.93 & 79.63 & 72.91 & 80.89 & 72.29 & 80.92 \\
piqa                         & 73.01 & 74.86 & 75.52 & 76.82 & 76.82 & 77.37 & 77.86 & 77.26 & 78.4 & 78.35 & 78.29 & 78.07 & 79 & 78.62 & 78.94 & 78.84 & 79.27 & 79.33 & 79.38 \\
siqa                         & 34.14 & 37.51 & 48.21 & 51.79 & 52.81 & 55.89 & 60.18 & 63.82 & 60.59 & 65.66 & 63.05 & 58.6 & 64.33 & 64.79 & 65.76 & 63.77 & 63.72 & 62.9 & 68.37 \\
hellaswag                    & 54.75 & 63.98 & 66.68 & 69.13 & 69.83 & 71.47 & 71.94 & 73.19 & 73.03 & 73.77 & 74.21 & 74.35 & 74.27 & 75.4 & 75.11 & 76.01 & 76.21 & 76.41 & 75.85 \\
winogrande                   & 54.46 & 58.33 & 60.06 & 61.25 & 62.9 & 63.77 & 65.82 & 65.59 & 64.33 & 65.75 & 65.11 & 64.96 & 66.22 & 66.46 & 67.64 & 64.48 & 69.46 & 67.72 & 67.8 \\
ARC-e                        & 26.46 & 35.1 & 46.21 & 62.08 & 67.9 & 71.25 & 69.66 & 74.43 & 73.54 & 73.19 & 79.72 & 79.19 & 80.6 & 83.25 & 83.25 & 84.13 & 83.42 & 85.54 & 85.01 \\
ARC-c                        & 24.75 & 31.86 & 37.29 & 41.36 & 47.8 & 53.22 & 56.95 & 58.31 & 55.25 & 56.27 & 60.68 & 63.05 & 67.12 & 62.71 & 66.44 & 65.76 & 65.42 & 68.47 & 67.12 \\
openbookqa\_fact             & 24 & 36.2 & 52 & 61.6 & 63.6 & 70.8 & 72.8 & 76.4 & 70.4 & 76.2 & 78.6 & 76.6 & 79.4 & 78 & 81.8 & 78 & 81.4 & 80.2 & 78.6 \\
commonsense\_qa              & 50.7 & 59.38 & 63.72 & 65.52 & 65.85 & 69.21 & 69.94 & 70.76 & 70.27 & 71.33 & 70.19 & 71.99 & 70.93 & 71.25 & 72.24 & 72.48 & 71.91 & 71.91 & 73.05 \\
mmlu                         & 27.86 & 24.68 & 38.37 & 44.24 & 47.22 & 49.13 & 53.8 & 53.81 & 56.42 & 55.78 & 57.2 & 58.62 & 59.31 & 58.23 & 58.64 & 60.29 & 61.48 & 60.92 & 61.13 \\
------ Code Generation ----- & - & - & - & - & - & - & - & - & - & - & - & - & - & - & - & - & - & - & - \\
openai\_humaneval            & 6.71 & 7.93 & 10.98 & 9.15 & 9.76 & 12.8 & 10.98 & 14.02 & 13.41 & 15.85 & 14.63 & 13.41 & 14.63 & 13.41 & 15.24 & 14.02 & 18.29 & 19.51 & 17.68 \\
mbpp                         & 4.2 & 9.6 & 11.4 & 14.8 & 14 & 18.6 & 19.8 & 21.8 & 22.4 & 22.2 & 22 & 22 & 24.4 & 25.8 & 25.4 & 24 & 25.2 & 27 & 27.6 \\
------ World Knowledge ----- & - & - & - & - & - & - & - & - & - & - & - & - & - & - & - & - & - & - & - \\
nq                           & 5.48 & 10.33 & 12.77 & 15.1 & 16.26 & 16.62 & 19.11 & 18.42 & 12.66 & 21.61 & 17.81 & 26.7 & 22.66 & 20.06 & 22.63 & 17.78 & 24.18 & 25.21 & 23.38 \\
triviaqa                     & 11.25 & 23.72 & 26.49 & 30.86 & 34.1 & 34.6 & 40.99 & 37.44 & 35.53 & 41.86 & 41.45 & 46.35 & 47.04 & 39.75 & 49.6 & 42.91 & 45.9 & 48.72 & 47.24 \\
--- Reading Comprehension -- & - & - & - & - & - & - & - & - & - & - & - & - & - & - & - & - & - & - & - \\
squad2.0                     & 10.76 & 2.93 & 10.99 & 55.67 & 58.13 & 55.97 & 61.13 & 61.33 & 61.46 & 63.38 & 62.35 & 68.94 & 63.35 & 64.12 & 60.86 & 59.36 & 65.81 & 59.93 & 65.51 \\
---------- Math -----------  & - & - & - & - & - & - & - & - & - & - & - & - & - & - & - & - & - & - & - \\
math                         & 1.38 & 2.14 & 2.2 & 2.32 & 2.78 & 2.78 & 3.68 & 2.94 & 2.84 & 3.16 & 3.54 & 3.34 & 3.04 & 3.9 & 4.58 & 4.22 & 4.46 & 4.52 & 4.56 \\
gsm8k                        & 3.03 & 5.03 & 5.32 & 8.49 & 9.01 & 10.31 & 15.22 & 17.45 & 19.68 & 20.12 & 22.01 & 22.44 & 22.52 & 24.94 & 26.37 & 27.73 & 27.81 & 28.79 & 30.08 \\
TheoremQA                    & 0.75 & 2.38 & 2.25 & 3.12 & 1.88 & 0.88 & 1.88 & 1.38 & 1.25 & 1.12 & 3.75 & 3.75 & 3 & 2.75 & 3 & 1.62 & 3.75 & 3 & 3.12 \\
--------- Chinese ---------- & - & - & - & - & - & - & - & - & - & - & - & - & - & - & - & - & - & - & - \\
ceval                        & 27.07 & 34.23 & 37.42 & 42.56 & 45.29 & 49.04 & 45.07 & 53.45 & 54.92 & 50.24 & 55.06 & 53.87 & 58.13 & 57.47 & 58.02 & 59.8 & 60.76 & 60.65 & 57.8 \\
ceval-stem                   & 26.96 & 31.71 & 30.12 & 35.61 & 36.68 & 38.32 & 34.6 & 45.18 & 45.11 & 41.92 & 45.68 & 45.61 & 50.59 & 47.49 & 48.92 & 50.79 & 52.41 & 49.49 & 47.81 \\
ceval-social-science         & 23.28 & 40.09 & 46.85 & 54.62 & 57.06 & 65.07 & 54.16 & 64.33 & 66.43 & 66.05 & 68.5 & 70.47 & 71.64 & 69.13 & 72.02 & 75.8 & 75.28 & 73.24 & 73.85 \\
ceval-humanities             & 28.33 & 32.66 & 43.26 & 44.33 & 45.94 & 50.97 & 52.86 & 58.24 & 58.92 & 50.72 & 60.7 & 55.72 & 60.46 & 65.06 & 62.57 & 63.72 & 65.48 & 68.75 & 62.51 \\
ceval-other                  & 29.44 & 35.05 & 36.28 & 42.43 & 49.6 & 52.02 & 48.03 & 53.79 & 58.27 & 50.5 & 54.27 & 51.97 & 57.22 & 57.4 & 57.28 & 57.71 & 58.04 & 61.37 & 56.69 \\
ceval-hard                   & 26.6 & 31.73 & 29.6 & 31.56 & 32.56 & 29.18 & 27.28 & 34.8 & 33.7 & 28.91 & 34.67 & 32.62 & 38.41 & 37.42 & 35.92 & 38.2 & 41.58 & 38.69 & 40.31 \\
cmmlu                        & 25.39 & 29.54 & 38.2 & 35.99 & 37.58 & 43.37 & 43.96 & 48.34 & 53.9 & 46.69 & 50.46 & 49.14 & 51.57 & 47.18 & 50.27 & 54.73 & 57.08 & 48.36 & 52.61 \\
cmmlu-humanities             & 25.72 & 31.43 & 41.29 & 35.15 & 40.22 & 46.87 & 45.6 & 52.75 & 56.93 & 51.79 & 54.12 & 52.52 & 53.82 & 50.86 & 53.13 & 57.86 & 60.22 & 50.74 & 56.16 \\
cmmlu-stem                   & 25.54 & 27.2 & 31.13 & 29.63 & 31.45 & 33.63 & 34.33 & 36.16 & 41.71 & 36.1 & 38.39 & 37.73 & 41.08 & 38.07 & 38.14 & 42.81 & 44.27 & 37.66 & 40.25 \\
cmmlu-social-science         & 25.04 & 30.58 & 39.75 & 38.37 & 39.6 & 46.43 & 48.56 & 52.61 & 59.13 & 49.5 & 55.02 & 53.7 & 54.87 & 49.65 & 54.3 & 59 & 62.1 & 52.75 & 56.6 \\
cmmlu-other                  & 25.43 & 29.02 & 41.25 & 40.45 & 39.29 & 46.87 & 46.73 & 52.04 & 57.42 & 50.16 & 54.29 & 52.44 & 56.65 & 50.69 & 55.63 & 59.26 & 61.51 & 51.96 & 57.7 \\
cmmlu-china-specific         & 25.87 & 29.8 & 38.69 & 37.12 & 36.56 & 42.67 & 45.49 & 48.97 & 55.21 & 45.57 & 51.69 & 50.64 & 51.95 & 47.35 & 52.53 & 55.69 & 58.06 & 49.51 & 53.56 \\

        \bottomrule
    \end{tabular}%
    }
    \caption{Summary of Yi-34B summary for the first 1000 billion tokens.}
    \label{tab:Yi-34B summary pre 1000b}
\end{table}

\begin{table}[t]
    \resizebox{\textwidth}{!}{%
    \begin{tabular}{llllllllllllllllllllllllllllllllllllllllll}
        \toprule
Dataset                       & 1101.0B & 1203.24B & 1305.48B & 1399.85B & 1502.09B & 1604.32B & 1698.69B & 1800.93B & 1903.17B & 1997.54B & 2099.77B & 2202.01B & 2304.25B & 2398.62B & 2500.85B & 2595.23B & 2697.46B & 2799.7B & 2901.93B & 2996.31B \\
\midrule
------- MMLU details ------- & - & - & - & - & - & - & - & - & - & - & - & - & - & - & - & - & - & - & - & - \\
mmlu-humanities              & 66.15 & 67.86 & 67.9 & 69.02 & 67.98 & 70.43 & 71.47 & 68.61 & 70.91 & 72.02 & 71.59 & 73.71 & 74.24 & 74.15 & 72.87 & 74.92 & 76.01 & 75.3 & 75.71 & 77.45 \\
mmlu-stem                    & 51.25 & 51.63 & 52.34 & 52.38 & 50.1 & 52.7 & 55.32 & 52.85 & 53.21 & 56.09 & 55.88 & 57.01 & 57.64 & 58.59 & 58.17 & 58.36 & 58.33 & 59.2 & 59.38 & 59.09 \\
mmlu-social-science          & 73.05 & 72.67 & 74.53 & 74.51 & 74.54 & 76.12 & 76.18 & 76.43 & 75.24 & 77.41 & 76.91 & 78.44 & 78.88 & 78.78 & 79.92 & 79.81 & 80.3 & 80.66 & 80.88 & 80.36 \\
mmlu-other                   & 65.93 & 65.52 & 66.63 & 67.88 & 68.27 & 68.54 & 67.69 & 69.76 & 69.72 & 71.07 & 69.62 & 72.78 & 72.3 & 72.7 & 73 & 72.84 & 73.82 & 73.59 & 73.42 & 75.34 \\
mmlu                         & 62.59 & 62.93 & 63.82 & 64.37 & 63.47 & 65.29 & 66.21 & 65.27 & 65.65 & 67.63 & 67.02 & 68.93 & 69.24 & 69.61 & 69.48 & 69.96 & 70.52 & 70.67 & 70.84 & 71.46 \\
---- Standard Benchmarks --- & - & - & - & - & - & - & - & - & - & - & - & - & - & - & - & - & - & - & - & - \\
BoolQ                        & 84.22 & 83.7 & 82.91 & 83.85 & 76.76 & 81.74 & 82.97 & 80.09 & 81.68 & 82.66 & 84.59 & 82.6 & 84.62 & 86.42 & 84.89 & 84.31 & 83.12 & 85.75 & 84.71 & 85.17 \\
piqa                         & 79.27 & 79.65 & 79.6 & 79.87 & 79.76 & 78.94 & 80.36 & 80.03 & 80.36 & 80.69 & 80.85 & 80.96 & 80.9 & 81.66 & 81.45 & 82.15 & 81.28 & 80.96 & 81.5 & 81.45 \\
siqa                         & 68.68 & 69.04 & 70.83 & 66.94 & 69.45 & 69.55 & 70.37 & 69.75 & 71.03 & 72.26 & 68.99 & 74.05 & 71.9 & 71.34 & 72.47 & 73.03 & 71.34 & 72.67 & 72.31 & 72.93 \\
hellaswag                    & 76.64 & 76.99 & 77.44 & 77.48 & 77.99 & 78.5 & 78.39 & 78.1 & 79.41 & 79.6 & 79.63 & 79.93 & 80.35 & 80.46 & 80.87 & 81.02 & 80.96 & 80.97 & 81.09 & 81.55 \\
winogrande                   & 66.54 & 67.01 & 66.93 & 68.43 & 68.67 & 66.93 & 68.35 & 67.01 & 70.32 & 69.69 & 67.88 & 69.85 & 72.77 & 71.9 & 70.88 & 69.69 & 72.93 & 71.11 & 67.32 & 71.35 \\
ARC-e                        & 83.95 & 86.77 & 88.54 & 89.42 & 88.89 & 89.95 & 88.36 & 90.65 & 90.65 & 89.77 & 92.06 & 91.71 & 92.24 & 93.47 & 92.95 & 93.3 & 94 & 92.77 & 93.47 & 92.59 \\
ARC-c                        & 70.85 & 71.86 & 75.59 & 73.22 & 72.2 & 75.25 & 73.22 & 75.25 & 75.93 & 77.63 & 80.68 & 84.07 & 83.05 & 82.37 & 81.69 & 82.03 & 85.42 & 83.39 & 84.07 & 85.42 \\
openbookqa\_fact             & 82.2 & 79.8 & 83.4 & 81.8 & 84.2 & 85 & 85.2 & 83.8 & 85.4 & 85 & 86.4 & 88.2 & 88.2 & 85.8 & 87.6 & 87 & 87 & 87.4 & 86.2 & 88.4 \\
commonsense\_qa              & 71.58 & 71.66 & 72.65 & 71.99 & 73.05 & 75.1 & 75.1 & 75.68 & 75.27 & 74.04 & 76 & 74.94 & 73.71 & 74.86 & 76.33 & 76.49 & 77.07 & 76.33 & 77.07 & 76.25 \\
mmlu                         & 62.59 & 62.93 & 63.82 & 64.37 & 63.47 & 65.29 & 66.21 & 65.27 & 65.65 & 67.63 & 67.02 & 68.93 & 69.24 & 69.61 & 69.48 & 69.96 & 70.52 & 70.67 & 70.84 & 71.46 \\
------ Code Generation ----- & - & - & - & - & - & - & - & - & - & - & - & - & - & - & - & - & - & - & - & - \\
openai\_humaneval            & 20.12 & 16.46 & 18.29 & 21.95 & 20.73 & 17.68 & 19.51 & 22.56 & 20.12 & 25 & 23.78 & 24.39 & 24.39 & 25.61 & 25 & 26.83 & 24.39 & 26.22 & 25 & 22.56 \\
mbpp                         & 28.2 & 27.6 & 28.8 & 30.8 & 31.2 & 31.8 & 33 & 31.6 & 34.8 & 35 & 35 & 36.6 & 36.4 & 39.6 & 36 & 38.4 & 36.2 & 39.6 & 38.8 & 42.2 \\
------ World Knowledge ----- & - & - & - & - & - & - & - & - & - & - & - & - & - & - & - & - & - & - & - & - \\
nq                           & 25.51 & 13.82 & 23.3 & 24.82 & 29.09 & 22.52 & 26.79 & 27.92 & 27.23 & 27.31 & 29.28 & 30.78 & 28.81 & 33.74 & 31.91 & 31.66 & 33.55 & 31.19 & 33.46 & 34.54 \\
triviaqa                     & 54.89 & 36.48 & 49.09 & 54.22 & 55.1 & 53.26 & 52.42 & 56.56 & 54.63 & 57.02 & 57.19 & 57 & 52.21 & 57.95 & 56.93 & 59.15 & 59.45 & 61.89 & 62.11 & 62.32 \\
--- Reading Comprehension -- & - & - & - & - & - & - & - & - & - & - & - & - & - & - & - & - & - & - & - & - \\
squad2.0                     & 69.2 & 59.88 & 64.36 & 72.89 & 59.78 & 69.89 & 52.3 & 57.53 & 58.49 & 50.95 & 60.24 & 50.75 & 58.21 & 46.13 & 56.98 & 55.17 & 54.59 & 60.73 & 55.66 & 73.43 \\
---------- Math -----------  & - & - & - & - & - & - & - & - & - & - & - & - & - & - & - & - & - & - & - & - \\
math                         & 5.74 & 5.68 & 6.42 & 6.34 & 6.8 & 6.8 & 7.64 & 7.1 & 7.78 & 8.22 & 8.86 & 8.96 & 10.16 & 9.58 & 9.82 & 10.86 & 10.82 & 10.42 & 11.7 & 11.72 \\
gsm8k                        & 29.38 & 31.3 & 32.43 & 33.99 & 33.03 & 34.32 & 36.38 & 41.06 & 43.08 & 42.76 & 44 & 44.19 & 45.02 & 45.83 & 45.19 & 46 & 46.91 & 45 & 48.07 & 50.2 \\
TheoremQA                    & 1.62 & 1.5 & 2.62 & 1.62 & 1.62 & 4.62 & 3.75 & 2.62 & 2.25 & 2.62 & 3.62 & 7.38 & 3.75 & 7.75 & 4.12 & 4.12 & 3.5 & 6 & 3 & 3.38 \\
--------- Chinese ---------- & - & - & - & - & - & - & - & - & - & - & - & - & - & - & - & - & - & - & - & - \\
ceval                        & 63.14 & 58.51 & 59.67 & 63.78 & 61.17 & 63.43 & 60.69 & 57.87 & 68.32 & 64.42 & 64.63 & 66.66 & 70.47 & 68.68 & 68.94 & 68.76 & 68.93 & 69.63 & 72.26 & 70.19 \\
ceval-stem                   & 55.07 & 49.65 & 47.8 & 54.03 & 49.49 & 51.51 & 50.15 & 43.77 & 56.2 & 51.26 & 51.84 & 55.44 & 59.12 & 56.96 & 56.2 & 57.29 & 59 & 58.3 & 62.56 & 60.02 \\
ceval-social-science         & 76.73 & 72.36 & 73.5 & 76.53 & 73.11 & 78.8 & 74.63 & 74.8 & 82.26 & 81.89 & 79.68 & 77.8 & 83.56 & 82.74 & 83.02 & 80.76 & 79.66 & 80.19 & 81.48 & 79.35 \\
ceval-humanities             & 68.35 & 64.13 & 68.9 & 71.3 & 71.33 & 71.45 & 67.49 & 66.8 & 78.45 & 75.44 & 74.1 & 76.73 & 79.42 & 78.44 & 76.78 & 78.49 & 78.24 & 80.34 & 81.65 & 80.66 \\
ceval-other                  & 60.24 & 56.41 & 59.45 & 62.41 & 61.4 & 63.11 & 60.39 & 59.17 & 67.58 & 61.45 & 64.7 & 66.85 & 70.25 & 67.44 & 71.45 & 69 & 67.95 & 69.94 & 72.14 & 69.89 \\
ceval-hard                   & 41.99 & 43.77 & 36.69 & 40.04 & 36.11 & 38.53 & 33.06 & 26.37 & 39.8 & 31.43 & 37.01 & 38.66 & 43.6 & 41.69 & 38.39 & 39.42 & 41.95 & 40.73 & 45.9 & 43.86 \\
cmmlu                        & 49.14 & 51.16 & 55.65 & 63.53 & 54.17 & 58.02 & 55.71 & 47.33 & 55.96 & 56.5 & 57.62 & 61.08 & 61.83 & 67.32 & 57.97 & 64.52 & 57.61 & 58.47 & 59.29 & 57.45 \\
cmmlu-humanities             & 53.28 & 54.6 & 60 & 69.04 & 58.29 & 61.18 & 58.92 & 50.88 & 61.37 & 59.99 & 62.04 & 65.38 & 64.24 & 71.23 & 60.37 & 69.23 & 58.76 & 61.06 & 59.89 & 61.74 \\
cmmlu-stem                   & 37.9 & 39.31 & 44.1 & 50.02 & 43.09 & 47.63 & 45.02 & 38.53 & 44.25 & 45.23 & 46.71 & 48.46 & 50.35 & 56.44 & 47.98 & 52.7 & 46.85 & 47.01 & 50.96 & 46.8 \\
cmmlu-social-science         & 51.85 & 56.13 & 57.69 & 67.25 & 58.42 & 61.99 & 58.43 & 49.73 & 58.99 & 59.88 & 60.78 & 64.7 & 65.5 & 70.08 & 61.4 & 67.55 & 62.51 & 61.65 & 63.12 & 60.45 \\
cmmlu-other                  & 54.31 & 54.33 & 61.98 & 68.61 & 56.91 & 61.26 & 61.03 & 50.7 & 60.08 & 61.31 & 61.52 & 66.37 & 67.37 & 72.22 & 62.16 & 69.37 & 61.62 & 64.56 & 62.57 & 61.4 \\
cmmlu-china-specific         & 49.41 & 51.8 & 55.54 & 63.33 & 55.2 & 58.99 & 56.04 & 46.12 & 55.6 & 58.15 & 58.04 & 62.47 & 61.75 & 66.95 & 59.65 & 63.77 & 58.35 & 58.6 & 58.89 & 58.45 \\

        \bottomrule
    \end{tabular}%
    }
    \caption{Summary of Yi-34B checkpoints after the first 1000 billion tokens.}
    \label{tab:Yi-34B summary after 1000b}
\end{table}

\begin{table}[t]
    \resizebox{\columnwidth}{!}{%
    \begin{tabular}{llllllllllllllll}
        \toprule
Dataset                       & 200B & 400B & 600B & 800B & 1000B & 1200B & 1400B & 1600B & 1800B & 2000B \\
\midrule
------- MMLU details ------- & - & - & - & - & - & - & - & - & - & - \\
mmlu-humanities              & 38.08 & 59.46 & 65.73 & 68.31 & 70.17 & 70.23 & 72.01 & 74.2 & 74.6 & 77.11 \\
mmlu-stem                    & 32.17 & 46.19 & 47.95 & 50.8 & 50.64 & 54.35 & 53.76 & 57.02 & 58.64 & 60.55 \\
mmlu-social-science          & 40.87 & 64.11 & 69.91 & 72.63 & 73.26 & 74.23 & 76.28 & 77.05 & 79.71 & 81.67 \\
mmlu-other                   & 39.84 & 58.83 & 63.18 & 64.76 & 65.99 & 67.44 & 67.85 & 70.49 & 72.42 & 74.38 \\
mmlu                         & 37.1 & 55.87 & 60.1 & 63.57 & 63.36 & 65.14 & 65.88 & 68.23 & 69.86 & 71.93 \\
---- Standard Benchmarks --- & - & - & - & - & - & - & - & - & - & - \\
BoolQ                        & 72.69 & 78.29 & 82.35 & 83.67 & 84.04 & 85.63 & 86.27 & 86.54 & 87.4 & 88.23 \\
piqa                         & 78.73 & 78.94 & 79.92 & 81.39 & 81.07 & 80.96 & 80.9 & 81.77 & 82.37 & 82.64 \\
siqa                         & 37.15 & 52.15 & 57.32 & 51.89 & 58.5 & 61.21 & 59.88 & 62.38 & 61.72 & 62.79 \\
hellaswag                    & 71.28 & 75.72 & 77.65 & 78.25 & 78.35 & 78.71 & 79.19 & 80.19 & 81.15 & 82.29 \\
winogrande                   & 64.8 & 68.9 & 70.56 & 72.69 & 72.14 & 72.77 & 74.98 & 73.16 & 75.37 & 76.01 \\
ARC-e                        & 33.33 & 70.19 & 83.25 & 83.77 & 88.01 & 88.71 & 89.95 & 90.48 & 93.83 & 93.65 \\
ARC-c                        & 26.44 & 51.19 & 68.81 & 69.83 & 74.92 & 74.24 & 77.63 & 80 & 84.07 & 86.44 \\
openbookqa\_fact             & 31 & 61 & 74 & 69.4 & 79.6 & 73.8 & 75.6 & 80.4 & 80.8 & 81 \\
commonsense\_qa              & 67.65 & 67.24 & 81.5 & 72.48 & 68.39 & 68.3 & 70.19 & 70.68 & 75.43 & 74.45 \\
mmlu                         & 37.1 & 55.87 & 60.1 & 62.57 & 63.36 & 65.14 & 65.88 & 68.23 & 69.86 & 71.93 \\
------ Code Generation ----- & - & - & - & - & - & - & - & - & - & - \\
openai\_humaneval            & 17.07 & 23.78 & 27.44 & 27.44 & 30.49 & 32.32 & 25.61 & 35.37 & 37.8 & 39.63 \\
mbpp                         & 25.2 & 36.2 & 40.6 & 42.2 & 41.2 & 45.2 & 48 & 50 & 53.4 & 55.4 \\
------ World Knowledge ----- & - & - & - & - & - & - & - & - & - & - \\
nq                           & 15.43 & 21.14 & 20.94 & 21.69 & 25.6 & 25.18 & 27.26 & 24.9 & 26.9 & 29.94 \\
triviaqa                     & 43.61 & 55.37 & 55.29 & 59.68 & 58.69 & 60.45 & 61.56 & 62.87 & 64.84 & 67.39 \\
--- Reading Comprehension -- & - & - & - & - & - & - & - & - & - & - \\
squad2.0                     & 40.39 & 43.58 & 46.77 & 44.5 & 42.34 & 49.15 & 48.43 & 51.69 & 57.64 & 57.02 \\
---------- Math -----------  & - & - & - & - & - & - & - & - & - & - \\
math                         & 2.28 & 4.78 & 7.06 & 7.88 & 8.5 & 9.34 & 10.66 & 11.24 & 14.42 & 15.9 \\
gsm8k                        & 9.17 & 25.55 & 37.6 & 44.58 & 49.51 & 49.81 & 54.59 & 57.01 & 61.56 & 66.57 \\
TheoremQA                    & 1 & 1.75 & 1.38 & 2.25 & 2.25 & 3.38 & 2.12 & 3.12 & 2.5 & 2.62 \\
--------- Chinese ---------- & - & - & - & - & - & - & - & - & - & - \\
ceval                        & 36.52 & 50.03 & 54.45 & 57.94 & 56.93 & 60.78 & 60.78 & 62.39 & 64.89 & 66.95 \\
ceval-stem                   & 30.89 & 42.38 & 44.05 & 49.04 & 46.91 & 50.34 & 51.37 & 50.02 & 52.03 & 56.93 \\
ceval-social-science         & 41.85 & 61 & 64.68 & 67.94 & 70.86 & 75.7 & 72.15 & 74.67 & 79.43 & 79.56 \\
ceval-humanities             & 37.91 & 55.93 & 65.1 & 63.29 & 63.97 & 68.34 & 67.3 & 73.11 & 73.76 & 74.09 \\
ceval-other                  & 40.53 & 48.07 & 53.41 & 59.68 & 55.44 & 58.66 & 61.02 & 62.98 & 66.19 & 66.57 \\
ceval-hard                   & 25.64 & 32.31 & 32.66 & 34.71 & 35.39 & 35.62 & 38.3 & 35.38 & 35.88 & 39.32 \\
cmmlu                        & 32.94 & 52.35 & 58.3 & 60.6 & 62.3 & 62.95 & 64.75 & 66.46 & 68.91 & 70.44 \\
cmmlu-humanities             & 34.9 & 56.88 & 63.84 & 67.72 & 69.79 & 70.11 & 72.53 & 73.97 & 76.95 & 78.52 \\
cmmlu-stem                   & 29.13 & 40.43 & 44.38 & 45.35 & 47.81 & 48.01 & 49.38 & 52.61 & 54.17 & 56.61 \\
cmmlu-social-science         & 33.64 & 56.83 & 63.45 & 65.56 & 66.85 & 67.61 & 69.44 & 70.59 & 73.49 & 73.95 \\
cmmlu-other                  & 34.51 & 55.37 & 61.7 & 64.45 & 65.56 & 66.84 & 68.53 & 69.58 & 71.94 & 73.94 \\
cmmlu-china-specific         & 32.69 & 52.14 & 58.98 & 61.57 & 62.54 & 63.55 & 65.32 & 66.58 & 70.23 & 71.73 \\

        \bottomrule
    \end{tabular}%
    }
    \caption{Summary of DeepSeek-67B checkpoints.}
    \label{tab:DeepSeek-67B summary}
\end{table}

\clearpage

\subsection{Detailed Experiment Prompts}
Here we present the prompts used for each dataset in the evaluation. Since the prompts for datasets (BoolQ, PIQA, SIQA, HellaSwag, WinoGrande, ARC easy and challenge, OpenBookQA, CommonsenseQA, MMLU, CEval, CMMLU), concatenat the question and the answers (options), we have not listed them separately. Below, we provide the prompts used for the other datasets.

\textbf{HumanEval}
\begin{lstlisting}
HUMAN: Complete the following python code:\n{prompt}
\end{lstlisting}

\textbf{MBPP}
\begin{lstlisting}
HUMAN: You are an expert Python programmer, and here is your task: Write a function to find the similar elements from the given two tuple lists. Your code should pass these tests:\n\n assert similar_elements((3, 4, 5, 6),(5, 7, 4, 10)) == (4, 5)\n assert similar_elements((1, 2, 3, 4),(5, 4, 3, 7)) == (3, 4) \n assert similar_elements((11, 12, 14, 13),(17, 15, 14, 13)) == (13, 14) \n
BOT: [BEGIN]\n 'def similar_elements(test_tup1, test_tup2):\r\n  res = tuple(set(test_tup1) & set(test_tup2))\r\n  return (res)' \n[DONE] \n\n
HUMAN: You are an expert Python programmer, and here is your task: Write a python function to identify non-prime numbers. Your code should pass these tests:\n\n assert is_not_prime(2) == False \n assert is_not_prime(10) == True \n assert is_not_prime(35) == True \n
BOT: [BEGIN]\n 'import math\r\ndef is_not_prime(n):\r\n    result = False\r\n    for i in range(2,int(math.sqrt(n)) + 1):\r\n        if n % i == 0:\r\n            result = True\r\n    return result' \n[DONE] \n\n
HUMAN: You are an expert Python programmer, and here is your task: Write a function to find the largest integers from a given list of numbers using heap queue algorithm. Your code should pass these tests:\n\n assert heap_queue_largest( [25, 35, 22, 85, 14, 65, 75, 22, 58],3)==[85, 75, 65] \n assert heap_queue_largest( [25, 35, 22, 85, 14, 65, 75, 22, 58],2)==[85, 75] \n assert heap_queue_largest( [25, 35, 22, 85, 14, 65, 75, 22, 58],5)==[85, 75, 65, 58, 35] \n
BOT: [BEGIN]\n 'import heapq as hq\r\ndef heap_queue_largest(nums,n):\r\n  largest_nums = hq.nlargest(n, nums)\r\n  return largest_nums' \n[DONE] \n\n
HUMAN: You are an expert Python programmer, and here is your task: {text} Your code should pass these tests:\n\n {test_list}  \n
BOT: [BEGIN]\n
\end{lstlisting}

\textbf{NaturalQuestions}

\begin{lstlisting}
HUMAN: Answer these questions, your answer should be as simple as possible, start your answer with the prompt \'The answer is \'.\nQ: {question}?
BOT: A:
\end{lstlisting}

\textbf{triviaqa}
\begin{lstlisting}
HUMAN: Answer these questions, your answer should be as simple as possible, start your answer with the prompt \'The answer is \'.\nQ: {question}?
BOT: A:
\end{lstlisting}

\textbf{SQuAD 2.0}

\begin{lstlisting}
HUMAN: {context}\nAccording to the above passage, answer the following question. If it is impossible to answer according to the passage, answer `impossible to answer`:\nQuestion: {question}
BOT: Answer:
\end{lstlisting}

\textbf{MATH}

\begin{lstlisting}
HUMAN: Problem:\nFind the domain of the expression $\\frac{{\sqrt{{x-2}}}}{{\sqrt{{5-x}}}}$.}}\nSolution:"
BOT: The expressions inside each square root must be non-negative. Therefore, $x-2 \ge 0$, so $x\ge2$, and $5 - x \ge 0$, so $x \le 5$. Also, the denominator cannot be equal to zero, so $5-x>0$, which gives $x<5$. Therefore, the domain of the expression is $\\boxed{{[2,5)}}$.\nFinal Answer: The final answer is $[2,5)$. I hope it is correct.\n
HUMAN: Problem:\nIf $\det \mathbf{{A}} = 2$ and $\det \mathbf{{B}} = 12,$ then find $\det (\mathbf{{A}} \mathbf{{B}}).$\nSolution:
BOT: We have that $\det (\mathbf{{A}} \mathbf{{B}}) = (\det \mathbf{{A}})(\det \mathbf{{B}}) = (2)(12) = \\boxed{{24}}.$\nFinal Answer: The final answer is $24$. I hope it is correct.\n
HUMAN: Problem:\nTerrell usually lifts two 20-pound weights 12 times. If he uses two 15-pound weights instead, how many times must Terrell lift them in order to lift the same total weight?\nSolution:
BOT: If Terrell lifts two 20-pound weights 12 times, he lifts a total of $2\cdot 12\cdot20=480$ pounds of weight. If he lifts two 15-pound weights instead for $n$ times, he will lift a total of $2\cdot15\cdot n=30n$ pounds of weight. Equating this to 480 pounds, we can solve for $n$: \\begin{{align*}} 30n&=480\\\\ \Rightarrow\qquad n&=480/30=\\boxed{{16}} \end{{align*}}\nFinal Answer: The final answer is $16$. I hope it is correct.\n
HUMAN: Problem:\nIf the system of equations: \\begin{{align*}} 6x-4y&=a,\\\\ 6y-9x &=b. \end{{align*}}has a solution $(x, y)$ where $x$ and $y$ are both nonzero, find $\\frac{{a}}{{b}},$ assuming $b$ is nonzero.\nSolution:
BOT: If we multiply the first equation by $-\\frac{{3}}{{2}}$, we obtain $$6y-9x=-\\frac{{3}}{{2}}a.$$Since we also know that $6y-9x=b$, we have $$-\\frac{{3}}{{2}}a=b\Rightarrow\\frac{{a}}{{b}}=\\boxed{{-\\frac{{2}}{{3}}}}.$$\nFinal Answer: The final answer is $-\\frac{{2}}{{3}}$. I hope it is correct.\n
HUMAN: Problem:\n{problem}\nSolution:\n
\end{lstlisting}

\textbf{GSM8K}

\begin{lstlisting}
HUMAN: Question: Angelo and Melanie want to plan how many hours over the next week they should study together for their test next week. They have 2 chapters of their textbook to study and 4 worksheets to memorize. They figure out that they should dedicate 3 hours to each chapter of their textbook and 1.5 hours for each worksheet. If they plan to study no more than 4 hours each day, how many days should they plan to study total over the next week if they take a 10-minute break every hour, include 3 10-minute snack breaks each day, and 30 minutes for lunch each day?\nLet's think step by step\nAnswer: ,

BOT: Angelo and Melanie think they should dedicate 3 hours to each of the 2 chapters, 3 hours x 2 chapters = 6 hours total.\nFor the worksheets they plan to dedicate 1.5 hours for each worksheet, 1.5 hours x 4 worksheets = 6 hours total.\nAngelo and Melanie need to start with planning 12 hours to study, at 4 hours a day, 12 / 4 = 3 days.\nHowever, they need to include time for breaks and lunch. Every hour they want to include a 10-minute break, so 12 total hours x 10 minutes = 120 extra minutes for breaks.\nThey also want to include 3 10-minute snack breaks, 3 x 10 minutes = 30 minutes.\nAnd they want to include 30 minutes for lunch each day, so 120 minutes for breaks + 30 minutes for snack breaks + 30 minutes for lunch = 180 minutes, or 180 / 60 minutes per hour = 3 extra hours.\nSo Angelo and Melanie want to plan 12 hours to study + 3 hours of breaks = 15 hours total.\nThey want to study no more than 4 hours each day, 15 hours / 4 hours each day = 3.75\nThey will need to plan to study 4 days to allow for all the time they need.\nThe answer is 4,

HUMAN :Question: Mark's basketball team scores 25 2 pointers, 8 3 pointers and 10 free throws.  Their opponents score double the 2 pointers but half the 3 pointers and free throws.  What's the total number of points scored by both teams added together?\nLet's think step by step\nAnswer:,

BOT: Mark's team scores 25 2 pointers, meaning they scored 25*2= 50 points in 2 pointers.\nHis team also scores 6 3 pointers, meaning they scored 8*3= 24 points in 3 pointers\nThey scored 10 free throws, and free throws count as one point so they scored 10*1=10 points in free throws.\nAll together his team scored 50+24+10= 84 points\nMark's opponents scored double his team's number of 2 pointers, meaning they scored 50*2=100 points in 2 pointers.\nHis opponents scored half his team's number of 3 pointers, meaning they scored 24/2= 12 points in 3 pointers.\nThey also scored half Mark's team's points in free throws, meaning they scored 10/2=5 points in free throws.\nAll together Mark's opponents scored 100+12+5=117 points\nThe total score for the game is both team's scores added together, so it is 84+117=201 points\nThe answer is 201\n,

HUMAN: Question: Bella has two times as many marbles as frisbees. She also has 20 more frisbees than deck cards. If she buys 2/5 times more of each item, what would be the total number of the items she will have if she currently has 60 marbles?\nLet's think step by step\nAnswer:,

BOT: When Bella buys 2/5 times more marbles, she'll have increased the number of marbles by 2/5*60 = 24\nThe total number of marbles she'll have is 60+24 = 84\nIf Bella currently has 60 marbles, and she has two times as many marbles as frisbees, she has 60/2 = 30 frisbees.\nIf Bella buys 2/5 times more frisbees, she'll have 2/5*30 = 12 more frisbees.\nThe total number of frisbees she'll have will increase to 30+12 = 42\nBella also has 20 more frisbees than deck cards, meaning she has 30-20 = 10 deck cards\nIf she buys 2/5 times more deck cards, she'll have 2/5*10 = 4 more deck cards.\nThe total number of deck cards she'll have is 10+4 = 14\nTogether, Bella will have a total of 14+42+84 = 140 items\nThe answer is 140\n,

HUMAN: Question: A group of 4 fruit baskets contains 9 apples, 15 oranges, and 14 bananas in the first three baskets and 2 less of each fruit in the fourth basket. How many fruits are there?\nLet's think step by step\nAnswer:,

BOT: For the first three baskets, the number of apples and oranges in one basket is 9+15=24\nIn total, together with bananas, the number of fruits in one basket is 24+14=38 for the first three baskets.\nSince there are three baskets each having 38 fruits, there are 3*38=114 fruits in the first three baskets.\nThe number of apples in the fourth basket is 9-2=7\nThere are also 15-2=13 oranges in the fourth basket\nThe combined number of oranges and apples in the fourth basket is 13+7=20\nThe fourth basket also contains 14-2=12 bananas.\nIn total, the fourth basket has 20+12=32 fruits.\nThe four baskets together have 32+114=146 fruits.\nThe answer is 146\n,

HUMAN: Question: {question}\nLet's think step by step\nAnswer:

\end{lstlisting}

\textbf{TheoremQA}

\begin{lstlisting}
HUAMN: You are a mathematician, you are supposed to answer the given question. You need to output the answer in your final sentence like "Therefore, the answer is ...". The answer can only be one of the following forms:\n1. a numerical value like 0.1, no symbol and no unit at all.\n2. a list of number like [2, 3, 4].\n3. True/False.\n4. an option like (a), (b), (c), (d)\nQuestion: {Question}\nLet\'s think step by step.
\end{lstlisting}





\section{Detailed Explanation of the Scaling Law Formula}
\label{app:explanation_scaling_law}
The scaling law formula, as given by:
\begin{equation}
  P = f(N, D, C) = a \cdot N^{\alpha} \cdot D^{\beta} \cdot C^{\gamma},
\end{equation}
provides a mathematical representation of how the performance ($P$) of machine learning models scales with respect to three critical factors: model size ($N$), data size ($D$), and computational budget ($C$). Each of these components plays a crucial role in the model's ability to learn from data and perform tasks accurately. Here is a detailed explanation of each parameter within the formula:

$P$ (Performance Metric): This denotes the measure of effectiveness or accuracy of the model in performing its designated task. Performance is often quantified in terms of the model's loss ($L$), where a lower loss signifies a higher performance. However, performance could also be measured using other metrics such as accuracy, F1 score, or any other relevant evaluation criteria depending on the specific task and model.

$N$ (Model Size): This represents the size of the model, quantified by the count of its parameters. A larger model size ($N$) means more parameters are available for the model to learn complex patterns and relationships within the data. However, increasing the model size also demands more computational resources for training and inference.

$D$ (Data Size): This is measured in terms of the number of tokens or data points available for training the model. A larger data size ($D$) provides the model with more examples to learn from, potentially improving its generalization capabilities and performance on unseen data.

$C$ (Computational Budget): This reflects the total computational resources allocated for training the model. It encompasses not just the raw compute power but also includes considerations of training time and energy consumption. The computational budget is effectively a proxy for the total amount of compute effort expended in training the model.

$a, \alpha, \beta, \gamma$ (Constants): These are constants that depend on the specific architecture of the model and the task it is being trained for. The constants $\alpha, \beta, \gamma$ are exponents that dictate how performance scales with changes in model size, data size, and computational budget, respectively. The constant $a$ serves as a scaling factor that adjusts the overall scale of the performance metric.

The scaling law formula thus provides a theoretical framework to predict how adjustments in model size, data size, and computational resources are likely to impact the performance of machine learning models. This formula is instrumental in guiding researchers and practitioners towards making informed decisions about resource allocation during the model development process, aiming for an optimal balance between these factors to achieve the best possible model performance.

\end{document}